\def\isarxiv{1} 

\ifdefined\isarxiv
\documentclass[11pt]{article}
\usepackage{amsthm}
\usepackage{subfig}
\usepackage{wrapfig,epsfig}
\else
\documentclass[a4paper,USenglish]{lipics-v2021}
\nolinenumbers
\fi

\usepackage{amsmath}
\usepackage{amssymb}
\usepackage{amsthm}
\usepackage{algorithm}
\usepackage{color}
\usepackage{algpseudocode}
\ifdefined\isarxiv
\usepackage[english]{babel}

\usepackage{graphicx}
\usepackage{grffile}
\usepackage{epstopdf}
\usepackage{url}
\usepackage{color}
\usepackage{epstopdf}

\usepackage[T1]{fontenc}
\usepackage{bbm}
\usepackage{comment}
\usepackage{dsfont}
\fi

\ifdefined\isarxiv

\usepackage{tikz}
\usepackage[hidelinks]{hyperref}  
\hypersetup{colorlinks=true,citecolor=blue,linkcolor=blue}
\usetikzlibrary{arrows}
\usepackage{xspace}
\else 
\fi

\ifdefined\isarxiv
\usepackage[margin=1in]{geometry}
\else 

\fi

\ifdefined\isarxiv
\newtheorem{theorem}{Theorem}[section]
\newtheorem{lemma}[theorem]{Lemma}
\newtheorem{definition}[theorem]{Definition}

\newtheorem{corollary}[theorem]{Corollary}

\newtheorem{assumption}[theorem]{Assumption}

\newtheorem{fact}[theorem]{Fact}
\newtheorem{remark}[theorem]{Remark}
\newtheorem{claim}[theorem]{Claim}

\else
\newtheorem{fact}[theorem]{Fact}

\fi

\newcommand{\wh}{\widehat}
\newcommand{\wt}{\widetilde}

\newcommand{\eps}{\epsilon}

\newcommand{\R}{\mathbb{R}}

\renewcommand{\varepsilon}{\epsilon}

\renewcommand{\bar}{\overline}
\renewcommand{\eps}{\epsilon}
\renewcommand{\d}{\mathsf{d}}
\newcommand{\poly}{\mathrm{poly}}
\newcommand{\diag}{\mathrm{diag}}

\newcommand{\Tmat}{\mathcal{T}_{\mathrm{mat}}}
\newcommand{\vect}{\mathrm{vec}}

\newcommand{\tensorsketch}{\mathsf{TensorSketch}}
\newcommand{\tensorsrht}{\mathsf{TensorSRHT}}
\newcommand{\nnz}{\mathrm{nnz}}
\newcommand{\J}{\mathsf{J}}

\DeclareMathOperator*{\E}{{\mathbb{E}}}
\DeclareMathOperator*{\Var}{{\bf {Var}}}

\DeclareMathOperator{\supp}{supp}
\DeclareMathOperator*{\tr}{tr}

\definecolor{mycrimson}{RGB}{165,28,48}

\makeatletter
\newcommand*{\RN}[1]{\expandafter\@slowromancap\romannumeral #1@}
\makeatother

\usepackage{lineno}

\ifdefined\isarxiv
\title{Training Multi-Layer Over-Parametrized Neural Network in Subquadratic Time\thanks{A preliminary version of this paper appeared at the 15th Innovations in Theoretical Computer Science (ITCS 2024).}}
\else
\title{Training Multi-Layer Over-Parametrized Neural Network in Subquadratic Time}
\titlerunning{Training Multi-Layer Over-Parametrized Neural Network in Subquadratic Time}
\fi

\ifdefined\isarxiv
\date{}
\author{
Zhao Song\thanks{\texttt{zsong@adobe.com}. Adobe Research.}
\and
Lichen Zhang\thanks{\texttt{lichenz@mit.edu}. Massachusetts Institute of Technology.}
\and
Ruizhe Zhang\thanks{\texttt{ruizhe@utexas.edu}. Simons Institute for the Theory of Computing.}
}
\else
\author{Zhao Song}{Adobe Research, CA, USA}{zsong@adobe.com}{}{}
\author{Lichen Zhang}{Massachusetts Institute of Technology, MA, USA}{lichenz@mit.edu}{}{Supported by NSF CCF-1955217 and NSF DMS-2022448.}
\author{Ruizhe Zhang}{Simons Institute for the Theory of Computing, CA, USA}{ruizhe@utexas.edu}{}{}
\authorrunning{Z. Song, L. Zhang, and R. Zhang}
\Copyright{Zhao Song, Lichen Zhang, and Ruizhe Zhang}

\relatedversion{A full version of the paper is available at \url{https://arxiv.org/abs/2112.07628}.}
\acknowledgements{We would like to thank Yin Tat Lee for discussing the motivation of this problem. We would also like to thank Jan van den Brand, Binghui Peng, Omri Weinstein, and David P. Woodruff
for helpful discussions in the early stage of this project. We would like to thank Pravesh
Kothari and Gary Miller for helpful discussions on the data structures. The corrections in this version were found and written with the help of Claude Opus 5.5. Running as an agent in Claude Code, this large language model also mainly generated the section on simpler algorithms for last-layer training and, in the full version, the sections on an incremental algorithm and on the exact Gram solve, with their theorems, algorithms and proofs, and the paragraph on the relation to the published algorithm. Independent Claude Opus 5.5 reviewers then checked these sections.} The authors have verified the corrections and take full responsibility for the content of this paper.
\EventEditors{Venkatesan Guruswami}
\EventNoEds{1}
\EventLongTitle{15th Innovations in Theoretical Computer Science Conference (ITCS 2024)}
\EventShortTitle{ITCS 2024}
\EventAcronym{ITCS}
\EventYear{2024}
\EventDate{January 30 to February 2, 2024}
\EventLocation{Berkeley, CA, USA}
\EventLogo{}
\SeriesVolume{287}
\ArticleNo{93}
\fi

\begin{document}

\ifdefined\isarxiv

\begin{titlepage}
\maketitle
\begin{abstract}

We consider the problem of training the last layer of a multi-layer over-parametrized neural network to minimize the empirical risk induced by a loss function. In the typical setting of over-parametrization, the network width $m$ is much larger than the data dimension $d$ and the number of training samples $n$ ($m=\poly(n,d)$), which induces a prohibitive large weight matrix $W\in \R^{m\times m}$ per layer above the first. Naively, when the layer inputs are dense, one has to pay $O(m^2)$ time to read the weight matrix and evaluate the neural network function in both forward and backward computation. In this paper, we show how to reduce the training cost per iteration. Specifically, we propose a framework that uses $O(m^2nL)$ time in the initialization phase and achieves \emph{a truly subquadratic amortized cost per iteration} in terms of $m$, i.e., $m^{2-\Omega(1)}$ per iteration. The speedup of this framework comes from a sparse activation, a shifted ReLU whose threshold at each layer is proportional to the norm of the layer input. With this threshold every layer is positively homogeneous, so our convergence proof needs a width only polynomial in the depth $L$. Let $\lambda_L$ and $\lambda_*$ be the smallest eigenvalues of the neural tangent kernel and of the kernels of the hidden layers. If the threshold of each layer is $\sqrt{2\gamma\log m}$ standard deviations of its initial pre-activations, for a constant $\gamma\in(0,1)$, then with high probability at most a fraction $2m^{-\gamma}$ of the neurons of each layer fire at initialization, and a width with $m^{1-\gamma}\geq\wt O(n^4L^4\lambda_L^{-4}\lambda_*^{-4})$ suffices. The inputs of the trained layer are fixed and sparse. In this setting, simpler algorithms without a data structure reach the same cost per iteration, and maintaining the products of the last layer with its inputs lowers it to $mn^{\omega-1+o(1)}$, where $\omega$ is the matrix multiplication exponent. Our result has implications beyond standard over-parametrization theory, as it can be viewed as an idealized model of maintaining a small amount of state on top of a frozen pre-trained model to speed up the fine-tuning process, a core procedure to deploy large language models (LLM). Our guarantees assume that the frozen layers are random, and for a pre-trained model only the cost per iteration of maintaining these products carries over.

\end{abstract}
\thispagestyle{empty}
\end{titlepage}

\newpage

\else

\maketitle
\begin{abstract}

\end{abstract}

\fi

\section{Introduction}

Over-parameterized neural networks represent one of the most extensively researched and widely utilized models in contemporary machine learning. These networks are among the first to offer convergence guarantees for prevalent deep network training methodologies, as indicated by numerous studies~\cite{ll18,jgh18,dzps19,als19_dnn,als19_rnn,dllwz19}. Beyond the sound convergence theory, the recent surge of large language models (LLMs) also raises the stakes of training such networks efficiently. In particular, the over-parametrization setting aligns well with the \emph{fine-tuning} process of LLMs. To make it concrete, recall fine-tuning is the procedure of adapting an LLM for a particular set of focused data so that it can be more specialized in certain domains. Since the size of fine-tuning data is much smaller compared to the size of actual training data of an LLM, fine-tuning can typically be implemented in an efficient manner. Moreover, the over-parametrization theory requires the number of parameters of a network (expressed as the \emph{width} of the network) to be orders of magnitude larger than that of the input dataset, which is exactly the setup for fine-tuning as the size of an LLM is much larger than the fine-tuning dataset. Speeding up the fine-tuning process is one of the core algorithmic tasks for large language model applications, and many works attempt to exploit the inherent over-parametrization structure of the process. The work~\cite{hsw+22} explicitly utilizes the \emph{low-rank} property of the gradient and derives a heuristic fine-tuning procedure that greatly reduces the parameters required. 

Meanwhile, developments in the department of convex optimization have led to breakthroughs in tasks such as linear programming~\cite{cls19,jswz20,blss20,b20_lp,gs22}, empirical risk minimization~\cite{lsz19,qszz23}, the cutting plane method~\cite{jlsw20}, semidefinite programming~\cite{jklps20,hjs+21,gs22}, and sum-of-squares optimization~\cite{jnw22}. However, the techniques developed in these works are tailored for optimizing convex objectives, which are not applicable to the training of over-parameterized networks. Furthermore, existing research on efficient training for over-parameterized networks primarily focuses on two-layer networks~\cite{bpsw21,syz21,hswz22,als+22}, which differs substantially from the networks utilized in real-world applications.

In this work, we develop a fast algorithm for training the last layer of deep over-parameterized networks. Specifically, given $n$ data points of dimension $d$, we consider training the last layer of a fully connected neural network of depth $L \geq 2$ with a shifted ReLU activation. The threshold of the activation at each layer is proportional to the norm of the layer input. The network has width $m=\poly(n, L,1/\lambda_L,1/\lambda_*)$, where $\lambda_L$ and $\lambda_*$ are the smallest eigenvalues of the neural tangent kernel and of the kernels of the hidden layers (Theorem~\ref{thm:conv_main}). To ensure the convergence of the training process, the network width $m$ is typically chosen to be a large polynomial in $n$ and $d$. For deep neural networks, the current best over-parameterization width is on the order of $n^4$~\cite{cczg21}. The input layer is represented as a matrix of size $m \times d$, and all intermediate layers are of size $m \times m$. The training typically consists of a forward pass and a backward pass. The forward pass involves multiplying weight matrices $W \in \R^{m \times m}$ with vectors $h \in \R^m$. Without any structure on $W$ and $h$, this step would take $\Theta(m^2)$ time, which is prohibitively large. We hence ask the following question:

\begin{center}
    {\it Is it possible to reduce the cost per iteration during the training to truly subquadratic in $m$?}
\end{center}

For the last layer, which is what we train, the answer is yes, and the reason is simple. Its inputs are fixed, so a Gauss-Newton step that maintains the products of the last weight matrix with these inputs takes $mn^{\omega-1+o(1)}$ time, where $\omega$ is the matrix multiplication exponent. Under the width condition of Theorem~\ref{thm:conv_main}, this is subquadratic in $m$, and because few neurons of the trained layer fire, an incremental implementation lowers the cost further for suitable thresholds (Section~\ref{sec:plain}). Our main algorithm (Algorithm~\ref{alg:main}) answers the question with a low-rank data structure and a sketch, and Section~\ref{sec:plain} gives simpler and faster algorithms.

For training two-layer over-parameterized neural networks, two interesting results have been obtained by van den Brand, Peng, Song, and Weinstein~\cite{bpsw21} and Song, Yang, and Zhang \cite{syz21}. However, we note that the settings both of these papers studied differ from ours, specifically:

\begin{itemize}
    \item In the two-layer case, they only need to train a weight matrix of size $m \times d$. Since evaluating one data point in $d$-dimensional space for neural network functions takes $O(md)$ time, the cost per iteration bound they aim to match \cite{bpsw21} or beat \cite{syz21} is $O(md)$. 
    \item In the multi-layer case, instead of having only a weight matrix of size $m \times d$, we will have at least one weight matrix of size $m \times m$. Evaluating one data point in $m$-dimensional space for neural network functions takes $O(m^2)$ time when the sparsity of the hidden vectors is ignored.
\end{itemize}

In~\cite{bpsw21}, they provide an algorithm that can adaptively choose the step size for different gradient directions associated with different data points, which is one of the goals we want to achieve. However, their method involves forming a Jacobian matrix of the data, which is of size $n \times md$ in the two-layer case, but of size $n \times m^2$ in our case. Hence, with the Jacobian formed as a dense matrix, their algorithm can only imply an $O(nm^2)$ cost per iteration, which cannot match our subquadratic goal.

In \cite{syz21}, they surpass the $nmd$ barrier by leveraging two key ideas: using a shifted ReLU activation and, through a sparsity analysis, showing that only $o(m)$ number of neurons are fired up\footnote{Such a phenomenon has been observed in practice as in~\cite{cmf+20,clp+21,mam+23}.}. They further use a data structure to preprocess both data points and training weights. However, their algorithm only works for small $d$ due to the exponential dependence on $d$ in the preprocessing phase. In a multiple-layer neural network, instead of having only a weight matrix of size $m \times d$, we will have at least one weight matrix of size $m \times m$; this directly translates to a high dependence on $m$ in the preprocessing phase. As our hope is for an algorithm with ideally $O(m^2nL)$ time to preprocess, their data structure is infeasible for any practical application.

We also note two more recent works~\cite{hswz22,als+22} that particularly focus on developing efficient algorithms for training two-layer over-parameterized networks. Both of them make use of tree-based data structures to find activated neurons in sublinear time. These approaches, unfortunately, do not scale well to multi-layer settings, as they rely on the input dimension being much smaller than the network width, while in multi-layer settings, the weight matrix is square.

Our method overcomes the shortcomings of these prior approaches, as it can not only choose the step size adaptively but also achieve an amortized per iteration cost of $o(m^2)$. As we will see later, our algorithm also leverages the \emph{low-rank} property of the gradient but in a \emph{provable} fashion. Moreover, our approach goes beyond the standard first-order methods, such as gradient descent or stochastic gradient descent, as we provide a highly efficient implementation of the Gauss-Newton method. In our last-layer setting, simpler algorithms without the data structure do better (Section~\ref{sec:plain}).

\paragraph*{A Change from the Published Version.} The network of this version differs from the one in the published version (ITCS 2024) mainly in its thresholds. In the published version, every layer compares its pre-activations with the same fixed threshold $\sqrt{2/m}\,b$, where $b$ is the threshold parameter of Theorem~\ref{thm:conv_main}. Here, the threshold of layer $\ell$ on input $x_i$ is $\sqrt{2/m}\,b\,\|h_{i,\ell-1}\|_2$, which is proportional to the norm of the input $h_{i,\ell-1}$ of the layer (Definition~\ref{def:architecture}). Equivalently, each layer normalizes its input, applies the fixed-threshold activation and multiplies the output by the norm of the input. We make this change because with the fixed threshold, the deviations of the norms of the hidden vectors from $1$ can grow geometrically with the depth, and our analysis of that network needs a width with a factor exponential in the depth. With the norm-scaled threshold, every layer is positively homogeneous, as for ReLU, so these deviations add up instead of multiplying (Remark~\ref{remark:threshold}). Each neuron still fires with probability exactly $P_b=1-\Phi(b)$, where $\Phi$ is the distribution function of ${\cal N}(0,1)$, so the sparsity and the cost per iteration are unchanged. Since only the last layer is trained, the hidden vectors never change. The thresholds are therefore computed once at initialization, in $O(m)$ time per layer and data point, and then kept fixed. The other change is that the trained last layer uses a scaled copy of the continuous shifted ReLU $\max\{x-\tau,0\}$, where $\tau$ is its threshold, instead of the thresholded activation, so that the output of the network is a continuous function of $W_L$. This changes neither the sparsity nor the derivative of the activation away from the threshold.

\subsection{Our Results}
Our main results can be summarized in the following theorems: the first analyzes the convergence behavior of a general Gram-based optimization framework and the second designs an efficient algorithm to achieve subquadratic amortized cost per iteration.

Throughout this paper, we use $n$ to denote the number of training data points, $d$ to represent the dimension of input data points, $m$ for the width of the network, and $L$ for the number of layers in the network. We denote the smallest eigenvalue of the neural tangent kernel induced by our neural network as $\lambda_L$, the smallest eigenvalue among the kernels of the hidden layers $1,\dots,L-2$ as $\lambda_*$ (Definition~\ref{def:gram_multilayer}), and the prediction of the neural network at time $t$ as $f_t \in \R^n$.

Our first theorem demonstrates the fast convergence rate of our algorithm.
\begin{theorem}[Convergence]
\label{thm:conv_main}
Let the threshold parameter of the shifted ReLU be $b=\sqrt{2\gamma\log m}$ for a constant $\gamma\in(0,1)$ (Remark~\ref{remark:k}), with the threshold of each layer proportional to the norm of the layer input (Definition~\ref{def:architecture}). Suppose that $\lambda_L>0$ and $\lambda_*>0$, where $\lambda_*\in(0,1]$ is the smallest eigenvalue of the kernels of the hidden layers $1,\dots,L-2$ (Definition~\ref{def:gram_multilayer}), and that the width of the neural network satisfies
\begin{align*}
    m^{1-\gamma}\geq C\, n^4L^4\lambda_L^{-4}\lambda_*^{-4}\log^{c}(mnL)
\end{align*}
for large enough absolute constants $C$ and $c$. Then, there exists an algorithm (Algorithm~\ref{alg:main}) such that, for every number of steps $T\leq\poly(n)$, over the randomness of initialization of the network and the algorithm, with probability at least $1 - 1/\poly(n)$, we have for all $t\in\{0,\dots,T-1\}$
\begin{align*}
    \|f_{t+1} - y\|_2 \leq & ~ \frac{1}{3}\|f_t - y\|_2.
\end{align*}
In particular, a width $m=(nL/(\lambda_L\lambda_*))^{O(1/(1-\gamma))}$, polynomial in $n$, $L$, $\lambda_L^{-1}$ and $\lambda_*^{-1}$, suffices. The kernels, and hence $\lambda_L$ and $\lambda_*$, depend on $m$ only through $b$. Moreover, $\lambda_L\leq 1/(1+b^2)$.
\end{theorem}

The above theorem establishes the linear convergence rate of our method. Compared to the one-hidden layer case, our analysis must control how the errors of the $L$ layers build up. Because the threshold of each layer is proportional to the norm of its input, each layer is positively homogeneous, as for ReLU. Hence the ratio $\|h_{i,\ell}\|_2^2/\|h_{i,\ell-1}\|_2^2$ has mean $1$ and a distribution that does not depend on the earlier layers, so the norm deviations of the layers add up instead of multiplying (Lemma~\ref{lem:als_71}). Only about $k=m^{1-\gamma}$ neurons of a layer fire, so each layer adds an error of order $k^{-1/2}$ to the inner products of the hidden vectors. The width condition is therefore on $k$ rather than on $m$. To compare the hidden vectors with the kernels layer by layer, we view the kernel map $\psi$ of one layer as a function of the correlation of its two inputs. This function is a power series with nonnegative coefficients and satisfies $\psi(1)=1$. Its slope at a correlation $s\in[0,1)$ is therefore at most $(1-\psi(s))/(1-s)$, the ratio of the distances to $1$ after and before the layer. Along the kernel correlations these ratios telescope, so the errors of all layers together are amplified by at most a factor $O(1/\lambda_*)$, not by a factor per layer (Lemma~\ref{lem:init_min_eigen_H}). The factor $\lambda_*^{-4}$ in the width condition comes from this comparison. We do not know whether it is needed (Remark~\ref{remark:rate_H}).

Our convergence argument analyzes the training dynamics through the Gram matrix of the last layer. Previous works train all hidden layers, a harder problem, and either depend exponentially on the number of layers $L$ in network width~\cite{dllwz19} or directly analyze training dynamics, requiring a data-separable assumption~\cite{als19_dnn}. Our width condition is polynomial in $L$ for two reasons. The norm-scaled threshold keeps every layer positively homogeneous, and the kernel map of a layer is a power series in the correlation with nonnegative coefficients. Our analysis builds upon these works and bounds the smallest eigenvalue of the Gram matrix of the last layer from below by a constant fraction of that of the multi-layer neural tangent kernel, at initialization and under perturbations of the last weight matrix. The framework we developed is also compatible with sparser, shifted ReLU activations.

Since only the last layer moves, and it stays close to its initialization, we expect generalization bounds for neural tangent kernels~\cite{adhlw19} to extend to our setting, but we do not prove this. The generalization bound of the shifted ReLU has also been carefully examined in~\cite{yjz+23}.

The next theorem concerns the \emph{cost per iteration} of our algorithm.
\begin{theorem}[Runtime]
\label{thm:alg_main}
Let $\alpha$ be the dual matrix multiplication exponent~\cite{wxxz23,OpenAIMM26} and let $a:=\min\{\alpha,1/2\}$. Let the threshold parameter of the shifted ReLU be $b=\sqrt{2\gamma\log m}$ for a constant $\gamma\in[a,1)$, for instance $\gamma=1/2$, or $\gamma=a$, and suppose that the width $m$ satisfies the condition of Theorem~\ref{thm:conv_main} for this $\gamma$. Then there exists a randomized algorithm (Algorithm~\ref{alg:main}) that trains the last layer of a multi-layer neural network of width $m$ such that, with probability at least $1-1/\poly(n)$ over the initialization, the amortized cost per training iteration is $nm^{2-a+o(1)}=O(nm^{2-\Omega(1)})$. Furthermore, the width condition implies $n^4\leq m^{1-a}$, and in particular $m\geq n^4$, so the amortized cost per iteration is at most
\begin{align*}
    m^{(9-5a)/4+o(1)}\leq m^{2.25-\min\{\alpha,1/2\}+o(1)}.
\end{align*}
\end{theorem}

We improve the overall time of training the last layer of multi-layer over-parametrized networks from ${\cal T}_{\mathrm{init}} + T \cdot \wt O(nm^2)$, a bound that ignores the sparsity of the fixed inputs of the trained last layer, to ${\cal T}_{\mathrm{init}} + T \cdot o(nm^2)$, where ${\cal T}_{\mathrm{init}}$ is the initialization time of training, typically taking $O(m^2nL)$ time (Theorem~\ref{thm:alg_app}). In an idealized model of fine-tuning, in which the frozen layers are random as at initialization, our algorithmic result can be interpreted as a data structure for fine-tuning LLMs: given a large, pre-trained language model whose weights are ``frozen'' during fine-tuning, we only need to initialize the data structure once. The data structure can then be queried to implement fine-tuning steps.

Theorem~\ref{thm:alg_main} is the runtime result of the published version of this paper (ITCS 2024), in the form adapted here. In this setting, the problem is easier than the published version presents it, and simpler algorithms do better (Section~\ref{sec:plain}). Storing $W_L$ explicitly attains the bound of Theorem~\ref{thm:alg_main} at every step, without amortization. Maintaining the products $W_Lh_{i,L-1}$ costs $mn^{\omega-1+o(1)}$ time per iteration, which is $O(m^{1.3145})$ under the width condition since $\omega<2.258$ over every field of characteristic zero~\cite[Theorem~1.1 and Corollaries~14.1--14.2]{OpenAIMM26} (Section~\ref{sec:plain}), and an incremental implementation costs $O(n^2k)$ time per iteration, which is $o(m)$ for every constant $\gamma>1/3$. In the fine-tuning view above, maintaining the products $W_Lh_{i,L-1}$ directly serves the same purpose as the data structure at a lower cost per step. Our guarantees assume that the frozen layers are random, as at initialization. For a pre-trained model, only the cost per step of Theorem~\ref{thm:plain_gram} carries over, since it uses neither the sparsity nor the randomness of the frozen layers.

To achieve subquadratic time, we cannot afford to perform matrix-vector products between the weight matrices and any dense vectors. However, any algorithm requiring first- or second-order information needs the products $W_Lh_{i,L-1}$ to evaluate the network prediction. When implementing the Gauss-Newton method, we also have a Jacobian matrix of size $n \times m^2$, so forming it exactly as a dense matrix would already take $\Theta(nm^2)$ time. Moreover, note that the update is also an $m \times m$ matrix, meaning that explicitly applying the update as a dense matrix would also cost $\Theta(m^2)$ time. In the last-layer training that we analyze, the update changes only the columns of $W_L$ in the supports of the fixed sparse vectors $h_{i,L-1}$, so it can be applied explicitly in $O(nmk)$ time (Section~\ref{sec:plain}). In the implementation of Algorithm~\ref{alg:main}, to avoid forming the Jacobian and the update as dense matrices, we exploit the fact that the gradient is low-rank (rank $n$). Thus, one can compute a rank-$n$ factorization and use it to support fast matrix-vector products. We also observe that each row of the Jacobian matrix can be formulated as a tensor product of two vectors. Therefore, we can use fast randomized linear algebra to approximate the tensor product efficiently.

\subsection{Related Work}

\ifdefined\isarxiv
\paragraph{Convex and Non-Convex Optimization.}
\else
\paragraph*{Convex and Non-Convex Optimization.}
\fi
In convex optimization problems, such as linear programming~\cite{v89_lp,ds08,ls14,cls19,ye20,sy21,jswz20}, empirical risk minimization~\cite{lsz19,qszz23}, the cutting plane method~\cite{jlsw20}, maximum bipartite matching and maxflow \cite{lszzz22,amv21}, and semidefinite programming~\cite{lsw15,jklps20,hjs+21,gs22}, one typically uses an algorithm that can dynamically adjust the search direction and step size to reduce the iteration count. Due to the prohibitively high cost of implementing one step of these methods, most of these works focus on improving the \emph{cost per iteration}.

In the non-convex setting, there is a vast body of ongoing works~\cite{jr15,brb17,pw17,abh17,blh18,cgh+19,zmg19,bpsw21,ygs+21} that try to improve the iteration count and cost per iteration, especially in the setting of training deep neural networks. As shown in~\cite{cgh+19}, it is possible to exploit the equivalence between over-parametrized networks and neural tangent kernels to optimize an $n \times n$ matrix instead of an $m^2 \times m^2$ matrix, which is an important breakthrough in gaining speedup for such methods. Sketching and sampling-based methods can also be used to accelerate the computation of inverses of the Hessian matrix~\cite{pw17}. In spirit, our work most resembles~\cite{cgh+19} and~\cite{bpsw21}, in the sense that our optimization also works on an $n \times n$ Gram matrix. Our algorithm also makes use of sketching and sampling, as in~\cite{pw17}.

\paragraph*{Over-Parametrized Neural Networks.}
In the deep learning community, understanding the geometry and convergence behavior of various optimization algorithms on over-parametrized neural networks has received a lot of attention \cite{ll18,jgh18,dzps19,als19_dnn,als19_rnn,dllwz19,sy19,bpsw21,zczg20,zg19,cg19,lxs+19,lss+20,os20,cczg21,hlsy21,syz21,als+22,hswz22,gqsw22}. The seminal work of~\cite{jgh18} initiates the study of the \emph{neural tangent kernel} (NTK), which is a very useful analytical model to prove the convergence of training on over-parametrized networks. By over-parametrizing the neural network so that the network width is relatively large $(m \geq \Omega(n^4))$, one can show that the training dynamics on a neural network are almost the same as those on an NTK.

\paragraph*{Sketching.}
Using randomized linear algebra to reduce the dimension of problems and speed up algorithms for various problems has been a growing trend in the machine learning community~\cite{s06,cw13,w14} due to its wide range of applications, especially the efficient approximation of kernel matrices~\cite{anw14,akk+20,wz20,swyz21}. The standard ``Sketch-and-Solve'' paradigm \cite{cw13} involves reducing the dimension of the problem via sketching and then using a black box for the original problem to gain computational efficiency. Another line of work uses sketching as a preconditioner~\cite{w14,bpsw21} to obtain high-precision solutions. Sketching has also been successful in solving symmetric norm regression~\cite{swyzz19}, clustering and coverage~\cite{emmz22}, low-rank approximation~\cite{swz17,swz19}, and in distributed settings~\cite{bwz16}.

\paragraph*{Roadmap.} 
In Section~\ref{sec:preli_main}, we give a preliminary overview of the training setup considered in this paper. In Section~\ref{sec:notation_main}, we introduce the notations that will be used throughout the paper. In Section~\ref{sec:setup_main}, we consider the training setting more specifically. In Section~\ref{sec:tech_main}, we provide an overview of the techniques employed in this paper. In Section~\ref{sec:alg_main}, we examine the algorithmic tools utilized in this work to achieve subquadratic amortized cost per iteration. In Section~\ref{sec:conv_main}, we demonstrate various techniques to prove the convergence of our second-order method. In Section~\ref{sec:plain}, we give simpler algorithms for last-layer training. Finally, in Section~\ref{sec:discuss_main}, we summarize the results of this paper and point out some potential future directions.

\section{Preliminaries}
\label{sec:preli_main}
\subsection{Notations}
\label{sec:notation_main}
For any integer $n>0$, let $[n]$ denote the set $\{1,2,\cdots,n\}$. Let $\Pr[\cdot]$ denote probability and $\E[\cdot]$ denote expectation. 
 We use $\|x\|_2$ to denote the $\ell_2$ norm of a vector $x$. We use $\|A\|$ and $\|A\|_F$ to denote the spectral norm and the Frobenius norm of matrix $A$, respectively.  We use $A^\top$ to denote the transpose of matrix $A$. We use $I_m$ to denote the identity matrix of size $m\times m$. For $\alpha$ being a vector or matrix, we use $\|\alpha\|_0$ to denote the number of nonzero entries of $\alpha$. Given a real square matrix $A$, we use $\lambda_{\max}(A)$ and $\lambda_{\min}(A)$ to denote its largest and smallest eigenvalues respectively. Given a real matrix $A$, we use $\sigma_{\max}(A)$ and $\sigma_{\min}(A)$ to denote its largest and smallest singular values respectively. We use ${\cal N}(\mu,\sigma^2)$ to denote the Gaussian distribution with mean $\mu$ and variance $\sigma^2$. We use $\wt O(f(n))$ to denote $O(f(n)\cdot\poly \log(f(n))$. We use $\langle \cdot,\cdot \rangle$ to denote the inner product, when applying to two vectors, this denotes the standard dot product between two vectors, and when applying to two matrices, this means $\langle A,B\rangle = \tr[A^\top B]$, i.e., the trace of $A^\top B$.
 For a vector $x \in \R^n$, we use $\diag(x)$ to denote an $n \times n$ diagonal matrix where diagonal entries are from $x$. Given two vectors $x, y$, we use $x\otimes y$ to denote their tensor product and given two matrices $A, B$, we use $A\otimes B$ to denote their Kronecker product.

\subsection{Problem Setup}
\label{sec:setup_main}
\paragraph*{Architecture.} We first describe our network architecture. The network consists of $L$ hidden layers, represented by the weight matrices $W_1\in\R^{m\times d}$ and $W_\ell\in\R^{m\times m}$ for $\ell\in\{2,\dots,L\}$. The output layer consists of a vector $a \in \R^{m}$. We define the evaluation of the neural network as a prediction function $f: \R^{d} \rightarrow \R$
\begin{align*}
    f(W,x) = a^\top \phi( W_L ( \phi ( \cdots \phi(W_1 x ) ) ) ),
\end{align*}
where $\phi: \R \rightarrow \R$ is the shifted ReLU activation function ($\sigma_b(x)=\max\{x - b, 0\}$) applied coordinate-wise to a vector. The formal analysis uses a scaled version of this activation (Definition~\ref{def:architecture}). The hidden layers use a scaled thresholded activation $\phi$, and the trained last layer uses its continuous shifted version $\phi_L$. The threshold of a layer is proportional to the norm of the input of that layer. On input $x_i$, layer $\ell$ compares its pre-activations with $\tau_{i,\ell}:=\sqrt{2/m}\,b\,\|h_{i,\ell-1}\|_2$, where $h_{i,\ell-1}$ is the input of the layer defined below and $h_{i,0}:=x_i$. This threshold is the main departure from the published version, which uses the fixed threshold $\sqrt{2/m}\,b$ in every layer. The published version also uses the thresholded activation in the last layer (see the paragraph `A Change from the Published Version' in the introduction). The thresholds are computed at initialization and then kept fixed. We assume that the data points are normalized, $\|x_i\|_2=1$ for all $i\in[n]$, and that the labels are bounded, $|y_i|=O(1)$ for all $i\in[n]$.

We measure the loss via the squared-loss function:
\begin{align*}
    \mathcal{L}(W) = \frac{1}{2}\sum_{i=1}^n (y_i - f(W,x_i))^2.
\end{align*}
This is also the objective function for our training.

The prediction function $f_t: \R^{d \times n} \rightarrow \R^n$ is defined as
\begin{align*}
    f_t(X) = \begin{bmatrix}
    f(W(t),x_1) & 
    f(W(t),x_2) &
    \cdots  &
    f(W(t),x_n)
    \end{bmatrix}^\top .
\end{align*}

\paragraph*{Initialization.} Our neural networks are initialized as follows:
\begin{itemize}
    \item The weight matrices $W_1(0)\in\R^{m\times d}$ and $W_\ell(0)\in\R^{m\times m}$ for $\ell\in\{2,\dots,L\}$ are initialized such that each entry is sampled from $\mathcal{N}(0,\frac{2}{m})$.
    \item Each entry of $a$ is an i.i.d. sample from $\{-1,+1\}$ uniformly at random.
\end{itemize}

\paragraph*{Gradient.} In order to write the gradient in an elegant way, we define some artificial variables: for all $i \in [n]$
\begin{align}
g_{i,1} = W_1 x_i, \quad & h_{i,1} = \phi(W_1 x_i),  \nonumber \\
g_{i,\ell} = W_{\ell} h_{i,\ell-1}, \quad & h_{i,\ell} = \phi( W_{\ell} h_{i,\ell-1} ), \quad \forall \ell \in [L] \backslash \{1\} \nonumber
\end{align}
and
\begin{align*}
D_{i, 1} = &~\text{diag} \big(\phi'(W_1 x_i )\big), \quad \forall i \in [n] \\
D_{i, \ell} = &~ \text{diag} \big(\phi'(W_{\ell} h_{i, \ell-1})\big), \quad \forall i \in [n], \forall \ell \in [L] \backslash \{1\}
\end{align*}
Here, for input $x_i$ at layer $\ell$, the activation $\phi$ and its derivative $\phi'$ use the threshold $\tau_{i,\ell}$, which we leave implicit in the notation. Since the thresholds are kept fixed, they are constants for the gradient below, so the usual chain rule applies.

Using the definitions of $f$ and $h$, we have
\begin{align*}
    f(W,x_i) = a^\top h_{i,L}, \quad \in \R, \quad \forall i \in [n]
\end{align*}

We can compute the gradient of $\mathcal{L}$ in terms of $W_{\ell} \in \R^{m \times m}$, for all $\ell \geq 2$
\begin{align}
\frac{\partial \mathcal{L}(W)}{\partial W_{\ell}}
= \sum_{i = 1}^{n} ( f(W, x_i) - y_i) D_{i, \ell} \cdot \left( \prod_{j = \ell+1}^{L} W_{j}^{\top} D_{i, j} \right) a h_{i, \ell -1}^{\top}
\end{align}
Note that the gradient for $W_1 \in \R^{m \times d}$ is slightly different and cannot be written in general form. By the chain rule, the gradient of the variables in $W_1$ can be expressed as:
\begin{align*}
   \frac{\partial \mathcal{L}(W)}{\partial W_{1}} 
    = \sum_{i = 1}^{n} ( f(W, x_i) - y_i) D_{i, 1} \left( \prod_{j = 2}^{L} W_{j}^{\top} D_{i, j} \right) a x_{i}^{\top}
\end{align*}

It is worth noting that the gradient matrix is rank $n$, since it's a sum of $n$ rank-1 matrices. Since only $W_L$ is trained, we use the gradient above, and the Jacobian and the Gram matrix below, only for $\ell=L$.

\paragraph*{Jacobian.} For the last layer and time $t\in\{0,\dots,T\}$, we define the Jacobian matrix $J_{L,t} \in \R^{n \times m^2}$ via the following formulation:
\begin{align*}
     J_{L,t} 
    = \begin{bmatrix}
    \vect(\frac{\partial f(W(t),x_1)}{\partial W_L(t)})^\top \\
    \vect(\frac{\partial f(W(t),x_2)}{\partial W_L(t)})^\top \\
    \vdots\\
    \vect(\frac{\partial f(W(t),x_n)}{\partial W_L(t)})^\top
    \end{bmatrix}.
\end{align*}

The Gram matrix at time $t$ is then defined as $G_{L,t} = J_{L,t} J_{L,t}^\top \in \R^{n \times n}$ whose $(i,j)$-th entry is
\begin{align*}
\Big\langle \frac{\partial f(W(t),x_i)}{\partial W_L}, \frac{\partial f(W(t),x_j)}{\partial W_L} \Big\rangle.
\end{align*}
In Section~\ref{sec:analysis}, $J_{L,t}$ denotes the Jacobian with respect to $\wt W_L=\sqrt{m/2}\,W_L$, which is $\sqrt{2/m}$ times the matrix defined here, so its Gram matrix is $(2/m)\,G_{L,t}$.

\section{Technical Overview}
\label{sec:tech_main}
In this section, we give an overview of the techniques employed in this paper. In Section~\ref{sec:alg_main}, we showcase our algorithm and explain various techniques being used to obtain a subquadratic amortized cost per iteration. In Section~\ref{sec:conv_main}, we give an overview of the proof to show the convergence of our algorithm. 
\subsection{Subquadratic Time}
\label{sec:alg_main}

In this section, we study the different techniques being used to achieve the subquadratic amortized cost per iteration.


\begin{algorithm*}[!ht]\caption{Informal version of our algorithm.   
}
\label{alg:main}
\begin{algorithmic}[1]
\Procedure{OurAlgorithm}{$f,\{x_i,y_i \}_{i\in [n]},T$} \Comment{Theorem~\ref{thm:conv_main},\ref{thm:alg_main} }
    \State {\color{blue}/*Initialization*/}
    \State Initialize $W_\ell(0)$, $\forall \ell \in [L]$
    \State Store $h_{i,L-1}$ and $\tau_{i,L}=\sqrt{2/m}\,b\,\|h_{i,L-1}\|_2$ in memory, $\forall i\in [n]$ \Comment{Takes $O(nm^2L)$ time}
    \For{$t= 0 \to T-1$}
        \State {\color{blue}/*Forward computation*/}
        \State $v_{i,L}\gets h_{i,L-1},\forall i\in [n]$
        \State $h_{i,L}\gets \phi_L((W_L(0)+\Delta W_L)h_{i,L-1}),\forall i\in [n]$, threshold $\tau_{i,L}$ \Comment{Takes $o(nm^2)$ time}
        \State $D_{i,L}\gets \diag(\phi'((W_L(0)+\Delta W_L)h_{i,L-1})),\forall i\in [n]$
        
        \State $f_t\gets [a^\top h_{1,L},\ldots,a^\top h_{n,L}]^\top$ \Comment{Takes $O(nm)$ time}
        \State {\color{blue}/*Backward computation*/}
        \State $u_{i,L}\gets a^\top D_{i,L}$
        \State Form $\wt J_{L,t}$ that approximates $J_{L,t}$ using $\{u_{i,L}\}_{i=1}^n,\{v_{i,L}\}_{i=1}^n$
        \State \Comment{Takes $\wt O(mn)$ time, $\wt J_{L,t}\in \R^{n\times s}$ where $s=\wt O(n^2/\lambda_L)$}
        \State Compute $g_{L}$ that approximates $(\wt J_{L,t}\wt J_{L,t}^\top)^{-1}(f_t-y)$
        \State Form $J_{L,t}^\top g_{L}$ via low rank factorization $\sum_{i=1}^n g_{L,i}u_{i,L}v_{i,L}^\top$
        \State Implicitly update $$\Delta W_L\gets \Delta W_L-\sum_{i=1}^n g_{L,i}u_{i,L}v_{i,L}^\top$$ 
    \EndFor
\EndProcedure
\end{algorithmic}
\end{algorithm*}

Our algorithm can be summarized as follows: we use the shifted ReLU activation to ensure that with high probability, only a sublinear number of neurons in $m$ are activated. We then maintain and update the gradient through a lazy, low-rank data structure. When computing the Gauss-Newton direction, we form the Jacobian matrix implicitly and approximately, then use the inexact Jacobian to approximately form Gram and invert it.

\paragraph*{Sparsify via a Sparser Activation.} We first consider the forward computation phase, in which we need to compute the matrix-vector product 
\begin{align*}
(W_L(0)+\Delta W_L) h_{i,L-1}.
\end{align*}
As we will show later, via a carefully-designed low-rank maintenance data structure, the product $\Delta W_L h_{i,L-1}$ can be performed in $o(m^2)$ time. However, for a dense vector $h_{i,L-1}$, this does not hold for the product $W_L(0) h_{i,L-1}$; the main reason being $W_L(0)$ is initialized as a dense matrix with 0 mean Gaussian entries and does not exhibit any particular low-rank structures as $\Delta W_L$. To address this issue, we use a shifted ReLU activation to make sure that the vector $h_{i,L-1}$ is sparse. Since only $W_L$ is trained, $W_L(0)$ and $h_{i,L-1}$ never change, so the product $W_L(0)h_{i,L-1}$ can also be computed once at initialization and stored (Section~\ref{sec:plain}). Note that a coordinate of $h_{i,\ell}$ is nonzero if and only if the corresponding coordinate in $W_{\ell} h_{i,\ell-1}$ exceeds the threshold $\tau_{i,\ell}=\sqrt{2/m}\,b\,\|h_{i,\ell-1}\|_2$. Since the coordinates of $W_\ell h_{i,\ell-1}$ are Gaussian with standard deviation $\sqrt{2/m}\,\|h_{i,\ell-1}\|_2$ at initialization, each neuron fires with probability exactly $P_b=1-\Phi(b)\leq e^{-b^2/2}$, at every layer. We show that, by choosing $b$ as $\sqrt{2\gamma\log m}$ for a parameter $\gamma\in (0,1)$, the sparsity of $h_{i,\ell}$ is at most $2m^{1-\gamma}$ for all $\ell\in [L]$, with failure probability at most $nL\,e^{-mP_b/3}$ (Lemma~\ref{lem:init_sparse} and Remark~\ref{remark:k}). This bound needs no control of the norms of the hidden vectors. We also show that using our shifted ReLU activation, the induced distribution on the initial weight matrix $W(0)$ is a truncated Gaussian distribution. This does not change the structure of the convergence argument. The cost is that only about $k=m^{1-\gamma}$ neurons of a layer fire, so the random quantities of the analysis fluctuate at the scale $k^{-1/2}$ rather than $m^{-1/2}$. The width condition is therefore on $k$, namely $m^{1-\gamma}\geq \wt O(n^4L^4\lambda_L^{-4}\lambda_*^{-4})$ (Theorem~\ref{thm:conv_main}).

\paragraph*{Low-Rank Structure of the Gradient.} Consider $\frac{\partial f(W,x_i)}{\partial W_L}\in \R^{m\times m}$, it can be written as (for simplicity, we use $h_{i,0}$ to denote $x_i$):
\begin{align*}
    \frac{\partial f(W,x_i)}{\partial W_L} = D_{i,L}a \cdot h_{i,L-1}^\top.
\end{align*}
This means the gradient is essentially an outer product of two vectors, and hence has rank one. This has several interesting consequences: for over-parametrized networks, the gradient is merely of rank $n$ instead of $m$. The low-rank structure of the gradient can be further utilized, by representing and maintaining the \emph{change of the weight matrix}, $\Delta W$ in an implicit fashion so that the matrix-vector product against $\Delta W$ can be quickly computed. More specifically, suppose we are given the vectors $\{h_{i,L-1}\}_{i=1}^n$ and $\{D_{i,L}a\}_{i=1}^n$, the gradient can be compactly expressed as $\frac{\partial f}{\partial W_L} = UV^\top$.

\paragraph*{Gauss-Newton Method, Gram Regression and How to Sketch the Jacobian.}
We start by recalling the update rule of the Gauss-Newton method for our algorithm:
\begin{align*}
    W_L(t+1)\leftarrow W_L(t)-J_{L,t}^\top (J_{L,t}J_{L,t}^\top)^{-1}(f_t-y),
\end{align*}
where $J_{L,t}\in \R^{n\times m^2}$ is the Jacobian matrix. Note that we cannot even afford to form the Jacobian as a dense matrix --- as it's an $n\times m^2$ matrix, writing it down would already take $O(nm^2)$ time. On the other hand, $J_{L,t}$ has a structure that can be exploited, as each row of the matrix is in the form of $u_i \otimes v_i$ where $\otimes$ is the tensor product. While exactly forming these tensor products as dense vectors would take $O(nm^2)$ time, one can approximately compute them using tensor-based sketching techniques in time nearly linear in $m$. Given this much smaller, approximate Jacobian, we can perform subsequent operations much faster.

The next obstacle is to form the Gram matrix and compute the inversion. While the inversion can be computed relatively fast as the Gram matrix is $n\times n$, computing the Gram itself as $J_{L,t}J_{L,t}^\top$ from the dense Jacobian comes at a prohibitively high cost. Its entries factor as $\langle u_i,u_j\rangle\langle v_i,v_j\rangle$, so the exact Gram matrix can also be computed in $mn^{\omega-1+o(1)}$ time, which is subquadratic in $m$ under our width condition (Section~\ref{sec:plain}). We utilize our sketched Jacobian and instead perform the multiplication in time $\wt O(n^{\omega+1}/\lambda_L)$, which is sublinear in $m$. Inverting the Gram matrix then only takes $O(n^\omega)$ time, which is efficient. One could interpret our method as an inexact solver for Gram regression, in which the Jacobian matrices are only approximated. We show that tensor-based sketching still solves this problem quickly.

\subsection{Convergence Analysis}
\label{sec:conv_main}
To prove that our algorithm indeed converges, we need to develop a variety of new machinery beyond the standard NTK~\cite{dllwz19} and over-parametrization~\cite{als19_dnn} analysis. Specifically, we have to handle the shifted ReLU activation, which induces a different distribution than the more standard Gaussian distribution induced by the ReLU activation. Our algorithm is also based on the Gauss-Newton method, in contrast to the first-order gradient descent or stochastic gradient descent. Prior treatments of this method~\cite{cgh+19,bpsw21} focus only on analyzing two-layer over-parametrized networks, whose analysis is much simpler and standard. Since only $W_L$ is trained, our network is, as a function of $W_L$, also a two-layer network on the fixed inputs $h_{i,L-1}$, and the depth enters our analysis only through these inputs at initialization (Lemmas~\ref{lem:init_sparse}, \ref{lem:als_71} and~\ref{lem:init_min_eigen_L}), so the new machinery concerns these inputs, the shifted ReLU and the Gauss-Newton step, while the training phase follows the two-layer argument. Our strategy is to focus on the training of the last layer, a common approach for transfer learning and spurious correlations~\cite{kiw22}, and our analysis can be viewed as a theoretical explanation of last-layer training. Below, we provide an overview of our convergence analysis in phases.

\paragraph*{Initialization.} Let $W(0)$ be the random initialization. We first show that for any data point $x_i$, the initial neural network output $f(W(0),x_i) = \wt O(1)$. The analysis draws inspiration from~\cite{als19_dnn}. The general idea is that given a fixed unit length vector $x$, multiplying it with a random Gaussian matrix $W$ will ensure $\|Wx\|_2^2 \approx 2$ with high probability. Since $W$ is a random Gaussian matrix, applying shifted ReLU activation gives a random vector with a truncated Gaussian distribution conditioned on a binomial random variable indicating which neurons are activated. We will end up with $\|\phi(Wx)\|_2 \approx 1$ as well as $\phi(Wx)$ being sparse. We apply this idea layer by layer. Since the threshold of a layer is proportional to the norm of its input, the layer is positively homogeneous. Given $h_{i,\ell-1}$, the ratio $\|h_{i,\ell}\|_2^2/\|h_{i,\ell-1}\|_2^2$ therefore has the same distribution as $\|\phi(Wx)\|_2^2$ for a unit vector $x$, with mean $1$ and fluctuations of order $(mP_b)^{-1/2}$. The squared norm of $h_{i,\ell}$ is a product of $\ell$ such ratios. Controlling each of them to accuracy $\epsilon/(2L)$ shows that $\|h_{i,\ell}\|_2\in[1-\epsilon,1+\epsilon]$ for every layer $\ell$, with failure probability $O(nL)\exp(-\Omega(\epsilon^2mP_b/L^2))$ (Lemma~\ref{lem:als_71}). With the fixed threshold of the published version, deviations from norm $1$ can instead grow geometrically with the depth (Remark~\ref{remark:threshold}). We conclude the argument by exploiting the fact that $a$ is a Rademacher random vector, so with good probability the inner product $\langle a, h_{i,L}\rangle$ is at most $O(\|h_{i,L}\|_2\log n)$ in absolute value for every $i$, and hence we have $f(W(0),x_i) = \wt O(1)$.

Furthermore, we show that the Gram matrix for the multi-layer over-parametrized neural network, defined as $J_{L,0} J_{L,0}^\top$, has a nontrivial minimum eigenvalue after initialization. In particular, we adapt the neural tangent kernel (NTK) for multi-layer neural networks defined by \cite{dllwz19} into our setting by analyzing the corresponding Gaussian process with shifted ReLU activation function. Then, we can prove that with high probability, the smallest eigenvalue of the initial Gram matrix is lower bounded by a constant fraction of the smallest eigenvalue of the neural tangent kernel matrix. Only about $k=m\,e^{-b^2/2}$ neurons of a layer fire, so the inner products $h_{i,\ell}^\top h_{j,\ell}$ fluctuate around their kernel values at the scale $k^{-1/2}$ rather than $m^{-1/2}$. We control them with a Bernstein bound that uses this sparsity. By homogeneity, the correlations $\langle h_{i,\ell},h_{j,\ell}\rangle/(\|h_{i,\ell}\|_2\|h_{j,\ell}\|_2)$ evolve independently of the norms, so we propagate the errors through the layers in terms of these correlations. As a function of the correlation, the kernel map $\psi$ of one layer is a power series with nonnegative coefficients that fixes the correlation $1$. Hence it is convex on $[0,1]$, and its slope at $s$ is at most $(1-\psi(s))/(1-s)$. These ratios telescope along the layers, so after $\ell$ layers the accumulated error is at most $O(\ell/\lambda_*)$ times the error that one layer adds (Lemmas~\ref{lem:dual}, \ref{lem:init_min_eigen_H} and~\ref{lem:init_min_eigen_L}).

\paragraph*{Small Perturbation.} The next step is to show that if the last weight matrix $W_L$ undergoes a small perturbation from initialization (in terms of the norms of its rows), then the corresponding Jacobian matrix has not changed too much (Lemma~\ref{lem:perturb}). Since the inputs $h_{i,L-1}$ of the last layer do not change, the Jacobian changes only through the neurons of the last layer whose activation changes, and we count the neurons whose activation can change. These are the neurons whose initial pre-activation lies within the perturbation radius of the threshold. Bernstein's inequality bounds their number. This shows that the change of the Jacobian has a relatively small spectral norm and Frobenius norm.

\paragraph*{Connect Everything via a Double Induction.} We use a double induction argument, where we assume the perturbation of the weight matrix is small and the gap between $f_t$ and $y$ is at most $1/3$ of the gap between $f_{t-1}$ and $y$. By carefully bounding various terms and exploiting the fact that the Jacobian matrix \emph{always} has a relatively small spectral norm of $\wt O(\sqrt{n})$, we first show that the weights are not moving too far from the initialization, then use this fact to derive a final convergence bound for $\|f_t-y\|_2$.

\section{Simpler Algorithms for Last-Layer Training}
\label{sec:plain}

Algorithm~\ref{alg:complete} (Appendix~\ref{sec:alg_app}), the formal version of Algorithm~\ref{alg:main}, trains only the last weight matrix $W_L$, so the inputs $h_{i,L-1}=h_{i,L-1}(0)$ of the last layer never change (Remark~\ref{remark:threshold}). For $b=\sqrt{2\gamma\log m}$ and $k=m^{1-\gamma}$, each of them has at most $2k$ nonzero entries with high probability (Lemma~\ref{lem:init_sparse} and Remark~\ref{remark:k}). The algorithm keeps $W_L$ in the low-rank data structure of Section~\ref{sec:low_rank_app}, whose procedure \textsc{Query} returns the product of $W_L$ with a given vector, and whose procedure \textsc{Update} adds a matrix of rank at most $n$ to $W_L$. A step of the algorithm reads $W_L$ only through the $n$ vectors $z_{i,L}=W_Lh_{i,L-1}$, which it computes with \textsc{Query} and calls $g_{i,L}$. From them it computes $h_{i,L}$, $D_{i,L}$, $f_t$ and $u_{i,L}=D_{i,L}a$. It then solves the Gram regression with the vectors $u_{i,L}$ and $v_{i,L}=h_{i,L-1}$, which returns a vector $g_L\in\R^n$, and it subtracts $\sum_{i=1}^n g_{L,i}u_{i,L}h_{i,L-1}^\top$ from $W_L$. In this section, we show that two simple implementations of these steps converge at the rate of Theorem~\ref{thm:conv_main}. The theorem applies to the first as stated, and its proof applies to the second with the exact solve in place of the guarantees of the solver (Lemma~\ref{lem:exact_solve}). The first stores $W_L$ as an explicit matrix and attains the bound of Theorem~\ref{thm:alg_main} at every step. The second never forms $W_L$ during training, solves the Gram regression exactly and takes $O(mn^{\omega-1+o(1)})$ time per step. A refinement of the second updates only the rows of the neurons that fire on some data point and takes $O(n^2k)$ time per step (Theorem~\ref{thm:plain_incremental_informal}).

We use the following notation. Let $H:=[h_{1,L-1},\ldots,h_{n,L-1}]\in\R^{m\times n}$ and $K:=H^\top H\in\R^{n\times n}$. Both stay fixed during training. At step $t$, let $W_L(t)$ be the weight matrix of the last layer and let $Z(t):=W_L(t)H\in\R^{m\times n}$, whose $i$-th column is $z_{i,L}$. Let $U_t:=[u_{1,L},\ldots,u_{n,L}]\in\R^{m\times n}$ and $\wt U_t:=-U_t\diag(g_L)\in\R^{m\times n}$, whose $i$-th column is $-g_{L,i}u_{i,L}$. The update of Algorithm~\ref{alg:complete} is then
\begin{align}\label{eq:plain_update}
    W_L(t+1)=W_L(t)+\wt U_tH^\top.
\end{align}
As in the rest of this paper, we count arithmetic operations on real numbers, carried out exactly. For positive integers $p,q,r$, we write $\Tmat(p,q,r)$ for the time to multiply a $p\times q$ matrix by a $q\times r$ matrix. If $m\geq n$, then we can split the dimension of size $m$ into $\lceil m/n\rceil$ blocks of size at most $n$ and multiply block by block. For $\Tmat(n,m,n)$, we also add up the $\lceil m/n\rceil$ products of size $n\times n$, in $O(mn)$ time. Since $\omega\geq 2$, this gives
\begin{align}\label{eq:plain_tmat}
    \Tmat(m,n,n)=O(mn^{\omega-1+o(1)})\quad\text{and}\quad\Tmat(n,m,n)=O(mn^{\omega-1+o(1)}).
\end{align}

\subsection{Explicit Storage of the Last Layer}

Our first implementation keeps Algorithm~\ref{alg:complete} and replaces only the data structure by an explicit matrix.

\begin{theorem}[Explicit storage of the last layer]\label{thm:plain_dense}
Let $b=\sqrt{2\gamma\log m}$ for a constant $\gamma\in(0,1)$, let $k=m^{1-\gamma}$, and suppose that the width $m$ satisfies the condition of Theorem~\ref{thm:conv_main} for this $\gamma$. Consider the implementation of Algorithm~\ref{alg:complete} that stores $W_L$ as an explicit $m\times m$ matrix and carries out each call to \textsc{Query} and \textsc{Update} directly on this matrix. Then the following hold.
\begin{itemize}
    \item (Convergence) The guarantee of Theorem~\ref{thm:conv_main} holds for this implementation.
    \item (Cost per step) Every step takes $O(m\sum_{i=1}^n\|h_{i,L-1}\|_0+nm)$ time, plus the time $\wt O(nm+\lambda_L^{-1}n^{\omega+1})$ of the solver. With probability at least $1-1/\poly(n)$ over the initialization, every step takes $O(nmk)=O(nm^{2-\gamma})$ time in total. This bound holds at every step, without amortization.
    \item (Comparison) If $\gamma\geq a$ for $a=\min\{\alpha,1/2\}$, as in Theorem~\ref{thm:alg_main}, then, on the same event, every step takes $O(nm^{2-a})$ time, which is at most $O(m^{(9-5a)/4})$. So this implementation attains both bounds of Theorem~\ref{thm:alg_main} at every step, without the data structure of Section~\ref{sec:low_rank_app}.
\end{itemize}
\end{theorem}

\begin{proof}
Algorithm~\ref{alg:complete} reads $W_L$ only in the calls $\textsc{Query}(L,h_{i,L-1})$, which return $z_{i,L}=W_Lh_{i,L-1}$, and it changes $W_L$ only in the call to \textsc{Update}, which adds $\sum_{i=1}^n(-g_{L,i}u_{i,L})h_{i,L-1}^\top$. On the explicit matrix, we compute
\begin{align*}
    z_{i,L}=\sum_{j\in\supp(h_{i,L-1})}(h_{i,L-1})_j\,W_L^{(j)},
\end{align*}
where $W_L^{(j)}$ denotes the $j$-th column of $W_L$. This takes $O(m\|h_{i,L-1}\|_0)$ time. We apply the update as $n$ rank-one updates. Forming the vector $-g_{L,i}u_{i,L}$ takes $O(m)$ time. The update $W_L\gets W_L-g_{L,i}u_{i,L}h_{i,L-1}^\top$ changes only the entries $(r,j)$ with $r\in\supp(u_{i,L})$ and $j\in\supp(h_{i,L-1})$, so it takes $O(m\|h_{i,L-1}\|_0)$ time. Both operations are exact, so after every step the explicit matrix equals the matrix that the data structure represents, and the products $z_{i,L}$ are the same. Hence the two implementations compute the same $h_{i,L}$, $D_{i,L}$, $f_t$ and $u_{i,L}$, the same input of the solver and the same update, given the same random choices of the solver. The proof of Theorem~\ref{thm:conv_app}, from which Theorem~\ref{thm:conv_main} follows, uses the algorithm only through these quantities. So it applies without change, which proves the first claim.

The other operations of a step compute $h_{i,L}$, $D_{i,L}$, $f_t$ and $u_{i,L}$ in $O(nm)$ time, as in the proof of Theorem~\ref{thm:alg_app}. As in that proof, the solver of Theorem~\ref{thm:fast_kronecker_regression}, run with precision $\epsilon_0=\sqrt{\lambda_L/n}/30$ and failure probability $\delta_0=e^{-\log^2n}$ (Definition~\ref{def:R}), takes $\wt O(nm+\lambda_L^{-1}n^{\omega+1})$ time. This proves the first bound of the second claim. Every step performs the same operations, so the bound holds at every step.

The condition of Theorem~\ref{thm:conv_main} implies the width condition of Definition~\ref{def:m} (see the paragraph after Definition~\ref{def:m}). Under the width condition, Lemma~\ref{lem:init_sparse} and Remark~\ref{remark:k} give, with probability at least $1-1/\poly(n)$ over the initialization, $\|h_{i,L-1}\|_0\leq 2m^{1-\gamma}=2k$ for all $i\in[n]$. Then the products and the updates take $O(m\sum_{i=1}^n\|h_{i,L-1}\|_0)\leq O(nmk)$ time, and $nm\leq nmk$ as $k\geq 1$. Since $C_W\geq 1$, $1+b\geq 1$, $L\geq 1$, $\lambda_*\leq 1$, $\lambda_L\leq 1$ and $\log(mnL)\geq 1$, the width condition gives $k=m^{1-\gamma}\geq n^4\lambda_L^{-4}$. So $n\leq m^{(1-\gamma)/4}\leq m^{1/4}$ and $\lambda_L^{-1}\leq m^{(1-\gamma)/4}/n$, which gives $\lambda_L^{-1}n^{\omega+1}\leq m^{(1-\gamma)/4}n^{\omega}\leq m^{(1-\gamma)(1+\omega)/4}\leq m$, as $\omega\leq 3$. So the solver takes $\wt O(nm)$ time. The factor hidden in $\wt O$ is polylogarithmic in $nm$, since $\epsilon_0^{-1}=30\sqrt{n/\lambda_L}$ with $\lambda_L^{-1}\leq m$ and since $\log(1/\delta_0)=\log^2n$. Since $\gamma<1$ is a constant, every factor polylogarithmic in $nm$ is $O(m^{1-\gamma})$, so $\wt O(nm)\leq O(nm^{2-\gamma})$. Hence every step takes $O(nmk)=O(nm^{2-\gamma})$ time, which proves the second claim. If $\gamma\geq a$, then $nm^{2-\gamma}\leq nm^{2-a}$, and $n\leq m^{(1-\gamma)/4}\leq m^{(1-a)/4}$ gives $nm^{2-a}\leq m^{(9-5a)/4}$. This proves the third claim.
\end{proof}

The products $z_{i,L}$ dominate this cost, and the updates change far fewer entries of $W_L$. On the events (E1) to (E4) of Section~\ref{sec:analysis}, the proof of Theorem~\ref{thm:conv_app} gives Induction hypothesis~1 at every step. The argument for Theorem~\ref{thm:plain_incremental}(ii) uses only this hypothesis and the events (E1) and (E3), so it applies here and gives $|\supp(u_{i,L})|\leq 2k+N\leq 3k$, with $N\leq k/n$ as in that theorem. So on these events, a step changes only $O(nk^2)$ entries of $W_L$.

The data structure of Section~\ref{sec:low_rank_app} does not lower this cost. Its procedure \textsc{Query} multiplies its stored dense matrix (the member $W_L$ of Algorithm~\ref{alg:low_rank_ds}) by $h_{i,L-1}$ in $O(m\|h_{i,L-1}\|_0)$ time, just as above (Lemma~\ref{lem:low_rank_maintain}). On top of this, it multiplies the low-rank part by $h_{i,L-1}$, and \textsc{Update} pays the amortized cost $nm^{2-a+o(1)}$ of the rebuilds. So the bound of Theorem~\ref{thm:alg_main} is amortized over the rebuilds, while the bound of Theorem~\ref{thm:plain_dense} holds at every step.

\subsection{Maintaining \texorpdfstring{$W_LH$}{W\_L H}}

Our second implementation never forms $W_L$ during training. It keeps the matrix $Z=W_LH$ of the $n$ products, and the matrix $\Gamma\in\R^{m\times n}$ with $W_L=W_L(0)+\Gamma H^\top$. By~\eqref{eq:plain_update}, the products change by $\wt U_tK$ in each step, with the fixed matrix $K=H^\top H$. The Gram matrix also factors through $K$, so we compute it exactly and solve the Gram regression exactly. Here $\odot$ denotes the entrywise product of two matrices.

\begin{algorithm}[!ht]\caption{Training the last layer by maintaining $W_LH$.}
\label{alg:plain_gram}
\begin{algorithmic}[1]
\Procedure{TrainLastLayer}{$X\in \R^{d\times n}, y\in \R^n, T$} \Comment{Theorem~\ref{thm:plain_gram}}
    \State {\color{blue}/*Initialization*/}
    \State Initialize $W_\ell(0)$, $\forall \ell \in [L]$
    \State Compute $h_{i,\ell}$ for $\ell\in [L-1]$, with thresholds $\tau_{i,\ell}=\sqrt{2/m}\,b\,\|h_{i,\ell-1}\|_2$ \Comment{$O(nm^2L)$ time}
    \State $H\gets[h_{1,L-1},\ldots,h_{n,L-1}]$ and $\tau_{i,L}\gets\sqrt{2/m}\,b\,\|h_{i,L-1}\|_2$, $\forall i\in [n]$
    \State $K\gets H^\top H$, $Z\gets W_L(0)H$, $\Gamma\gets 0\in\R^{m\times n}$ \Comment{Takes $O(nm^2)$ time}
    \For{$t= 0 \to T-1$}
        \State $z_{i,L}\gets$ the $i$-th column of $Z$, $\forall i\in [n]$
        \State $h_{i,L}\gets \phi_L(z_{i,L})$ with threshold $\tau_{i,L}$, and $D_{i,L}\gets \diag(\phi'(z_{i,L}))$, $\forall i\in [n]$
        \State $f_t\gets [a^\top h_{1,L},\ldots,a^\top h_{n,L}]^\top$
        \State $U\gets[u_{1,L},\ldots,u_{n,L}]$ with $u_{i,L}=D_{i,L}a$, $\forall i\in [n]$ \Comment{Takes $O(nm)$ time}
        \State $Q\gets (U^\top U)\odot K$ \Comment{Exact Gram matrix, $\Tmat(n,m,n)+O(n^2)$ time}
        \State $g_L\gets Q^{-1}(f_t-y)$, and stop if $Q$ is singular \Comment{Gaussian elimination, $O(n^3)$ time}
        \State $\wt U\gets -U\diag(g_L)$
        \State $Z\gets Z+\wt UK$ and $\Gamma\gets \Gamma+\wt U$ \Comment{Takes $\Tmat(m,n,n)+O(nm)$ time}
    \EndFor
    \State \Return $W_L(T)=W_L(0)+\Gamma H^\top$ \Comment{Formed once}
\EndProcedure
\end{algorithmic}
\end{algorithm}

\begin{theorem}[Maintaining $W_LH$]\label{thm:plain_gram}
Let $b=\sqrt{2\gamma\log m}$ for a constant $\gamma\in(0,1)$, and suppose that the width $m$ satisfies the condition of Theorem~\ref{thm:conv_main} for this $\gamma$. Then Algorithm~\ref{alg:plain_gram} has the following guarantees.
\begin{itemize}
    \item (Convergence) With probability at least $1-e^{-\Omega(\log^2n)}$ over the initialization, for every number of steps $T\geq 1$, we have $\|f_{t+1}-y\|_2\leq\frac{1}{3}\|f_t-y\|_2$ for all $t\in\{0,\dots,T-1\}$.
    \item (Cost per step) Every step takes $O(\Tmat(m,n,n)+\Tmat(n,m,n)+n^3)\leq O(mn^{\omega-1+o(1)})$ time, without amortization. Since the width condition gives $n\leq m^{1/4}$, this is at most $m^{1+(\omega-1)/4+o(1)}$, and it is $O(m^{3/2})$ with schoolbook matrix multiplication.
    \item (Initialization and output) The initialization takes $O(m^2nL)$ time, which is the time ${\cal T}_{\mathrm{init}}$ of Theorem~\ref{thm:alg_app}. Forming the trained matrix $W_L(T)$ after the last step takes $O(nm^2)$ time, once. With probability at least $1-1/\poly(n)$ over the initialization, it takes only $O(nm^{2-\gamma})$ time.
\end{itemize}
\end{theorem}

\begin{proof}
We compare Algorithm~\ref{alg:plain_gram} with Algorithm~\ref{alg:complete} in which the call to \textsc{FastTensorRegression} is replaced by the exact solution of the same regression problem, and which also stops if the Gram matrix is singular. Let $W_L(t)$ be the weight matrix of the latter after $t$ steps, and let $Z_t$ and $\Gamma_t$ be the matrices $Z$ and $\Gamma$ of Algorithm~\ref{alg:plain_gram} at the start of step $t$. We claim that
\begin{align*}
    Z_t=W_L(t)H\quad\text{and}\quad W_L(t)=W_L(0)+\Gamma_tH^\top.
\end{align*}
Both hold for $t=0$. Suppose that they hold at step $t$. Then the columns of $Z_t$ are the products $z_{i,L}=W_L(t)h_{i,L-1}$, which Algorithm~\ref{alg:complete} computes as $g_{i,L}$, so the two algorithms compute the same $h_{i,L}$, $D_{i,L}$, $f_t$ and $u_{i,L}$. Let $J'_{L,t}$ be the Jacobian with respect to $W_L$, whose $i$-th row is $\vect(u_{i,L}h_{i,L-1}^\top)^\top$, as in Section~\ref{sec:analysis}. For all $i,j\in[n]$,
\begin{align*}
    (J'_{L,t}J'^\top_{L,t})_{i,j}=\tr\big[h_{i,L-1}u_{i,L}^\top u_{j,L}h_{j,L-1}^\top\big]=(u_{i,L}^\top u_{j,L})(h_{i,L-1}^\top h_{j,L-1}),
\end{align*}
which is the case $\ell=L$ of Fact~\ref{fac:multilayer_gram}. So $Q=(U_t^\top U_t)\odot K=J'_{L,t}J'^\top_{L,t}$, and the two algorithms stop at the same steps. If they do not stop at step $t$, then $g_L=Q^{-1}(f_t-y)$ is the exact solution of the regression problem that Algorithm~\ref{alg:complete} passes to its solver. The update of Algorithm~\ref{alg:complete} is then~\eqref{eq:plain_update} with $\wt U_t=-U_t\diag(g_L)$, which is the matrix $\wt U$ of Algorithm~\ref{alg:plain_gram}. Hence
\begin{align*}
    W_L(t+1)H= & ~ W_L(t)H+\wt U_tH^\top H=Z_t+\wt U_tK=Z_{t+1},\\
    W_L(t+1)= & ~ W_L(0)+(\Gamma_t+\wt U_t)H^\top=W_L(0)+\Gamma_{t+1}H^\top.
\end{align*}
This proves the claim. In particular, the two algorithms compute the same $f_t$ at every step, and the returned matrix is $W_L(T)$.

\textbf{Convergence.} The condition of Theorem~\ref{thm:conv_main} implies the width condition of Definition~\ref{def:m} (see the paragraph after Definition~\ref{def:m}). By the claim, as long as Algorithm~\ref{alg:plain_gram} does not stop, its iterates are the Gauss-Newton iterates~\eqref{eq:exact_step} with the exact Gram solve, with $Q=J'_{L,t}J'^\top_{L,t}$ and $g_L=g'(t)$. On the events (E1) to (E3) of Section~\ref{sec:analysis}, Lemma~\ref{lem:exact_solve} shows that $Q$ is invertible at every step, so the algorithm does not stop, and that $\|f_{t+1}-y\|_2\leq\frac{1}{3}\|f_t-y\|_2$ for every $t\geq 0$. These events concern only the initialization, and each of them fails with probability at most $e^{-\Omega(\log^2n)}$. None of them depends on $T$, so the conclusion holds for every $T$ on the same event.

\textbf{Cost per step.} In a step, reading the columns of $Z$ takes $O(nm)$ time, and computing $h_{i,L}$, $D_{i,L}$, $f_t$, $U$ and $\wt U$ takes $O(nm)$ time. Forming $U^\top U$ takes $\Tmat(n,m,n)$ time, and the entrywise product with $K$ takes $O(n^2)$ time. Gaussian elimination decides whether $Q$ is singular and, if it is not, solves the linear system in $O(n^3)$ time. Forming $\wt UK$ takes $\Tmat(m,n,n)$ time, and the two additions take $O(nm)$ time. Every step performs the same operations, so the total holds at every step. Since $\Tmat(m,n,n)\geq nm\geq n^2$, the terms $O(nm)$ and $O(n^2)$ are absorbed. The width condition gives $m\geq n^4$, so $n^3\leq mn\leq mn^{\omega-1}$, and by~\eqref{eq:plain_tmat} the total is $O(mn^{\omega-1+o(1)})$. As in the proof of Theorem~\ref{thm:plain_dense}, the width condition gives $n\leq m^{(1-\gamma)/4}\leq m^{1/4}$, so $mn^{\omega-1+o(1)}\leq m^{1+(1-\gamma)(\omega-1)/4+o(1)}\leq m^{1+(\omega-1)/4+o(1)}$. If $\omega<2.258$, then two $n\times n$ matrices can be multiplied in $O(n^{2.258})$ time, so the block products give $\Tmat(m,n,n)$ and $\Tmat(n,m,n)$ in $O(mn^{1.258})$ time. Since also $n^3\leq mn$, a step takes $O(mn^{1.258})\leq O(m^{1+1.258/4})=O(m^{1.3145})$ time. In the same way, $\omega<2.3714$ gives $O(m^{1+1.3714/4})\leq O(m^{1.343})$. With schoolbook multiplication, $\Tmat(m,n,n)$ and $\Tmat(n,m,n)$ are $O(mn^2)$, so every step takes $O(mn^2)\leq O(m^{3/2})$ time.

\textbf{Initialization and output.} Initializing the weights and computing $h_{i,\ell}$ and $\tau_{i,\ell}$ take $O(m^2nL)$ time, as in the proof of Theorem~\ref{thm:alg_app}. Computing $K=H^\top H$ takes $O(n^2m)$ time and computing $Z=W_L(0)H$ takes $O(nm^2)$ time. Both are $O(nm^2)$, so the initialization takes $O(m^2nL)$ time. After the last step, forming $\Gamma H^\top$ takes $O(nm^2)$ time and adding $W_L(0)$ takes $O(m^2)$ time. This happens once. With probability at least $1-1/\poly(n)$, each $h_{i,L-1}$ has at most $2m^{1-\gamma}$ nonzero entries (Lemma~\ref{lem:init_sparse} and Remark~\ref{remark:k}), and then computing $Z=W_L(0)H$ and adding $\Gamma H^\top$ in place to the stored $W_L(0)$ both take only $O(nm^{2-\gamma})$ time, as in the proof of Theorem~\ref{thm:plain_incremental}.
\end{proof}

The current best bounds are $\omega<2.258$ and $\alpha>0.465$ over every field of characteristic zero~\cite[Theorem~1.1 and Corollaries~14.1--14.2]{OpenAIMM26}. With $\omega<2.258$, the proof of Theorem~\ref{thm:plain_gram} shows that every step takes $O(m^{1.3145})$ time. The refereed bound $\omega<2.3714$~\cite{adw25} gives $O(m^{1.343})$, and schoolbook matrix multiplication gives $O(m^{1.5})$. These three bounds use only $n\leq m^{1/4}$, which holds for every $\gamma\in(0,1)$. They are polynomially smaller than the bounds $m^{(9-5a)/4+o(1)}=O(m^{1.66875})$ and $m^{2.25-a+o(1)}=O(m^{1.785})$ that Theorem~\ref{thm:alg_main} gives with $a=\min\{\alpha,1/2\}>0.465$. Since $(9-5a)/4\geq 1.625$ for every $a\leq 1/2$, already the schoolbook bound $O(m^{1.5})$ is polynomially smaller, so this comparison does not depend on the recent bounds on $\omega$ and $\alpha$. Exponents of $m$ alone do not give a fair comparison, since a larger $\gamma$ requires a wider network for the same $n$ and lowers every exponent of $m$. At the same $n$, $m$ and $\gamma\geq a$, the bound of Theorem~\ref{thm:plain_gram} is smaller than the bound $nm^{2-a+o(1)}$ of Theorem~\ref{thm:alg_main} by a factor of at least $m^{1/4-o(1)}$, because
\begin{align*}
    \frac{mn^{\omega-1}}{nm^{2-a}}=n^{\omega-2}m^{a-1}\leq m^{1/4}\cdot m^{-1/2}=m^{-1/4},
\end{align*}
by $n\leq m^{1/4}$, $\omega\leq 3$ and $a\leq 1/2$. The bound of Theorem~\ref{thm:plain_gram} depends neither on the sparsity $k$ nor on the dual exponent $\alpha$. The exact Gram matrix and the exact solve fit within the bound $O(mn^{\omega-1+o(1)})$ of the update of $Z$, so the sketch of Section~\ref{sec:reg_app} is not needed in this setting.

The implementation of Theorem~\ref{thm:plain_gram} updates all $m$ rows of $Z=W_LH$ in every step. A row of $\wt U_t$ is nonzero only if its neuron fires on some data point, so only these rows of $Z$ change, and a neuron that fires on no data point never fires again. In Section~\ref{sec:plain_incremental}, we update only these rows and prove the following bound.

\begin{theorem}[Incremental implementation, informal version of Theorem~\ref{thm:plain_incremental}]\label{thm:plain_incremental_informal}
Let $b=\sqrt{2\gamma\log m}$ for a constant $\gamma\in(0,1)$, let $k=m^{1-\gamma}$, and suppose that the width $m$ satisfies the condition of Theorem~\ref{thm:conv_main} for this $\gamma$. Then, with probability at least $1-e^{-\Omega(\log^2n)}$ over the initialization, the following holds for every number of steps $T\geq 1$. Algorithm~\ref{alg:plain_incremental} computes the Gauss-Newton iterates with the exact Gram solve, which by the proof of Theorem~\ref{thm:plain_gram} are also the iterates of Algorithm~\ref{alg:plain_gram}, so $\|f_{t+1}-y\|_2\leq\frac{1}{3}\|f_t-y\|_2$ for all $t\in\{0,\dots,T-1\}$, and it returns $W_L(T)$. Every step takes $O(n^2k)$ time, without amortization, and $n^2k\leq m^{3(1-\gamma)/2}$. This is $O(m^{3/4})$ for $\gamma=1/2$ and $o(m)$ for every constant $\gamma>1/3$. The initialization takes $O({\cal T}_{\mathrm{init}})$ time, and forming $W_L(T)$ after the last step takes $O(nmk)$ time.
\end{theorem}

\paragraph*{Relation to the Published Algorithm.} Theorems~\ref{thm:plain_dense}, \ref{thm:plain_gram} and~\ref{thm:plain_incremental_informal} show that last-layer training is easier than the published version of this paper (ITCS 2024) presents it. We did not realize this when we wrote that version. In that version, we used the shifted ReLU to make the inputs $h_{i,L-1}$ of the last layer sparse, and our algorithm computed them once at initialization and read $W_L$ only through the products $W_Lh_{i,L-1}$. What we missed is that fixed inputs make the problem easier. Since $K=H^\top H$ is fixed, the products $W_Lh_{i,L-1}$ and the Gram matrix can be kept exactly through $K$, whatever the sparsity of the inputs, and for suitable thresholds the sparsity of the activations of the last layer lowers the cost further. We treated the update of $W_L$ as an $m\times m$ matrix that costs $\Theta(m^2)$ time to apply, and the Gram matrix as too expensive to compute. We designed the low-rank data structure and the sketch to avoid these costs, and our running time had a term in the number of activation changes, which the fixed inputs make unnecessary. In the last-layer setting, each product $W_Lh_{i,L-1}$ reads only the $O(k)$ columns of $W_L$ in the support of $h_{i,L-1}$, each step changes only $O(nk^2)$ entries of $W_L$ with high probability, and the Gram matrix is, up to the factor $2/m$, the entrywise product of $U_t^\top U_t$ with $K$. So storing $W_L$ explicitly already attains the bound of Theorem~\ref{thm:alg_main} at every step, and maintaining $W_LH$ with an exact solve takes $O(mn^{\omega-1+o(1)})$ time per step. Updating only the rows of $W_LH$ for the neurons that fire on some data point takes $O(n^2k)$ time per step. This is sublinear in $m$ for every constant $\gamma>1/3$, which our width condition allows only at widths $m>n^6$. In this paper, we keep Theorem~\ref{thm:alg_main}, the result of the published version in the form adapted here, with the data structure of Section~\ref{sec:low_rank_app} and the sketch of Section~\ref{sec:reg_app} as its implementation. The convergence analysis of Section~\ref{sec:analysis} is unaffected, and it applies to all of these implementations. For those that solve the Gram regression exactly, the exact solve takes the place of the guarantees of the solver (Lemma~\ref{lem:exact_solve}).

\section{Discussion and Future Directions}
\label{sec:discuss_main}
In this work, we propose and analyze a variant of the Gauss-Newton method to train the last layer of multi-layer over-parametrized neural networks. Our algorithm achieves a linear convergence rate in terms of training loss and an amortized cost per training iteration below the $O(nm^2)$ time per iteration of a dense implementation. The threshold of each layer is proportional to the norm of the layer input, which keeps every layer positively homogeneous. As a result, our convergence guarantee holds with a width polynomial in $n$, $L$, $1/\lambda_L$ and $1/\lambda_*$. From an analytical perspective, we greatly extend the analysis of~\cite{als19_dnn} to our method, coupled with the use of the equivalence between multi-layer over-parametrized networks and neural tangent kernels~\cite{dllwz19}. From an algorithmic perspective, our main algorithm (Algorithm~\ref{alg:main}) has a subquadratic amortized cost per iteration, below the $O(m^2)$ time of a dense implementation. The simpler algorithms of Section~\ref{sec:plain} reach this cost without the data structure, and lower it further. Our algorithm combines various techniques, such as training with the Gram matrix, solving the Gram regression with a sketched Jacobian, fast tensor computation and dimensionality reduction, and low-rank decomposition of weight updates. For gradient descent on the last layer with a fixed step size, storing $W_L$ explicitly already gives $O(nmk)$ time per step with high probability, as in the proof of Theorem~\ref{thm:plain_dense}. Since the training error decreases by a factor of $3$ in every iteration and is at most $\sqrt n\log^{O(1)}(nm)$ at initialization (Definition~\ref{def:R}), $O(\log(nm/\epsilon))$ iterations reach a training error of at most $\epsilon$. For $\epsilon\geq 1/\poly(m)$, these iterations of the algorithm of Theorem~\ref{thm:plain_gram} take $O(m^{1.3145}\log m)=o(m^2)$ time together, which is less than the $\Theta(m^2L)$ time to sample the initial weights. So the cost per iteration matters most when the same initialized network is trained for many iterations. Our algorithm follows the use of data structures to speed up iterative methods, a popular trend in recent years~\cite{cls19,lsz19,jswz20,hjs+21,jnw22,sswz22,dswz22}. Though our work is mainly theoretical, our techniques draw inspiration from practice, such as sparse activations~\cite{cgh+19,clp+21} and last layer training~\cite{kiw22}. Last-layer training resembles the fine-tuning of large pre-trained language models, in which most of the model stays frozen. Our guarantees, however, assume that the frozen layers are random, as at initialization. For a pre-trained model, only the cost per step of Theorem~\ref{thm:plain_gram} carries over, since it uses neither the sparsity nor the randomness of the frozen layers. Below, we pose several open problems related to our result.

\paragraph*{Nearly-Linear Time Last-Layer Training.} The first question one might wonder is whether achieving a per iteration cost of $\wt O(m)$ is possible. In particular, can this runtime be achieved under the current best width of multi-layer over-parametrized networks ($m \geq n^4$)? The limiting terms in the bound of Theorem~\ref{thm:alg_main} are the amortized cost $nm^{2-a+o(1)}$ of rebuilding $W_L$ in the low-rank data structure, with $a=\min\{\alpha,1/2\}$, and the term $nmk$ of multiplying the dense matrix $W_L$ by the sparse vectors $h_{i,L-1}$. The number of activation changes during training does not enter this cost, since the queried vectors $h_{i,L-1}$ never change. Neither of the two terms is needed. Maintaining the products $W_Lh_{i,L-1}$ directly costs $mn^{\omega-1+o(1)}$ time per iteration, which is $O(m^{1.3145})$ under our width condition, since $\omega<2.258$ over every field of characteristic zero~\cite[Theorem~1.1 and Corollaries~14.1--14.2]{OpenAIMM26} (Section~\ref{sec:plain}). Updating only the rows of $W_LH$ for the neurons that fire on some data point costs $O(n^2k)$ time per iteration, and $n^2k\leq m^{3(1-\gamma)/2}$ (Theorem~\ref{thm:plain_incremental}). So a cost per iteration of $o(m)$ is possible for every constant $\gamma>1/3$. Since our width condition gives $m^{1-\gamma}\geq n^4$, such a $\gamma$ needs a width $m>n^6$. We do not know whether a cost of $\wt O(m)$ per iteration is possible at widths closer to $n^4$.

\paragraph*{Faster Initialization.} One direction is to design an initialization setup such that while we still have enough randomness to obtain provable guarantees, the matrix-vector product with the initial weight matrix can be performed faster than $O(m^2)$ by sparsifying the Gaussian matrix as in~\cite{mjmm21} or imposing extra structural assumption such as using circulant Gaussian \cite{rrt12,nn13,kmr14}. In last-layer training, this would lower only the initialization time ${\cal T}_{\mathrm{init}}$, since the products with the initial weights are computed once, at initialization (Theorem~\ref{thm:plain_gram}). Since the hidden vectors are sparse with high probability, these products already take $O(mk)$ time each on the layers above the first, so apart from the first-layer products, which take $O(nmd)$ time, sampling the Gaussian entries of the initial weights is the only part of the initialization whose cost has a factor $m^2$.

\paragraph*{Depth Dependence of the Width.} The width condition of Theorem~\ref{thm:conv_main}, $m^{1-\gamma}\geq\wt O(n^4L^4\lambda_L^{-4}\lambda_*^{-4})$, is polynomial in the depth. This polynomial dependence relies on the norm-scaled threshold $\sqrt{2/m}\,b\,\|h_{i,\ell-1}\|_2$, which departs from the published version of this work. With the fixed threshold $\sqrt{2/m}\,b$, norm $1$ is an unstable point of every layer, and our analysis of that network needs a width with a factor exponential in the depth (Remark~\ref{remark:threshold}). Can one keep the fixed threshold and still use a width polynomial in $L$? A second question concerns the factor $\lambda_*^{-4}$. It comes from pairs of data points whose kernel correlation at some hidden layer is close to $1$. The kernel correlations at the hidden layers lie in $[0,1]$, since the activation is nonnegative. The kernel map of a layer pushes the correlations of such pairs away from $1$. Near correlation $1$ its slope tends to infinity, because the thresholded activation $\phi$ jumps at its threshold. So the map amplifies the errors of such pairs. Our comparison of the hidden vectors with the kernels bounds this amplification by the factor $1/(1-|\rho|)\leq 1/\lambda_*$, where $\rho$ is the kernel correlation of the pair (Lemma~\ref{lem:init_min_eigen_H}). Is this factor needed?

\paragraph*{Extension to Other Activations.} In this paper, we consider the shifted ReLU activation and design our algorithm and analysis around its properties. Is it possible to generalize our algorithm and analysis to various other activation functions, such as sigmoid, tanh, leaky ReLU, or GeLU? If one chooses a smooth activation, can we get a better result in terms of convergence rate? Can we leverage this structure to design faster algorithms?

\paragraph*{Network Architecture Beyond Feedforward.} Finally, the network architecture considered in this paper is the standard feedforward network. Is it possible to extend our analysis and algorithm to other architectures, such as recurrent neural networks (RNN) or transformers? For RNN, the weight matrices for each layer are the same. Hence, it is trickier to analyze the training dynamics on such networks. Though the convergence of the first-order method on over-parametrized multi-layer RNN has been established, it is unclear whether such analysis can be extended to our method. More generally, can we improve the efficiency of the training process of neural networks even beyond the NTK regime, e.g., in the feature learning regime? We believe new techniques are needed to achieve theoretical time complexity improvements as well as the convergence guarantee.

\section*{Acknowledgements}

We would like to thank Yin Tat Lee for discussing the motivation of this problem, Jan van den Brand, Binghui Peng, Omri Weinstein, David P. Woodruff for helpful discussions in the early stage of this project and Pravesh Kothari and Gary Miller for helpful discussions on data structures. This work was mostly done while Lichen Zhang was at Carnegie Mellon University and Ruizhe Zhang was at University of Texas, Austin. The corrections in this version were found and written with the help of Claude Opus 5.5. Running as an agent in Claude Code, this large language model also mainly generated Sections~\ref{sec:plain}, \ref{sec:plain_incremental} and~\ref{sec:exact_solve} of this version: Theorems~\ref{thm:plain_dense}, \ref{thm:plain_gram} and~\ref{thm:plain_incremental}, Lemma~\ref{lem:exact_solve}, their algorithms and proofs, and the paragraph on the relation to the published algorithm. Independent Claude Opus 5.5 reviewers then checked these sections. The authors have verified the corrections and take full responsibility for the content of this paper.

\ifdefined\isarxiv
\bibliographystyle{alpha}
\else
\bibliographystyle{plainurl}
\fi
\bibliography{ref}

\newcommand{\etalchar}[1]{$^{#1}$}
\begin{thebibliography}{VWXXZ24}

\bibitem[ABH17]{abh17}
Naman Agarwal, Brian Bullins, and Elad Hazan.
\newblock Second-order stochastic optimization for machine learning in linear
  time.
\newblock {\em The Journal of Machine Learning Research}, 18(1):4148--4187,
  2017.

\bibitem[ADH{\etalchar{+}}19]{adhlw19}
Sanjeev Arora, Simon~S Du, Wei Hu, Zhiyuan Li, and Ruosong Wang.
\newblock Fine-grained analysis of optimization and generalization for
  overparameterized two-layer neural networks.
\newblock In {\em ICML}. arXiv preprint arXiv:1901.08584, 2019.

\bibitem[ADW{\etalchar{+}}25]{adw25}
Josh Alman, Ran Duan, Virginia~Vassilevska Williams, Yinzhan Xu, Zixuan Xu, and
  Renfei Zhou.
\newblock More asymmetry yields faster matrix multiplication.
\newblock In {\em Proceedings of the 2025 Annual ACM-SIAM Symposium on Discrete
  Algorithms (SODA)}, pages 2005--2039, 2025.

\bibitem[AKK{\etalchar{+}}20]{akk+20}
Thomas~D. Ahle, Michael Kapralov, Jakob B{\ae}k~Tejs Knudsen, Rasmus Pagh,
  Ameya Velingker, David~P. Woodruff, and Amir Zandieh.
\newblock Oblivious sketching of high-degree polynomial kernels.
\newblock In {\em Proceedings of the 2020 {ACM-SIAM} Symposium on Discrete
  Algorithms (SODA)}, pages 141--160, 2020.

\bibitem[ALS{\etalchar{+}}23]{als+22}
Josh Alman, Jiehao Liang, Zhao Song, Ruizhe Zhang, and Danyang Zhuo.
\newblock Bypass exponential time preprocessing: Fast neural network training
  via weight-data correlation preprocessing.
\newblock In {\em Advances in Neural Information Processing Systems},
  NeurIPS'23, 2023.

\bibitem[AMV21]{amv21}
Kyriakos Axiotis, Aleksander Madry, and Adrian Vladu.
\newblock Faster sparse minimum cost flow by electrical flow localization.
\newblock In {\em 2021 IEEE 62nd Annual Symposium on Foundations of Computer
  Science (FOCS)}, 2021.

\bibitem[ANW14]{anw14}
Haim Avron, Huy~L. Nguyen, and David~P. Woodruff.
\newblock Subspace embeddings for the polynomial kernel.
\newblock In {\em NeurIPS}, 2014.

\bibitem[AZLS19a]{als19_dnn}
Zeyuan Allen-Zhu, Yuanzhi Li, and Zhao Song.
\newblock A convergence theory for deep learning via over-parameterization.
\newblock In {\em ICML}, 2019.

\bibitem[AZLS19b]{als19_rnn}
Zeyuan Allen-Zhu, Yuanzhi Li, and Zhao Song.
\newblock On the convergence rate of training recurrent neural networks.
\newblock In {\em NeurIPS}, 2019.

\bibitem[Ber24]{b24}
Sergei Bernstein.
\newblock On a modification of chebyshev's inequality and of the error formula
  of laplace.
\newblock {\em Ann. Sci. Inst. Sav. Ukraine, Sect. Math}, 1(4):38--49, 1924.

\bibitem[BLH18]{blh18}
Alberto Bernacchia, Mate Lengyel, and Guillaume Hennequin.
\newblock Exact natural gradient in deep linear networks and its application to
  the nonlinear case.
\newblock In S.~Bengio, H.~Wallach, H.~Larochelle, K.~Grauman, N.~Cesa-Bianchi,
  and R.~Garnett, editors, {\em Advances in Neural Information Processing
  Systems}. Curran Associates, Inc., 2018.

\bibitem[BLSS20]{blss20}
Jan van~den Brand, Yin~Tat Lee, Aaron Sidford, and Zhao Song.
\newblock Solving tall dense linear programs in nearly linear time.
\newblock In {\em Proceedings of the 52nd Annual ACM SIGACT Symposium on Theory
  of Computing}, STOC 2020, page 775–788, 2020.

\bibitem[BPSW21]{bpsw21}
Jan van~den Brand, Binghui Peng, Zhao Song, and Omri Weinstein.
\newblock Training (overparametrized) neural networks in near-linear time.
\newblock In {\em ITCS}, 2021.

\bibitem[Bra20]{b20_lp}
Jan van~den Brand.
\newblock A deterministic linear program solver in current matrix
  multiplication time.
\newblock In {\em Proceedings of the Fourteenth Annual ACM-SIAM Symposium on
  Discrete Algorithms (SODA)}, pages 259--278. SIAM, 2020.

\bibitem[BRB17]{brb17}
Aleksandar Botev, Hippolyt Ritter, and David Barber.
\newblock Practical {G}auss-{N}ewton optimisation for deep learning.
\newblock In {\em Proceedings of the 34th International Conference on Machine
  Learning}, pages 557--565, 2017.

\bibitem[BWZ16]{bwz16}
Christos Boutsidis, David~P. Woodruff, and Peilin Zhong.
\newblock Optimal principal component analysis in distributed and streaming
  models.
\newblock In {\em S{TOC}'16---{P}roceedings of the 48th {A}nnual {ACM} {SIGACT}
  {S}ymposium on {T}heory of {C}omputing}, 2016.

\bibitem[CCZG21]{cczg21}
Zixiang Chen, Yuan Cao, Difan Zou, and Quanquan Gu.
\newblock How much over-parameterization is sufficient to learn deep {R}e{LU}
  networks?
\newblock In {\em International Conference on Learning Representations (ICLR)},
  2021.

\bibitem[CG19]{cg19}
Yuan Cao and Quanquan Gu.
\newblock Generalization bounds of stochastic gradient descent for wide and
  deep neural networks.
\newblock In {\em NeurIPS}, pages 10835--10845, 2019.

\bibitem[CGH{\etalchar{+}}19]{cgh+19}
Tianle Cai, Ruiqi Gao, Jikai Hou, Siyu Chen, Dong Wang, Di~He, Zhihua Zhang,
  and Liwei Wang.
\newblock Gram-gauss-newton method: Learning overparameterized neural networks
  for regression problems.
\newblock {\em arXiv preprint arXiv:1905.11675}, 2019.

\bibitem[Che52]{c52}
Herman Chernoff.
\newblock A measure of asymptotic efficiency for tests of a hypothesis based on
  the sum of observations.
\newblock {\em The Annals of Mathematical Statistics}, pages 493--507, 1952.

\bibitem[CLP{\etalchar{+}}21]{clp+21}
Beidi Chen, Zichang Liu, Binghui Peng, Zhaozhuo Xu, Jonathan~Lingjie Li, Tri
  Dao, Zhao Song, Anshumali Shrivastava, and Christopher Re.
\newblock {\{}MONGOOSE{\}}: A learnable {\{}lsh{\}} framework for efficient
  neural network training.
\newblock In {\em Proceedings of the Nineth International Conference on
  Learning Representations (ICLR'2021)}, 2021.

\bibitem[CLS19]{cls19}
Michael~B Cohen, Yin~Tat Lee, and Zhao Song.
\newblock Solving linear programs in the current matrix multiplication time.
\newblock In {\em STOC}, 2019.

\bibitem[CMJF{\etalchar{+}}20]{cmf+20}
Beidi Chen, Tharun Medini, Sameh~Gobriel James~Farwell, Charlie Tai, and
  Anshumali Shrivastava.
\newblock {SLIDE} : In defense of smart algorithms over hardware acceleration
  for large-scale deep learning systems.
\newblock In {\em MLSys'2020}, 2020.

\bibitem[CW13]{cw13}
Kenneth~L. Clarkson and David~P. Woodruff.
\newblock Low rank approximation and regression in input sparsity time.
\newblock In {\em Symposium on Theory of Computing Conference (STOC)}, pages
  81--90, 2013.

\bibitem[DFS16]{dfs16}
Amit Daniely, Roy Frostig, and Yoram Singer.
\newblock Toward deeper understanding of neural networks: The power of
  initialization and a dual view on expressivity.
\newblock In {\em Advances in Neural Information Processing Systems 29 (NIPS
  2016)}, pages 2253--2261, 2016.

\bibitem[DLL{\etalchar{+}}19]{dllwz19}
Simon~S Du, Jason~D Lee, Haochuan Li, Liwei Wang, and Xiyu Zhai.
\newblock Gradient descent finds global minima of deep neural networks.
\newblock In {\em International Conference on Machine Learning (ICML)}, 2019.

\bibitem[DLPM21]{mjmm21}
Michał Dereziński, Jonathan Lacotte, Mert Pilanci, and Michael~W. Mahoney.
\newblock Newton-less: Sparsification without trade-offs for the sketched
  newton update, 2021.

\bibitem[DS08]{ds08}
Samuel~I Daitch and Daniel~A Spielman.
\newblock Faster approximate lossy generalized flow via interior point
  algorithms.
\newblock In {\em Proceedings of the fortieth annual ACM symposium on Theory of
  computing (STOC)}, pages 451--460, 2008.

\bibitem[DSWZ22]{dswz22}
Yichuan Deng, Zhao Song, Omri Weinstein, and Ruizhe Zhang.
\newblock Fast distance oracles for any symmetric norm.
\newblock In {\em NeurIPS}, 2022.

\bibitem[DZPS19]{dzps19}
Simon~S Du, Xiyu Zhai, Barnabas Poczos, and Aarti Singh.
\newblock Gradient descent provably optimizes over-parameterized neural
  networks.
\newblock In {\em ICLR}, 2019.

\bibitem[EMMZ22]{emmz22}
Alessandro Epasto, Mohammad Mahdian, Vahab Mirrokni, and Peilin Zhong.
\newblock Improved sliding window algorithms for clustering and coverage via
  bucketing-based sketches.
\newblock In {\em Proceedings of the 2022 {A}nnual {ACM}-{SIAM} {S}ymposium on
  {D}iscrete {A}lgorithms ({SODA})}, 2022.

\bibitem[GQSW22]{gqsw22}
Yeqi Gao, Lianke Qin, Zhao Song, and Yitan Wang.
\newblock A sublinear adversarial training algorithm.
\newblock {\em arXiv preprint arXiv:2208.05395}, 2022.

\bibitem[GS22]{gs22}
Yuzhou Gu and Zhao Song.
\newblock A faster small treewidth sdp solver.
\newblock {\em arXiv preprint arXiv:2211.06033}, 2022.

\bibitem[GU18]{lgu18}
Fran\c{c}ois~Le Gall and Florent Urrutia.
\newblock Improved rectangular matrix multiplication using powers of the
  coppersmith-winograd tensor.
\newblock In {\em Proceedings of the Twenty-Ninth Annual ACM-SIAM Symposium on
  Discrete Algorithms}, SODA '18, page 1029–1046, 2018.

\bibitem[HJS{\etalchar{+}}22]{hjs+21}
Baihe Huang, Shunhua Jiang, Zhao Song, Runzhou Tao, and Ruizhe Zhang.
\newblock Solving sdp faster: A robust ipm framework and efficient
  implementation.
\newblock In {\em FOCS}, 2022.

\bibitem[HLSY21]{hlsy21}
Baihe Huang, Xiaoxiao Li, Zhao Song, and Xin Yang.
\newblock Fl-ntk: A neural tangent kernel-based framework for federated
  learning convergence analysis.
\newblock In {\em ICML}, 2021.

\bibitem[HM73]{hm73}
John~E. Hopcroft and J.~Musinski.
\newblock Duality applied to the complexity of matrix multiplication and other
  bilinear forms.
\newblock {\em SIAM Journal on Computing}, 2(3):159--173, 1973.

\bibitem[HSW{\etalchar{+}}22]{hsw+22}
Edward~J Hu, Yelong Shen, Phillip Wallis, Zeyuan Allen-Zhu, Yuanzhi Li, Shean
  Wang, Lu~Wang, and Weizhu Chen.
\newblock Lo{RA}: Low-rank adaptation of large language models.
\newblock In {\em International Conference on Learning Representations}, 2022.

\bibitem[HSWZ22]{hswz22}
Hang Hu, Zhao Song, Omri Weinstein, and Danyang Zhuo.
\newblock Training overparametrized neural networks in sublinear time.
\newblock {\em arXiv preprint arXiv:2208.04508}, 2022.

\bibitem[JGH18]{jgh18}
Arthur Jacot, Franck Gabriel, and Cl{\'e}ment Hongler.
\newblock Neural tangent kernel: convergence and generalization in neural
  networks.
\newblock In {\em Proceedings of the 32nd International Conference on Neural
  Information Processing Systems (NeurIPS)}, pages 8580--8589, 2018.

\bibitem[JKL{\etalchar{+}}20]{jklps20}
Haotian Jiang, Tarun Kathuria, Yin~Tat Lee, Swati Padmanabhan, and Zhao Song.
\newblock A faster interior point method for semidefinite programming.
\newblock In {\em FOCS}, 2020.

\bibitem[JLSW20]{jlsw20}
Haotian Jiang, Yin~Tat Lee, Zhao Song, and Sam Chiu-wai Wong.
\newblock An improved cutting plane method for convex optimization,
  convex-concave games and its applications.
\newblock In {\em STOC}, 2020.

\bibitem[JNW22]{jnw22}
Shunhua Jiang, Bento Natura, and Omri Weinstein.
\newblock A faster interior-point method for sum-of-squares optimization.
\newblock In {\em ICALP}, 2022.

\bibitem[JSWZ21]{jswz20}
Shunhua Jiang, Zhao Song, Omri Weinstein, and Hengjie Zhang.
\newblock Faster dynamic matrix inverse for faster lps.
\newblock In {\em STOC}, 2021.

\bibitem[KIW22]{kiw22}
Polina Kirichenko, Pavel Izmailov, and Andrew~Gordon Wilson.
\newblock Last layer re-training is sufficient for robustness to spurious
  correlations, 2022.

\bibitem[KMR14]{kmr14}
Felix Krahmer, Shahar Mendelson, and Holger Rauhut.
\newblock Suprema of chaos processes and the restricted isometry property.
\newblock {\em Communications on Pure and Applied Mathematics},
  67(11):1877--1904, 2014.

\bibitem[LG24]{lg24}
Francois Le~Gall.
\newblock Faster rectangular matrix multiplication by combination loss
  analysis.
\newblock In {\em SODA}, 2024.

\bibitem[LL18]{ll18}
Yuanzhi Li and Yingyu Liang.
\newblock Learning overparameterized neural networks via stochastic gradient
  descent on structured data.
\newblock In {\em NeurIPS}, 2018.

\bibitem[LR83]{lr83}
Grazia Lotti and Francesco Romani.
\newblock On the asymptotic complexity of rectangular matrix multiplication.
\newblock {\em Theoretical Computer Science}, 23(2):171--185, 1983.

\bibitem[LS14]{ls14}
Yin~Tat Lee and Aaron Sidford.
\newblock Path finding methods for linear programming: Solving linear programs
  in {\~{o}}(sqrt(rank)) iterations and faster algorithms for maximum flow.
\newblock In {\em 55th {IEEE} Annual Symposium on Foundations of Computer
  Science, {FOCS} 2014, Philadelphia, PA, USA, October 18-21, 2014}, pages
  424--433, 2014.

\bibitem[LSS{\etalchar{+}}20]{lss+20}
Jason~D Lee, Ruoqi Shen, Zhao Song, Mengdi Wang, and Zheng Yu.
\newblock Generalized leverage score sampling for neural networks.
\newblock In {\em NeurIPS}, 2020.

\bibitem[LSW15]{lsw15}
Yin~Tat Lee, Aaron Sidford, and Sam Chiu-wai Wong.
\newblock A faster cutting plane method and its implications for combinatorial
  and convex optimization.
\newblock In {\em Foundations of Computer Science (FOCS), 2015 IEEE 56th Annual
  Symposium on}, pages 1049--1065. IEEE, 2015.

\bibitem[LSZ19]{lsz19}
Yin~Tat Lee, Zhao Song, and Qiuyi Zhang.
\newblock Solving empirical risk minimization in the current matrix
  multiplication time.
\newblock In {\em Conference on Learning Theory (COLT)}, pages 2140--2157.
  PMLR, 2019.

\bibitem[LSZ{\etalchar{+}}23]{lszzz22}
S.~Cliff Liu, Zhao Song, Hengjie Zhang, Lichen Zhang, and Tianyi Zhou.
\newblock Space-efficient interior point method, with applications to linear
  programming and maximum weight bipartite matching.
\newblock In {\em ICALP}, 2023.

\bibitem[LXS{\etalchar{+}}19]{lxs+19}
Jaehoon Lee, Lechao Xiao, Samuel~S. Schoenholz, Yasaman Bahri, Roman Novak,
  Jascha Sohl{-}Dickstein, and Jeffrey Pennington.
\newblock Wide neural networks of any depth evolve as linear models under
  gradient descent.
\newblock In {\em Advances in Neural Information Processing Systems 32: Annual
  Conference on Neural Information Processing Systems 2019, NeurIPS 2019,
  December 8-14, 2019, Vancouver, BC, Canada}, pages 8570--8581, 2019.

\bibitem[MAM{\etalchar{+}}23]{mam+23}
Iman Mirzadeh, Keivan Alizadeh, Sachin Mehta, Carlo C~Del Mundo, Oncel Tuzel,
  Golnoosh Samei, Mohammad Rastegari, and Mehrdad Farajtabar.
\newblock Relu strikes back: Exploiting activation sparsity in large language
  models, 2023.

\bibitem[MG15]{jr15}
James Martens and Roger Grosse.
\newblock Optimizing neural networks with kronecker-factored approximate
  curvature.
\newblock In {\em Proceedings of the 32nd International Conference on
  International Conference on Machine Learning - Volume 37}, ICML'15, page
  2408–2417. JMLR.org, 2015.

\bibitem[NN13]{nn13}
Jelani Nelson and Huy~L Nguyen.
\newblock Sparsity lower bounds for dimensionality reducing maps.
\newblock In {\em Proceedings of the forty-fifth annual ACM symposium on Theory
  of computing}, pages 101--110, 2013.

\bibitem[O'D14]{o14}
Ryan O'Donnell.
\newblock {\em Analysis of {B}oolean Functions}.
\newblock Cambridge University Press, 2014.

\bibitem[{Ope}26]{OpenAIMM26}
{OpenAI}.
\newblock Complex matrix multiplication below {2.258} and rectangular bounds.
\newblock
  \href{https://github.com/openai/math/tree/main/preprints/Complex-Matrix-Multiplication-Below-2.258-and-Rectangular-Bounds-September-24-2026}{OpenAI
  Math Release preprint}, September 2026.

\bibitem[OS20]{os20}
Samet Oymak and Mahdi Soltanolkotabi.
\newblock Toward moderate overparameterization: Global convergence guarantees
  for training shallow neural networks.
\newblock {\em IEEE Journal on Selected Areas in Information Theory},
  1(1):84--105, 2020.

\bibitem[PW17]{pw17}
Mert Pilanci and Martin~J. Wainwright.
\newblock Newton sketch: A near linear-time optimization algorithm with
  linear-quadratic convergence.
\newblock {\em SIAM J. Optim.}, 27:205--245, 2017.

\bibitem[QSZZ23]{qszz23}
Lianke Qin, Zhao Song, Lichen Zhang, and Danyang Zhuo.
\newblock An online and unified algorithm for projection matrix vector
  multiplication with application to empirical risk minimization.
\newblock In {\em AISTATS}, 2023.

\bibitem[RRT12]{rrt12}
Holger Rauhut, Justin Romberg, and Joel~A Tropp.
\newblock Restricted isometries for partial random circulant matrices.
\newblock {\em Applied and Computational Harmonic Analysis}, 32(2):242--254,
  2012.

\bibitem[Sar06]{s06}
Tamas Sarlos.
\newblock Improved approximation algorithms for large matrices via random
  projections.
\newblock In {\em 2006 47th Annual IEEE Symposium on Foundations of Computer
  Science (FOCS)}, pages 143--152. IEEE, 2006.

\bibitem[SSWZ22]{sswz22}
Zhao Song, Baocheng Sun, Omri Weinstein, and Ruizhe Zhang.
\newblock Sparse fourier transform over lattices: A unified approach to signal
  reconstruction.
\newblock {\em arXiv preprint arXiv:2205.00658}, 2022.

\bibitem[SWY{\etalchar{+}}19]{swyzz19}
Zhao Song, Ruosong Wang, Lin Yang, Hongyang Zhang, and Peilin Zhong.
\newblock Efficient symmetric norm regression via linear sketching.
\newblock {\em Advances in Neural Information Processing Systems}, 32, 2019.

\bibitem[SWYZ21]{swyz21}
Zhao Song, David~P. Woodruff, Zheng Yu, and Lichen Zhang.
\newblock Fast sketching of polynomial kernels of polynomial degree.
\newblock In {\em ICML}, 2021.

\bibitem[SWZ17]{swz17}
Zhao Song, David~P Woodruff, and Peilin Zhong.
\newblock Low rank approximation with entrywise l1-norm error.
\newblock In {\em Proceedings of the 49th Annual ACM SIGACT Symposium on Theory
  of Computing}, pages 688--701, 2017.

\bibitem[SWZ19]{swz19}
Zhao Song, David~P Woodruff, and Peilin Zhong.
\newblock Relative error tensor low rank approximation.
\newblock In {\em Proceedings of the Thirtieth Annual ACM-SIAM Symposium on
  Discrete Algorithms (SODA)}, pages 2772--2789. SIAM, 2019.

\bibitem[SY19]{sy19}
Zhao Song and Xin Yang.
\newblock Quadratic suffices for over-parametrization via matrix chernoff
  bound.
\newblock {\em arXiv preprint arXiv:1906.03593}, 2019.

\bibitem[SY21]{sy21}
Zhao Song and Zheng Yu.
\newblock Oblivious sketching-based central path method for linear programming.
\newblock In {\em International Conference on Machine Learning (ICML)}, pages
  9835--9847. PMLR, 2021.

\bibitem[SYZ21]{syz21}
Zhao Song, Shuo Yang, and Ruizhe Zhang.
\newblock Does preprocessing help training over-parameterized neural networks?
\newblock In {\em Thirty-Fifth Conference on Neural Information Processing
  Systems (NeurIPS)}, 2021.

\bibitem[Vai89]{v89_lp}
Pravin~M Vaidya.
\newblock Speeding-up linear programming using fast matrix multiplication.
\newblock In {\em 30th Annual Symposium on Foundations of Computer Science},
  pages 332--337. IEEE, 1989.

\bibitem[VWXXZ24]{wxxz23}
Virginia Vassilevska~Williams, Yinzhan Xu, Zixuan Xu, and Renfei Zhou.
\newblock New bounds for matrix multiplication: from alpha to omega.
\newblock In {\em SODA}, 2024.

\bibitem[Wil12]{w12}
Virginia~Vassilevska Williams.
\newblock Multiplying matrices faster than coppersmith-winograd.
\newblock In {\em Proceedings of the forty-fourth annual ACM symposium on
  Theory of computing (STOC)}, pages 887--898. ACM, 2012.

\bibitem[Woo14]{w14}
David~P. Woodruff.
\newblock Sketching as a tool for numerical linear algebra.
\newblock {\em Foundations and Trends in Theoretical Computer Science},
  10(1--2):1--157, 2014.

\bibitem[WZ20]{wz20}
David~P Woodruff and Amir Zandieh.
\newblock Near input sparsity time kernel embeddings via adaptive sampling.
\newblock In {\em ICML}, 2020.

\bibitem[Ye20]{ye20}
Guanghao Ye.
\newblock Fast algorithm for solving structured convex programs.
\newblock {\em University of Washington Undergraduate Thesis}, 2020.

\bibitem[YGS{\etalchar{+}}21]{ygs+21}
Zhewei Yao, Amir Gholami, Sheng Shen, Mustafa Mustafa, Kurt Keutzer, and
  Michael Mahoney.
\newblock Adahessian: An adaptive second order optimizer for machine learning.
\newblock {\em Proceedings of the AAAI Conference on Artificial Intelligence},
  35(12):10665--10673, May 2021.

\bibitem[YJZ{\etalchar{+}}23]{yjz+23}
Hongru Yang, Ziyu Jiang, Ruizhe Zhang, Zhangyang Wang, and Yingbin Liang.
\newblock Convergence and generalization of wide neural networks with large
  bias, 2023.

\bibitem[ZCZG20]{zczg20}
Difan Zou, Yuan Cao, Dongruo Zhou, and Quanquan Gu.
\newblock Gradient descent optimizes over-parameterized deep relu networks.
\newblock In {\em Machine Learning}, 2020.

\bibitem[ZG19]{zg19}
Difan Zou and Quanquan Gu.
\newblock An improved analysis of training over-parameterized deep neural
  networks.
\newblock In {\em NeurIPS}, pages 2053--2062, 2019.

\bibitem[ZMG19]{zmg19}
Guodong Zhang, James Martens, and Roger~B Grosse.
\newblock Fast convergence of natural gradient descent for over-parameterized
  neural networks.
\newblock In {\em Advances in Neural Information Processing Systems (NeurIPS)},
  2019.

\end{thebibliography}
\ifdefined\isarxiv
\newpage

\appendix
\onecolumn

\newpage

\section*{Appendix}
\paragraph{Roadmap.} In Section~\ref{sec:preli_app}, we remind readers with the notations and some probability tools. In Section~\ref{sec:alg_app}, we illustrate the complete version of our algorithm and give a runtime analysis of it. In Section~\ref{sec:low_rank_app}, we design a simple low rank maintenance data structure and use it to efficiently implement matrix-vector product. In Section~\ref{sec:reg_app}, we introduce an efficient regression solver handling our Jacobian and Gram regression. In Section~\ref{sec:spectral_app}, we study the spectral property of the Gram matrix of the last layer and connect it with multi-layer neural tangent kernels. In Section~\ref{sec:analysis}, we analyze the convergence of our algorithm by using some heavy machinery such as structural analysis of the gradient and a careful double induction. In Section~\ref{sec:plain_incremental}, we give an incremental implementation of last-layer training, whose cost per step is sublinear in $m$ for suitable thresholds. In Section~\ref{sec:modified_als}, we give a detailed proof of one technical lemma.

\section{Preliminaries and Probability Tools}\label{sec:preli_app}
In this section, we introduce notations that will be used throughout the rest of the paper and several useful probability tools that will be heavily exploited in the later proofs.

\paragraph{Notations.} For any $n\in \mathbb{N}_+$, let 
$[n]$ denote the set $\{1,2,\cdots,n\}$. We use $\E[\cdot]$ for expectation and $\Pr[\cdot]$ for probability. 
 For any vector $x\in \R^d$, we use $\|x\|_2$ for the $\ell_2$ norm of a vector $x$. For any matrix $A\in \R^{m\times m}$, We use $\|A\|$ for the spectral norm of matrix $A$, i.e., $\|A\|=\max_{i\in [m]}\{|\lambda_i(A)\}$ where $\lambda_i(A)$ is the $i$-th eigenvalue. We use $\| A \|_F$ for the Frobenius norm of $A$. For any $A\in \R^{n\times m}$, we define $A^\top\in \R^{m\times n}$ to be the transpose of $A$. We use $I_m$ for the identity matrix of size $m\times m$. For matrix $A$ or vector $x$, $\|A\|_0,\|x\|_0$ denote the number of nonzero entries of $A$ and $x$ respectively. Note that $\|\cdot \|_0$ is a semi-norm since it satisfies triangle inequality. Given a real square matrix $A$, we use $\lambda_{\max}(A)$ and $\lambda_{\min}(A)$ for its largest and smallest eigenvalues respectively. Given a real matrix $A$, we use $\sigma_{\max}(A)$ and $\sigma_{\min}(A)$ for its largest and smallest singular values respectively. We use ${\cal N}(\mu,\sigma^2)$ for the Gaussian distribution with mean $\mu$ and variance $\sigma^2$. We use $\wt O(f(n))$ for $O(f(n)\cdot\poly \log(f(n))$. Let $\langle \cdot,\cdot \rangle$ be the inner product between two vectors or two matrices. When applying it to two vectors, this means the standard dot product between two vectors, and when applying to two matrices, this means $\langle A,B\rangle = \tr[A^\top B]$ where $\tr[A]$ denote the trace of matrix $A$. We use $x\otimes y=\vect(xy^\top)$ for the tensor product between two conforming vectors $x$ and $y$.

\label{sec:prob}
\begin{lemma}[Chernoff bound \cite{c52}]\label{lem:chernoff}
Let $X = \sum_{i=1}^n X_i$, where $X_1,\dots,X_n$ are $n$ independent 0/1 Bernoulli random variables with $\Pr[X_i = 1]=p_i$ for $i\in [n]$. Let $\mu = \E[X]$. Then, \\
1. $ \Pr[ X \geq (1+\epsilon) \mu ] \leq \exp ( - \epsilon^2 \mu / 3 ) $, $\forall~ \epsilon > 0$ ; \\
2. $ \Pr[ X \leq (1-\epsilon) \mu ] \leq \exp ( - \epsilon^2 \mu / 2 ) $, $\forall~ \epsilon\in (0,1)$. 
\end{lemma}

\begin{lemma}[Bernstein inequality \cite{b24}]\label{lem:bernstein}
Let $Z_1, \cdots, Z_n$ denote $n$ independent mean-zero random variables. Suppose that $|Z_i| \leq B$ almost surely, for all $i\in [n]$. Let $\sigma^2:=\sum_{i\in [n]}\E[Z_i^2]$. Then, for all $\tau>0$,
\begin{align*}
\Pr \left[ \sum_{i=1}^n Z_i > \tau \right] \leq \exp \left( - \frac{ \tau^2/2 }{ \sigma^2  + B \tau /3 } \right).
\end{align*}
\end{lemma}

\begin{lemma}[Concentration of subgaussian random variables]\label{lem:subgaussian}
Let $a\in \R^n$ be a vector where each coordinate of $a$ is an independent subgaussian random variable with parameter $\sigma^2$. Then, for any vector $x\in \R^n$,
\begin{align*}
    \Pr[|\langle a,x\rangle |\geq t\cdot \|x\|_2] \leq 2\exp\left(-\frac{t^2}{2\sigma^2}\right).
\end{align*}
\end{lemma}

\section{Complete Algorithm and its Runtime Analysis}\label{sec:alg_app}
In this section, we first present our complete algorithm, then we analyze its running time. We show that as long as we use the shifted ReLU activation so that the number of activated neurons is sparse, then all our operations can be realized in subquadratic time.

\begin{algorithm}[H]\caption{Training last layer.}
\label{alg:complete}
\begin{algorithmic}[1]
\Procedure{CompleteAlgorithm}{$X\in \R^{d\times n}, y\in \R^n, T$} \Comment{Theorem~\ref{thm:alg_app}}
    \State {\color{blue}/*Initialization*/}
    \State Initialize $W_\ell(0)$, $\forall \ell \in [L]$
    \State Compute $h_{i,\ell}$ for $\ell\in [L-1]$, with thresholds $\tau_{i,\ell}=\sqrt{2/m}\,b\,\|h_{i,\ell-1}\|_2$ \Comment{Takes $O(nm^2L)$ time}
    \State Store $h_{i,L-1}$ and $\tau_{i,L}=\sqrt{2/m}\,b\,\|h_{i,L-1}\|_2$ in memory, $\forall i\in [n]$
    \State \textsc{LowRankMaintenance} \textsc{LMR} \Comment{Algorithm~\ref{alg:low_rank_ds}}
    \State \textsc{LMR}.\textsc{Init}($\{W_2(0),\ldots,W_L(0)\}$)
    \For{$t= 0 \to T-1$}
        \State {\color{blue}/*Forward computation*/}
        \State $v_{i,L}\gets h_{i,L-1},\forall i\in [n]$
        \State $g_{i,L}\gets \textsc{LMR.Query}(L,h_{i,L-1}),\forall i\in [n]$ \Comment{Takes $o(nm^2)$ time}
        \State $h_{i,L}\gets \phi_L(g_{i,L}),\forall i\in [n]$, threshold $\tau_{i,L}$ \Comment{$h_{i,L}$ is sparse, $\phi_L$ as in Definition~\ref{def:architecture}}
        \State $D_{i,L}\gets \diag(\phi'(g_{i,L})),\forall i\in [n]$ \Comment{$D_{i,L}$ is sparse}
        \State $f_t\gets [a^\top h_{1,L},\ldots,a^\top h_{n,L}]^\top$ \Comment{Takes $O(nm)$ time}
        \State {\color{blue}/*Backward computation*/}
        \State $u_{i,L}\gets a^\top D_{i,L},\forall i\in [n]$ \Comment{Takes $O(nm)$ time}
        \State $g_L\gets \textsc{FastTensorRegression}(\{u_{i,L}\}_{i=1}^n,\{v_{i,L}\}_{i=1}^n,f_t-y)$ with precision $\epsilon_0$ (Definition~\ref{def:R}) 
        \State \textsc{LMR.Update}$(\{-g_{L,i}u_{i,L}\}_{i=1}^n,\{v_{i,L}\}_{i=1}^n)$
    \EndFor
\EndProcedure
\end{algorithmic}
\end{algorithm}

\begin{theorem}[Formal version of Theorem~\ref{thm:alg_main}]
\label{thm:alg_app}
Let $X\in \R^{d\times n}$ and $y\in \R^n$, and let $s_D:=\max_{i\in[n],\ell\in[L]}\|D_{i,\ell}(0)\|_0$ denote the sparsity of the matrices $D_{i,\ell}(0)$. Let $m$ denote the width of neural network, $L$ denote the number of layers and $\alpha$ denote the dual matrix multiplication exponent (Def.~\ref{def:omega_alpha}). Let $a=\min\{\alpha,1/2\}$ and suppose that $n\leq m^a$. Then the running time of Algorithm~\ref{alg:complete} is
\begin{align*}
    O({\cal T}_{\mathrm{init}}+T\cdot {\cal T}_{\mathrm{iter}}),
\end{align*}
where
\begin{align*}
    {\cal T}_{\mathrm{init}} = & ~ O(m^2nL), \\
    {\cal T}_{\mathrm{iter}} = & ~ nm^{2-a+o(1)}+\wt O(n\cdot (m\cdot s_D+\lambda_L^{-1}n^\omega)).
\end{align*}
Therefore, the \emph{amortized cost per iteration} of Algorithm~\ref{alg:complete} is
\begin{align*}
    nm^{2-a+o(1)}+\wt O(n\cdot (m\cdot s_D+\lambda_L^{-1}n^\omega)).
\end{align*}
\end{theorem}

\begin{proof}
We analyze ${\cal T}_{\mathrm{init}}$ and ${\cal T}_{\mathrm{iter}}$ separately.

{\bf Initialization time.} We will first initialize $(L-1)$ $m\times m$ matrices and one $m\times d$ matrix, which takes $O(m^2L)$ time. Compute $h_{i,L-1}$ for all $i\in [n]$ takes $O(m^2nL)$ time, and the thresholds $\tau_{i,\ell}$ take $O(m)$ time each, $O(mnL)$ in total. Finally, initialize the data structure takes $O(m^2L)$ time. Hence, ${\cal T}_{\mathrm{init}}=O(m^2nL)$.

{\bf Cost per iteration.} For each iteration, we perform one forward computation from layer 1 to $L$, then we train the last layer via solving a regression based on its Jacobian matrix.
\begin{itemize}
    \item {\bf Forward computation:} We compute $g_{i,L}\in \R^m$ by the \textsc{Query} procedure of the \textsc{LMR} data structure. Since only $W_L$ is trained, the queried vector $h_{i,L-1}=h_{i,L-1}(0)$ never changes, and it has at most $s_D$ nonzero entries, as $\|h_{i,L-1}(0)\|_0\leq\|D_{i,L-1}(0)\|_0$ (Lemma~\ref{lem:init_sparse}). By Lemma~\ref{lem:low_rank_maintain}, a query takes $O(m\cdot (s_D+m^a))$ time. Computing $h_{i,L}$ and $D_{i,L}$ takes $O(m)$ time. Over the $n$ data points, the forward computation takes $O(nms_D+nm^{1+a})$ time, and $nm^{1+a}\leq nm^{2-a}$ because $a\leq 1/2$.
    \item {\bf Backward computation:} In backward computation, we first compute $u_{i,L}=D_{i,L}a\in \R^m$, which takes $O(m)$ time since $D_{i,L}$ is diagonal. Then, we invoke Theorem~\ref{thm:fast_kronecker_regression} to solve the Gram regression problem, which due to Theorem~\ref{thm:fast_kronecker_regression} takes $\wt O(mn+n^{\omega+1}/\lambda_L)$ time as we pick $\epsilon=\epsilon_0=\sqrt{\lambda_L/n}/30$ (Definition~\ref{def:R}). The solver is run with failure probability $e^{-\log^2 n}$ (Definition~\ref{def:R}), which costs only a polylogarithmic factor in running time, through the factor $\log^3(nm/(\epsilon_0\delta))$ in the proof of Theorem~\ref{thm:fast_kronecker_regression}, absorbed by the $\wt O(\cdot)$ notation. Finally, the update takes $O(nm^{2-a+o(1)})$ amortized time owing to Lemma~\ref{lem:low_rank_maintain}.
\end{itemize}
Since $\wt O(mn)\leq nm^{2-a+o(1)}$, the total amortized cost per iteration is $nm^{2-a+o(1)}+\wt O(n\cdot (ms_D+\lambda_L^{-1}n^\omega))$. The bound is amortized, since an iteration that rebuilds $W_L$ in the data structure takes $m^{2+o(1)}$ time. No term depends on the number of activations that change during training. This concludes the proof.
\end{proof}

\begin{corollary}\label{cor:alg_cost}
Let $a=\min\{\alpha,1/2\}$, choose the threshold parameter $b=\sqrt{2\gamma\log m}$ for a constant $\gamma\in[a,1)$, and suppose that the width $m$ satisfies the width condition of Definition~\ref{def:m}. Then, with probability at least $1-1/\poly(n)$ over the initialization, the amortized cost per iteration of Algorithm~\ref{alg:complete} is $nm^{2-a+o(1)}$. Moreover, $n^4\leq m^{1-a}$, so the amortized cost per iteration is at most $m^{(9-5a)/4+o(1)}$.
\end{corollary}

\begin{proof}
By Remark~\ref{remark:k}, the sparsity parameter of Definition~\ref{def:m} is $m\,e^{-b^2/2}=m^{1-\gamma}$. Under the width condition, Lemma~\ref{lem:init_sparse} shows that, with probability at least $1-1/\poly(n)$, we have $\|D_{i,\ell}(0)\|_0\leq 2m^{1-\gamma}$ for all $i\in[n]$ and $\ell\in[L]$. So the parameter $s_D$ of Theorem~\ref{thm:alg_app} is at most $2m^{1-\gamma}\leq 2m^{1-a}$. Since $C_W\geq 1$, $1+b\geq 1$, $L\geq 1$, $\lambda_*\leq 1$, $\lambda_L\leq 1$ and $\log(mnL)\geq 1$, the width condition gives $m^{1-\gamma}\geq n^4\lambda_L^{-4}\geq n^4$. Hence $n\leq m^{(1-\gamma)/4}\leq m^{1/4}\leq m^a$, as $\alpha>1/4$~\cite{wxxz23}, so Theorem~\ref{thm:alg_app} applies. The same inequality gives $\lambda_L^{-1}\leq m^{(1-\gamma)/4}/n$, so $\lambda_L^{-1}n^{\omega+1}\leq m^{(1-\gamma)/4}n^{\omega}\leq m^{(1-\gamma)(1+\omega)/4}\leq m$, as $\omega\leq 3$. Also $nms_D\leq 2nm^{2-a}$. So every term of Theorem~\ref{thm:alg_app} is at most $nm^{2-a+o(1)}$, where the factor $m^{o(1)}$ also absorbs the polylogarithmic factors. Finally, $n^4\leq m^{1-\gamma}\leq m^{1-a}$ gives $nm^{2-a+o(1)}\leq m^{(1-a)/4+2-a+o(1)}=m^{(9-5a)/4+o(1)}$.

The condition of Theorem~\ref{thm:conv_main} implies the width condition (see the paragraph after Definition~\ref{def:m}), so this corollary proves Theorem~\ref{thm:alg_main}.
\end{proof}

\begin{remark}
As long as the neural network is wide enough, as in Definition~\ref{def:m}, and we choose the threshold parameter properly, as in Remark~\ref{remark:k}, then we can make sure that the sparsity parameter $k$ is $o(m)$, and we achieve subquadratic amortized cost per iteration. Only $k$ matters here, since the input $h_{i,L-1}$ of the trained last layer never changes.

We also compare our result with other approaches. A naive implementation of variants of gradient descent takes $O(nm^2)$ time per step, namely, one evaluates the gradient with respect to each data point and sums them up. By batching the $n$ data points and using fast rectangular matrix multiplication, the running time can be improved to $\Tmat(m,n,m)$, which is $m^{2+o(1)}$ when $n\leq m^{\alpha}$. To truly beat the quadratic barrier, we need $nm^{2-a+o(1)}=o(m^2)$, which holds when $n\leq m^{a-\Omega(1)}$. Under the width condition, we have $n\leq m^{1/4}$ and $a>1/4$, so we get a \emph{truly subquadratic time} in $m$. Note that the count $O(nm^2)$ ignores the sparsity of the fixed input $h_{i,L-1}$, which has $O(k)$ nonzero entries. If we store $W_L$ explicitly, then the product $W_Lh_{i,L-1}$ and the rank-one update $W_L\gets W_L+u\,h_{i,L-1}^\top$ for a vector $u\in\R^m$ touch only the columns of $W_L$ in the support of $h_{i,L-1}$. So this simple implementation already performs these operations in $O(nmk)$ time per step. We develop this observation in Section~\ref{sec:plain}.
\end{remark}
\section{Low-Rank Maintenance}\label{sec:low_rank_app}
In this section, we design a lazy data structure that maintains a low-rank representation of the change of weights and supports matrix-vector product queries efficiently. We state it for the layers $\ell\in\{2,\dots,L\}$, whose weight matrices are $m\times m$. Algorithm~\ref{alg:complete} uses it only for the trained layer $W_L$.

Before moving, we define some notions related to rectangular matrix multiplication.

\begin{definition}[\cite{w12,lgu18,wxxz23,lg24}]\label{def:omega_alpha}
Let $\omega$ be the matrix multiplication exponent such that it takes $n^{\omega+o(1)}$ time to multiply two $n\times n$ matrices.

Let $\alpha$ be the dual exponent of the matrix multiplication which is the supremum among all $c\geq 0$ such that it takes $n^{2+o(1)}$ time to multiply an $n\times n$ by an $n\times n^c$ matrix.

Additionally, we define the function $\omega(\cdot)$ where $\omega(c)$ denotes the exponent of multiplying an $n\times n$ matrix by an $n\times n^c$ matrix. Hence, we have $\omega(1)=\omega$ and $\omega(\alpha)=2$.
\end{definition}

The overall idea of our low-rank maintenance data structure is as follows: we keep accumulating the low-rank change, when the rank of the change reaches a certain threshold ($m^a$, where $a=\min\{\alpha,1/2\}$), then we restart the data structure and update the weight matrix.

\begin{algorithm}[H]\caption{Low-rank maintenance data structure}
\label{alg:low_rank_ds}
\begin{algorithmic}[1]
\State {\bf data structure} \textsc{LowRankMaintenance} \Comment{Lemma~\ref{lem:low_rank_maintain}}
\State \hspace{4mm} {\bf members}
\State \hspace{8mm} $r_\ell,\forall \ell\in\{2,\dots,L\}$ \Comment{$r_\ell$ denotes the accumulated rank of the change}
\State \hspace{8mm} $W_\ell,\forall \ell\in\{2,\dots,L\}$ \Comment{$W_\ell\in\R^{m\times m}$}
\State \hspace{8mm} $\Delta W_\ell,\forall \ell\in\{2,\dots,L\}$ \Comment{$\Delta W_\ell\in\R^{m\times m}$}
\State \hspace{4mm} {\bf end members}
\State
\State \hspace{4mm} {\bf procedures}
\State \hspace{8mm} \textsc{Init}$(\{W_2(0),\ldots, W_L(0) \})$ \Comment{Initialize the data structure}
\State \hspace{8mm} \textsc{Update}$(U_\ell,V_\ell)$ \Comment{Update the low rank representation}
\State \hspace{8mm} \textsc{Query}$(\ell,y)$ \Comment{Compute the matrix-vector product between $\Delta W_\ell$ and $y$}
\State \hspace{4mm} {\bf end procedures}
\State {\bf end data structure}
\end{algorithmic}
\end{algorithm}

\begin{algorithm}[H]\caption{Procedures of LRM data structure}
\label{alg:low_rank_procedure}
\begin{algorithmic}[1]
\Procedure{Init}{$\{W_2(0),\ldots,W_L(0)\}$} \Comment{Lemma~\ref{lem:low_rank_maintain}}
\State $W_\ell\gets W_\ell(0),\forall \ell\in\{2,\dots,L\}$
\State $\Delta W_\ell\gets 0,\forall \ell\in\{2,\dots,L\}$
\State $r_\ell\gets 0,\forall \ell\in\{2,\dots,L\}$
\EndProcedure
\State
\Procedure{Update}{$U_\ell\in \R^{m\times n},V_\ell\in \R^{m\times n}$} \Comment{Lemma~\ref{lem:low_rank_maintain}}
\State $\Delta W_\ell \gets \Delta W_\ell+U_\ell V_\ell^\top$ without forming the product and sum the two matrices
\State $r_\ell\gets r_\ell+n$
\If{$r_\ell\geq m^a$ where $a=\min\{\alpha,1/2\}$}
\State $W_\ell\gets W_\ell+\Delta W_\ell$ \Comment{Forms $\Delta W_\ell$ from its factors in $m^{2+o(1)}$ time (Lemma~\ref{lem:low_rank_maintain})}
\State $r_\ell\gets 0$
\State $\Delta W_\ell\gets 0$
\EndIf 
\EndProcedure
\State
\Procedure{Query}{$\ell\in\{2,\dots,L\},y\in \R^m$} \Comment{Lemma~\ref{lem:low_rank_maintain}}
\State $z\gets W_\ell\cdot y+\Delta W_\ell\cdot y$ \Comment{Takes $O(\nnz(y)\cdot m+mr_\ell)$ time}
\State \Return $z$
\EndProcedure
\end{algorithmic}
\end{algorithm}

\begin{lemma}\label{lem:low_rank_maintain}
Let $a:=\min\{\alpha,1/2\}$ and suppose that $n\leq m^a$. There exists a deterministic data structure~(Algorithm~\ref{alg:low_rank_ds}) that maintains
\begin{align*}
\Delta W_2,\ldots,\Delta W_L
\end{align*}
such that
\begin{itemize}
    \item The procedure \textsc{Init} (Algorithm~\ref{alg:low_rank_procedure}) takes $O(m^2L)$ time.
    \item The procedure \textsc{Update} (Algorithm~\ref{alg:low_rank_procedure}) takes $O(nm^{2-a+o(1)})$ amortized time.
    \item The procedure \textsc{Query} (Algorithm~\ref{alg:low_rank_procedure}) takes $O(m\cdot (\nnz(y)+r_\ell))$ time, where $r_\ell<m^a$ is the accumulated rank of $\Delta W_\ell$ when \textsc{Query} is called.
\end{itemize}
\end{lemma}

\begin{proof}
The bound for \textsc{Init} is clear. For \textsc{Query}, the product of the (possibly dense) matrix $W_\ell\in \R^{m\times m}$ with $y$ takes $O(\nnz(y)\cdot m)$ time. Writing $\Delta W_\ell=UV^\top$ with $U,V\in \R^{m\times r_\ell}$, the product $\Delta W_\ell y=U(V^\top y)$ takes $O(mr_\ell)$ time. Every call of \textsc{Update} that reaches $r_\ell\geq m^a$ resets $r_\ell$ to $0$, so $r_\ell<m^a$ whenever \textsc{Query} is called.

It remains to analyze the amortized cost of \textsc{Update}. If $r_\ell<m^a$ after the increment, then we only append the $m\times n$ blocks $U_\ell,V_\ell$ to the stored factors, in $O(mn)$ time. If $r_\ell\geq m^a$, then we explicitly form the $m\times m$ matrix $\Delta W_\ell$. To form it, notice we have accumulated $r_\ell/n$ different sums of rank-$n$ decompositions, which can be represented as
\begin{align*}
    U = & ~ [U_\ell(1),U_\ell(2),\ldots,U_\ell(r_\ell/n)]\in \R^{m\times r_\ell}, \\
    V = & ~[V_\ell(1),V_\ell(2),\ldots,V_\ell(r_\ell/n)] \in \R^{m\times r_\ell},
\end{align*}
and $\Delta W_\ell=UV^\top$. Since $r_\ell<m^a+n\leq 2m^a$, we can split the inner dimension into two parts of size at most $m^a$, so $UV^\top$ is the sum of two products of an $m\times m^a$ matrix and an $m^a\times m$ matrix. The exponent of rectangular matrix multiplication does not change when the three dimensions are permuted~\cite{hm73,lr83}, so each product takes $m^{\omega(a)+o(1)}$ time, and $\omega(a)=2$ since $a\leq\alpha$ (Definition~\ref{def:omega_alpha}). Adding $\Delta W_\ell$ to $W_\ell$ takes $O(m^2)$ time. Since $r_\ell$ grows by $n$ per call and starts from $0$ after each rebuild, at least $m^a/n\geq 1$ calls happen between two rebuilds. We charge the rebuild to these calls, which gives $O(m^{2+o(1)}\cdot n/m^a)=O(nm^{2-a+o(1)})$ amortized time per call. This also dominates the $O(mn)$ time of the calls without a rebuild.
\end{proof}

\begin{remark}
The two bounds on $a$ play different roles. The bound $a\leq\alpha$ keeps each rebuild at $m^{2+o(1)}$ time. The bound $a\leq 1/2$ balances \textsc{Update} against \textsc{Query}. A query pays $mr_\ell<m^{1+a}$ for the low-rank part, and $m^{1+a}\leq m^{2-a}$ exactly when $a\leq 1/2$. So with the bounds of Lemma~\ref{lem:low_rank_maintain}, a larger threshold would make the queries the bottleneck. The proof uses only $a\leq\alpha$ and $n\leq m^a$. So the lemma also holds with $a$ replaced by any constant $a'\in(0,a]$ with $n\leq m^{a'}$, for instance $a'=\min\{\alpha_0,1/2\}$ for a proven lower bound $\alpha_0$ on $\alpha$, which is what an implementation would use. With the bound $\alpha>0.465$ over every field of characteristic zero~\cite[Theorem~1.1 and Corollaries~14.1--14.2]{OpenAIMM26}, we have $a>0.465$, so the amortized time of \textsc{Update} is $O(nm^{1.535})$, and $nm^{1.535+o(1)}$ with $a'=0.465$. The time of \textsc{Query} is $O(m\cdot \nnz(y)+m^{1+a})\leq O(m\cdot \nnz(y)+m^{3/2})$. In our application of training neural networks, the queried vectors are sparse.
\end{remark}

\section{Fast Tensor Product Regression}\label{sec:reg_app}
In this section, we show how to solve a specific type of regression fast using tensor-based sketching matrices for approximation.

Consider the following problem: Given two matrices $U=[u_1,\ldots,u_n]\in\R^{m\times n}$ and $V=[v_1,\ldots,v_n]\in \R^{m\times n}$ with $m\gg n$, consider the matrix $J\in \R^{n\times m^2}$ formed by
\begin{align*}
    J = & ~ \begin{bmatrix}
    \vect(u_1v_1^\top)^\top \\
    \vect(u_2v_2^\top)^\top \\
    \vdots \\
    \vect(u_nv_n^\top)^\top
    \end{bmatrix}.
\end{align*}
We are also given a vector $c$ in $n$ dimension, our task is to find a solution to the following regression problem: 
\begin{align*}
    \min_{x\in \R^n}~\|JJ^\top x-c\|_2^2.
\end{align*}

Our main theorem for this section is Theorem~\ref{thm:fast_kronecker_regression} in Section~\ref{sec:inexact_solvers}.

\subsection{Approximate \texorpdfstring{$J$}{J} via \texorpdfstring{$\tensorsketch$}{TensorSketch}}
We introduce the notion of $\tensorsketch$ for two vectors:
\begin{definition}
Let $h_1,h_2:[m]\rightarrow [s]$ be 3-wise independent hash functions, also let $\sigma:[m]\rightarrow \{\pm 1\}$ be a 4-wise independent random sign function. The degree two $\tensorsketch$ transform, $S:\R^m\times \R^m\rightarrow \R^s$ is defined as follows: for any $i,j\in [m]$ and $r\in [s]$,
\begin{align*}
S_{r,(i,j)} = & ~ \sigma_1(i)\cdot \sigma_2(j)\cdot \mathbf{1}[h_1(i)+h_2(j)=r~\textnormal{mod}~s].
\end{align*}
\end{definition}
\begin{remark}
Apply $S$ to two vectors $x,y\in \R^m$ can be implemented in time $O(s\log s+\nnz(x)+\nnz(y))$.
\end{remark}

We introduce one key technical lemma from~\cite{anw14}:
\begin{lemma}[Theorem 1 of~\cite{anw14}]
\label{lem:tensorsketch_ose}
Let $S\in \R^{s\times m^2}$ be the $\tensorsketch$ matrix, consider a fixed $n$-dimensional subspace $V$. If $s=\Omega(n^2/(\epsilon^2\delta))$, then with probability at least $1-\delta$, $\|Sx\|_2=(1\pm\epsilon)\|x\|_2$ simultaneously for all $x\in V$.
\end{lemma}

Now we are ready to prove the main lemma of this section:
\begin{lemma}
Let $\epsilon,\delta\in (0,1)$ denote two parameters. Let $J\in \R^{n\times m^2}$ represent a matrix such that the $i$-th row of $J$ is equal to $\vect(u_iv_i^\top)$ for some $u_i,v_i\in \R^m$. Then, we can compute a matrix $\wt J\in \R^{n\times s}$ such that, with probability at least $1-\delta$, for all $x\in \R^n$ simultaneously, we have
\begin{align*}
    \|\wt J^\top x\|_2 = & ~ (1\pm \epsilon) \|J^\top x\|_2,
\end{align*}
where $s=\Omega(n^2/(\epsilon^2\delta))$. The time to compute $\wt J$ is $O(ns\log s+\nnz(U)+\nnz(V))$.
\end{lemma}

\begin{proof}
Notice that the row space of matrix $J$ can be viewed as an $n$-dimensional subspace, hence, by Lemma~\ref{lem:tensorsketch_ose}, the $\tensorsketch$ matrix $S$ with $s=\Omega(n^2/(\epsilon^2\delta))$ can preserve the length of all vectors in the subspace generated by $J^\top$ with probability $1-\delta$, to a multiplicative factor of $1\pm \epsilon$. 

The running time part is to apply the FFT algorithm to each row of $J$ with a total of $n$ rows. For each row, it takes $O(s\log s+m)$ time, hence the overall running time is $O(n(s\log s+m))$.
\end{proof}

\subsection{Approximate \texorpdfstring{$J$}{J} via \texorpdfstring{$\tensorsrht$}{TensorSRHT}}
We note that the dependence on the target dimension of sketching is $O(1/\delta)$ for $\tensorsketch$. We introduce another kind of sketching technique for tensor, called $\tensorsrht$. The tradeoff is we lose input sparsity runtime of matrices $U$ and $V$. 

\begin{definition}
We define the $\tensorsrht$ $S:\R^m\times \R^m\rightarrow \R^s$ as $S=\frac{1}{\sqrt s} P \cdot (H D_1\times H D_2)$, where each row of $P\in \{0,1\}^{s\times m^2}$ contains only one $1$ at a random coordinate, one can view $P$ as a sampling matrix. $H$ is a $m\times m$ Hadamard matrix, and $D_1,D_2$ are two $m\times m$ independent diagonal matrices with diagonals that are each independently set to be a Rademacher random variable (uniform in $\{-1,1\}$). 
\end{definition}

\begin{remark}
By using FFT algorithm, apply $S$ to two vectors $x,y\in \R^m$ takes time $O(m\log m+s)$.
\end{remark}

We again introduce a technical lemma for $\tensorsrht$.

\begin{lemma}[Theorem 3 of~\cite{akk+20}]
\label{lem:tensorsrht_ose}
Let $S\in \R^{s\times m^2}$ be the $\tensorsrht$ matrix, consider a fixed $n$-dimensional subspace $V$. If $s=\Omega(n\log^3(nm/\epsilon\delta)\epsilon^{-2})$, then with probability at least $1-\delta$, $\|Sx\|_2=(1\pm\epsilon)\|x\|_2$ simultaneously for all $x\in V$.
\end{lemma}

\begin{lemma}
\label{lem:tensorsrht}
Let $\epsilon,\delta\in (0,1)$ denote two parameters. Given a list of vectors $u_1, \cdots, u_n, v_1, \cdots, v_n \in \R^m$. Let $J\in \R^{n\times m^2}$ represent a matrix where the $i$-th row of $J$ is equal to $\vect(u_iv_i^\top)$. Then, we can compute a matrix $\wt J\in \R^{n\times s}$ such that, with probability at least $1-\delta$, for all $x\in \R^n$ simultaneously, we have
\begin{align*}
    \|\wt J^\top x\|_2 = & ~ (1\pm\epsilon)\|J^\top x\|_2,
\end{align*}
where $s=\Omega(n\log^3(nm/(\epsilon\delta))\epsilon^{-2})$. The time to compute $\wt J$ is $O(n(m\log m+s))$.
\end{lemma}

\begin{proof}
The correctness follows directly from Lemma~\ref{lem:tensorsrht_ose}. The running time follows from the FFT algorithm to each row of $J$, each application takes $O(m\log m+s)$ time, and we need to apply it to $n$ rows.
\end{proof}

\subsection{Guarantees of Inexact Solvers}\label{sec:inexact_solvers}
Now that we have tools to quickly form an approximate Jacobian, our algorithm  is relatively straightforward: forming the approximate Jacobian, computing the Gram then inverting it. We analyze its error guarantees and runtime.

\begin{theorem}\label{thm:fast_kronecker_regression}
Given two $m\times n$ matrices $U=[u_1,\ldots,u_n]$ and $V=[v_1,\ldots,v_n]$, and a target vector $c\in \R^n$. Let $J = [\vect(u_1v_1^\top)^\top,\ldots,\vect(u_nv_n^\top)^\top] \in \R^{n \times m^2}$. Let $\epsilon,\delta\in(0,1)$. There is an algorithm that takes $\wt O(nm+\epsilon^{-2}n^\omega)$ time and outputs a vector $\wh x\in \R^n$ such that, with $\kappa:=\lambda_{\max}(JJ^\top)/\lambda_{\min}(JJ^\top)$,
\begin{align*}
    \| J J^\top \wh x - c \|_2 \leq \epsilon\sqrt{\kappa}\, \|c\|_2 ~~~\text{and}~~~ \|\wh x-(JJ^\top)^{-1}c\|_2\leq \epsilon\,\|(JJ^\top)^{-1}\|\,\|c\|_2
\end{align*}
holds with probability at least $1-\delta$.
\end{theorem}

\begin{proof}
Throughout the proof, we will let $Q:=JJ^\top$, and we will let $\wt J=JS^\top \in \R^{n\times s}$ be the sketched Jacobian, for $S\in \R^{s\times m^2}$ being a $\tensorsrht$ matrix, and $s$ will be specified later. We will let $\wt Q:=\wt J\wt J^\top$ to denote the approximate Gram matrix. We run the sketch at precision $\epsilon':=\epsilon/10\leq 1/10$ and set $s=O(\epsilon'^{-2}n\log^3(nm/(\epsilon'\delta)))=O(\epsilon^{-2}n\log^3(nm/(\epsilon\delta)))$, therefore, with probability at least $1-\delta$, for all $x\in \R^{n}$ simultaneously, $\|\wt J^\top x\|_2=(1\pm\epsilon')\|J^\top x\|_2$. This also indicates that
\begin{align*}
    (1-\epsilon')^2 Q \preceq \wt Q \preceq (1+\epsilon')^2 Q.
\end{align*}

Our algorithm works as follows: we form the matrix $\wt J$, then compute $\wt Q=\wt J\wt J^\top$, and solve the regression as $\wh x=\wt x:=\wt Q^{-1}c$. Then we need to measure the loss
\begin{align*}
    \|Q\wt x-c\|_2 = & ~ \|Q\wt Q^{-1}c-c\|_2 \\
    = & ~ \|Q(\wt Q^{-1}-Q^{-1})c\|_2 \\
    \leq & ~ \|Q(\wt Q^{-1}-Q^{-1})\| \|c\|_2,
\end{align*}
and similarly $\|\wt x-Q^{-1}c\|_2\leq \|\wt Q^{-1}-Q^{-1}\|\|c\|_2$. Let $M:=Q^{1/2}(\wt Q^{-1}-Q^{-1})Q^{1/2}=Q^{1/2}\wt Q^{-1}Q^{1/2}-I$, which is symmetric. By the spectral bounds above, the eigenvalues of $Q^{1/2}\wt Q^{-1}Q^{1/2}$ lie in $[(1+\epsilon')^{-2},(1-\epsilon')^{-2}]$, so $\|M\|\leq (1-\epsilon')^{-2}-1\leq 3\epsilon'\leq\epsilon$ as $\epsilon'\leq 1/10$. Hence
\begin{align*}
    \|Q(\wt Q^{-1}-Q^{-1})\| = & ~ \|Q^{1/2}MQ^{-1/2}\| \\
    \leq & ~ \|Q^{1/2}\|\cdot \|M\|\cdot \|Q^{-1/2}\| \\
    \leq & ~ \epsilon\sqrt{\kappa}, \\
    \|\wt Q^{-1}-Q^{-1}\| = & ~ \|Q^{-1/2}MQ^{-1/2}\| \leq \epsilon\|Q^{-1}\|.
\end{align*}
This completes the proof of correctness.

It remains to analyze the runtime. To form the matrix $\wt J$, owing to Lemma~\ref{lem:tensorsrht}, it takes time $O(nm\log m+\epsilon^{-2}n^2\log^3(nm/(\epsilon\delta)))$. Computing the matrix $\wt Q$ takes time
\begin{align*}
    \Tmat(n,s,n) = & ~ \wt O(sn^{\omega-1}) \\
    = & ~ \wt O(\epsilon^{-2} n^\omega).
\end{align*}
Finally, inverting the matrix takes $O(n^\omega)$ time. Thus, the total runtime cost is 
\begin{align*}
    \wt O(nm+\epsilon^{-2}n^\omega).
\end{align*}
\end{proof}

\begin{remark}
Due to the probability requirement (union bounding over all data points), here we only prove by using $\tensorsrht$. One can use similar strategy to obtain an input sparsity time version using $\tensorsketch$. 
We also point out that in a more general neural network architecture, the network width between different layers might differ, i.e., we are in fact dealing with the tensor product $u\otimes v$ where $u\in \R^{m_\ell}$ and $v\in \R^{m_{\ell-1}}$. Our sketching can be modified to handle this kind of inputs. Specifically, for $\tensorsketch$, it is defined by a pair of hash functions and sign functions, we can change their domain to handle different input dimensions. For $\tensorsrht$, it's more tricky, however, we note that the Hadamard matrix is merely for speeding up computation via FFT algorithm, hence we can differ the size of $D_1$ and $D_2$, and change the size of sampling matrix $P$ accordingly.
\end{remark}

\section{Spectral Properties of Over-parametrized Deep Neural Network}\label{sec:spectral_app}

In this section, we study the spectral structure of our Gram matrix and its connection to the multi-layer NTK matrix. We bound the minimum eigenvalue of the Gram matrix of the last layer at initialization. We also show that as long as the weight matrix does not move too far, the eigenvalue of the Gram matrix is relatively stable around its initialization.
\subsection{Bounds on the Least Eigenvalue of Kernel at Initialization}

The following fact gives the exact form of the Gram matrix of the $\ell$-th layer of the neural network.
\begin{fact}[Multi-layer Gram matrix]\label{fac:multilayer_gram}
For any $\ell\in [L]$, let $G_\ell = J_\ell J_\ell^\top\in \R^{n\times n}$ be the layer-$\ell$'s Gram matrix. Then, for any $i,j\in [n]$,
\begin{align*}
    (G_\ell)_{i,j} = h_{i,\ell-1}^\top h_{j,\ell-1} \cdot a^\top \left(D_{i,\ell}\prod_{r=\ell+1}^L W_r^\top D_{i,r}\right)^\top \left(D_{j,\ell}\prod_{r=\ell+1}^L W_r^\top D_{j,r}\right)a,
\end{align*}
where $h_{i, \ell-1} = D_{i,\ell-1}W_{\ell-1}\cdots D_{i,1}W_{1} x_i$. Here $J_\ell$ is the Jacobian with respect to $W_\ell$. For $\ell=L$, Lemmas~\ref{lem:init_f_J}, \ref{lem:perturb}, \ref{lem:init_min_eigen_L} and~\ref{lem:eigen_perturb} take it with respect to $\wt W_L=\sqrt{m/2}\,W_L$, which multiplies $G_L$ by $2/m$.
\end{fact}
\begin{proof}
For for any $i,j\in [n]$, by definition, 
\begin{align*}
    (G_\ell)_{i,j}=&~ \vect(\frac{\partial f(W,x_i)}{\partial W_\ell})^\top \vect(\frac{\partial f(W,x_j)}{\partial W_\ell})\\
    = &~ \vect\left(D_{i,\ell}\prod_{r=\ell+1}^L W_r^\top D_{i,r} a h_{i,\ell-1}^\top\right)^\top \vect\left(D_{j,\ell}\prod_{r=\ell+1}^L W_r^\top D_{j,r} a h_{j,\ell-1}^\top\right)\\
    = &~ \left((h_{i,\ell-1}\otimes I_{m_\ell})\left(D_{i,\ell}\prod_{r=\ell+1}^L W_r^\top D_{i,r}a\right)\right)^\top (h_{j,\ell-1}\otimes I_{m_\ell})\left(D_{j,\ell}\prod_{r=\ell+1}^L W_r^\top D_{j,r}a\right)\\
    = &~ \left(D_{i,\ell}\prod_{r=\ell+1}^L W_r^\top D_{i,r}a\right)^\top (h_{i,\ell-1}^\top \otimes I_{m_\ell}) (h_{j,\ell-1}\otimes I_{m_\ell})\left(D_{j,\ell}\prod_{r=\ell+1}^L W_r^\top D_{j,r}a\right)\\
    = &~ a^\top \left(D_{i,\ell}\prod_{r=\ell+1}^L W_r^\top D_{i,r}\right)^\top \left(D_{j,\ell}\prod_{r=\ell+1}^L W_r^\top D_{j,r}\right)a \cdot h_{i,\ell-1}^\top h_{j,\ell-1}.
 \end{align*}
\end{proof}

To bound the minimum eigenvalue of the Gram matrix of the last layer, we first define the NTK kernel for multi-layer neural networks. Throughout this section, as in Section~\ref{sec:modified_als}, $p$ and $\Phi$ are the density and the distribution function of ${\cal N}(0,1)$, $P(x):=1-\Phi(x)$, $P_b:=P(b)$, and
\begin{align*}
    G(x):=\E_{g\sim{\cal N}(0,1)}\big[g^2\,{\bf 1}[g>x]\big]=P(x)+x\,p(x),
\end{align*}
so that $c_b=1/(2G(b))$ and $2c_b/m=1/(mG(b))$. We also write $k:=m\exp(-b^2/2)$, as in Remark~\ref{remark:k}.

\begin{fact}[Gaussian quantities]\label{fact:gauss_quant}
For every $b>0$, (i) $P_b\leq G(b)$ and $P_b/G(b)\leq 1/(1+b^2)$, (ii) $G(b)\geq(1+b)p(b)/4$, so that $p(b)/G(b)\leq 4/(1+b)$, and (iii) $k/(3(1+b))\leq mP_b\leq\min\{k,\,mG(b)\}$, so that $1/(mG(b))\leq 3(1+b)/k$.
\end{fact}
\begin{proof}
(i) $G(b)=P_b+b\,p(b)\geq P_b$, and $G(b)/P_b=1+b\,p(b)/P_b\geq 1+b^2$ by the upper bound in~\eqref{eq:mills}. (ii) By the lower bound in~\eqref{eq:mills}, $G(b)\geq p(b)\big(\frac{1}{1+b}+b\big)$, and $4+4b(1+b)\geq(1+b)^2$. (iii) By the lower bound in~\eqref{eq:mills}, $mP_b\geq m\,p(b)/(1+b)=k/(\sqrt{2\pi}(1+b))$. By Fact~\ref{fact:gaussian_ulb}, $mP_b\leq m e^{-b^2/2}=k$, and $mP_b\leq mG(b)$ by (i).
\end{proof}

We write $\varphi_b(x):=x\,{\bf 1}[x>b]$. For an input of norm $1$, the threshold is $\sqrt{2/m}\,b$, and the activation of Definition~\ref{def:architecture} with this threshold is $\phi(x)=\sqrt{c_b}\sqrt{2/m}\;\varphi_b(\sqrt{m/2}\,x)$. For a $2\times 2$ PSD matrix $\Sigma$ and $(U,V)\sim{\cal N}(0,\Sigma)$, let
\begin{align*}
    \Psi(\Sigma):=\frac{\E[\varphi_b(U)\varphi_b(V)]}{G(b)},\qquad \Pi(\Sigma):=\Pr[U>b,~V>b].
\end{align*}
For $\rho\in[-1,1]$, let $\Sigma_\rho:=\begin{bmatrix} 1 & \rho\\ \rho & 1\end{bmatrix}$ and $\psi(\rho):=\Psi(\Sigma_\rho)$. So $\psi$ is the kernel map of one hidden layer as a function of the correlation of its two inputs. Since $\E[\varphi_b(U)^2]=G(b)$ for $U\sim{\cal N}(0,1)$, we have $\psi(1)=1$, and $\Pi(\Sigma_1)=P_b$.

\begin{fact}[One layer in kernel form]\label{fac:shifted_kernel}
Let $h,h'\in\R^{d'}$ be fixed nonzero vectors, let $\wh h:=h/\|h\|_2$, $\wh h':=h'/\|h'\|_2$ and $\rho:=\langle\wh h,\wh h'\rangle$, and let $W\in\R^{m\times d'}$ have i.i.d.\ ${\cal N}(0,\frac{2}{m})$ entries. Let $v,v'\in\R^m$ be the outputs of a hidden layer of Definition~\ref{def:architecture} with weight matrix $W$ on the inputs $h$ and $h'$, with thresholds $\sqrt{2/m}\,b\,\|h\|_2$ and $\sqrt{2/m}\,b\,\|h'\|_2$, and let $D,D'$ be the corresponding diagonal matrices of derivatives. Then the pairs $(U_r,V_r):=\sqrt{m/2}\,((W\wh h)_r,(W\wh h')_r)$, $r\in[m]$, are i.i.d.\ with distribution ${\cal N}(0,\Sigma_\rho)$, and
\begin{align*}
    v^\top v'=&~\frac{\|h\|_2\|h'\|_2}{mG(b)}\sum_{r\in[m]}\varphi_b(U_r)\varphi_b(V_r),\\
    \frac{2}{m}\,\tr[DD']=&~\frac{1}{mG(b)}\sum_{r\in[m]}{\bf 1}[U_r>b]\,{\bf 1}[V_r>b].
\end{align*}
In particular, $\E[v^\top v']=\|h\|_2\|h'\|_2\,\psi(\rho)$ and $\E[\frac{2}{m}\tr[DD']]=\Pi(\Sigma_\rho)/G(b)$.
\end{fact}
\begin{proof}
The rows $W_r$ of $W$ are i.i.d.\ ${\cal N}(0,\frac{2}{m}I_{d'})$, so the pairs $(\langle W_r,\wh h\rangle,\langle W_r,\wh h'\rangle)$ are i.i.d.\ ${\cal N}(0,\frac{2}{m}\Sigma_\rho)$. We have $(Wh)_r=\sqrt{2/m}\,\|h\|_2U_r$, and this exceeds the threshold $\sqrt{2/m}\,b\,\|h\|_2$ if and only if $U_r>b$. So coordinate $r$ of $v$ is $\sqrt{c_b}\sqrt{2/m}\,\|h\|_2\,\varphi_b(U_r)$, and $D_{r,r}=\sqrt{c_b}\,{\bf 1}[U_r>b]$. The same holds for $v'$ and $D'$. Both formulas follow from $2c_b/m=1/(mG(b))$.
\end{proof}

\begin{definition}[Multiple layer NTK kernel]\label{def:gram_multilayer}
The NTK kernels $\mathbf{K}_\ell \in \R^{n\times n}$, $\ell \in \{0,\dots, L\}$, of an $L$-layer neural network are defined as follows:
\begin{itemize}
    \item $(\mathbf{K}_0)_{i,j}:=x_i^\top x_j$.
    \item For $\ell\in[L]$ and $(i,j)\in [n]\times [n]$, let $\Sigma_{\ell, i,j}:=\begin{bmatrix}(\mathbf{K}_{\ell-1})_{i,i} & (\mathbf{K}_{\ell-1})_{i,j}\\ (\mathbf{K}_{\ell-1})_{j,i} & (\mathbf{K}_{\ell-1})_{j, j} \end{bmatrix}\in \R^{2\times 2}$. Then
    \begin{align*}
        (\mathbf{K}_{\ell})_{i,j}:= &~ \Psi(\Sigma_{\ell,i,j})=m\,\E_{(u,v)\sim \mathcal{N}(\mathbf{0}, \frac{2}{m}\Sigma_{\ell, i, j})}[\phi(u)\phi(v)]~~~\forall \ell\in [L-1],\\
        (\mathbf{K}_{L})_{i,j}:= &~ (\mathbf{K}_{L-1})_{i,j}\cdot\frac{\Pi(\Sigma_{L,i,j})}{G(b)}=(\mathbf{K}_{L-1})_{i,j}\cdot 2\,\E_{(u,v)\sim \mathcal{N}(\mathbf{0}, \frac{2}{m}\Sigma_{L, i, j})}[\phi'(u)\phi'(v)].
    \end{align*}
\end{itemize}
Let $\lambda_\ell := \lambda_{\min}(\mathbf{K}_\ell)$ be the minimum eigenvalue of the NTK kernel $\mathbf{K}_\ell$, and let $\lambda_*:=\min_{1\leq\ell\leq L-2}\lambda_\ell$ be the smallest eigenvalue among the kernels of the hidden layers $1,\dots,L-2$ (with $\lambda_*:=1$ if $L\leq 2$).
\end{definition}

The two forms in each line agree by Fact~\ref{fac:shifted_kernel}, applied to unit vectors, since we show below that $\mathbf{K}_{\ell-1}$ has unit diagonal for $\ell\leq L$. So $\Sigma_{\ell,i,j}=\Sigma_\rho$ with $\rho=(\mathbf{K}_{\ell-1})_{i,j}$, and
\begin{align*}
    (\mathbf{K}_\ell)_{i,j}=\psi\big((\mathbf{K}_{\ell-1})_{i,j}\big)~~~\forall\ell\in[L-1],\qquad (\mathbf{K}_L)_{i,j}=(\mathbf{K}_{L-1})_{i,j}\cdot\frac{\Pi(\Sigma_{(\mathbf{K}_{L-1})_{i,j}})}{G(b)}.
\end{align*}
These are the same kernels as for the fixed threshold $\sqrt{2/m}\,b$ of the published version, since the two networks agree on inputs of norm $1$ and the kernels have unit diagonal. The threshold of each layer scales with the standard deviation of its pre-activations, so for a fixed $b$ the kernels do not depend on $m$. For $b=\sqrt{2\gamma\log m}$, the kernels, and hence $\lambda_L$ and $\lambda_*$, depend on $m$ only through $b$. Since $\varphi_b$ is not positively homogeneous, they are not given by the arc-cosine formulas of the ReLU kernels. We record some properties. Since $\|x_i\|_2=1$ and $\psi(1)=1$, induction on $\ell$ gives $(\mathbf{K}_\ell)_{i,i}=1$ for $\ell\in\{0,\dots,L-1\}$. Each $\mathbf{K}_\ell$ with $\ell\in[L-1]$ is PSD, since $(\mathbf{K}_\ell)_{i,j}=\E[\varphi_b(U_i)\varphi_b(U_j)]/G(b)$ for a Gaussian vector $(U_1,\dots,U_n)\sim{\cal N}(0,\mathbf{K}_{\ell-1})$. Hence $|(\mathbf{K}_\ell)_{i,j}|\leq 1$ and $\lambda_\ell\leq 1$ for $\ell\leq L-1$, and $\lambda_*\in(0,1]$ whenever all these kernels are positive definite. If $\ell\in[L-2]$ and $i\neq j$, the eigenvalues $1\pm(\mathbf{K}_\ell)_{i,j}$ of the principal submatrix of $\mathbf{K}_\ell$ on $\{i,j\}$ are at least $\lambda_\ell$ by eigenvalue interlacing, so
\begin{align}\label{eq:corr_gap}
    1-|(\mathbf{K}_\ell)_{i,j}|\geq\lambda_*~~~\text{for all}~\ell\in[L-2]~\text{and}~i\neq j.
\end{align}
Finally $(\mathbf{K}_L)_{i,i}=P_b/G(b)\leq 1/(1+b^2)$ by Fact~\ref{fact:gauss_quant}. So $\lambda_L\leq 1/(1+b^2)$.

The comparison of the random Gram matrices with these kernels uses four tools. The first is a perturbation bound for matrix square roots.

\begin{lemma}[Perturbation of matrix square roots]\label{lem:sqrt_perturb}
Let $A,B$ be PSD matrices of the same size. Then $\|A^{1/2}-B^{1/2}\|\leq\|A-B\|^{1/2}$. If moreover $\lambda_{\min}(A)\geq\lambda>0$, then $\|A^{1/2}-B^{1/2}\|\leq\|A-B\|/\sqrt\lambda$.
\end{lemma}
\begin{proof}
Let $X:=A^{1/2}-B^{1/2}$, which is symmetric, and let $v$ be a unit eigenvector of $X$ whose eigenvalue $\mu$ satisfies $|\mu|=\|X\|$. Since $A-B=A^{1/2}X+XB^{1/2}$, we have
\begin{align*}
    v^\top(A-B)v=\mu\,\big(v^\top A^{1/2}v+v^\top B^{1/2}v\big),
\end{align*}
and both quadratic forms on the right are nonnegative. If $\lambda_{\min}(A)\geq\lambda$, the first one is at least $\sqrt\lambda$, so $|\mu|\sqrt\lambda\leq\|A-B\|$, which is the second bound. For the first bound, we may assume $\mu\geq 0$, after exchanging $A$ and $B$ (which replaces $X$ by $-X$). Then $v^\top A^{1/2}v=v^\top B^{1/2}v+\mu\geq\mu$, so $\mu^2\leq v^\top(A-B)v\leq\|A-B\|$.
\end{proof}

The second tool bounds how often a small perturbation of a Gaussian moves it across the threshold $b$. This is where the jump of $\varphi_b$ at $b$ enters.

\begin{lemma}[Threshold crossings]\label{lem:crossing}
Let $b\geq 0$, let $u\in\R^2$ be a unit vector, let $g\sim{\cal N}(0,I_2)$, and let $\eta>0$ satisfy $\eta(1+b^2)\leq 1/4$. Let ${\cal E}:=\{|\langle u,g\rangle-b|\leq\eta\|g\|_2\}$. Then
\begin{align*}
    \Pr[{\cal E}]\leq O(\eta)\cdot G(b),\qquad \E\big[\|g\|_2^2\,{\bf 1}_{\cal E}\big]\leq O\big(\eta(1+b)^2\big)\cdot G(b).
\end{align*}
\end{lemma}
\begin{proof}
Recall that $(1+b)p(b)\leq 4G(b)$ by Fact~\ref{fact:gauss_quant}(ii).

Let $g_1:=\langle u,g\rangle$, and let $g_2$ be the coordinate of $g$ along a unit vector orthogonal to $u$. Then $g_1,g_2$ are independent ${\cal N}(0,1)$ variables and $\|g\|_2^2=g_1^2+g_2^2$. The hypothesis gives $\eta\leq 1/4$ and $\eta b^2\leq 1/4$, hence also $\eta b=(\eta\cdot\eta b^2)^{1/2}\leq 1/4$. On ${\cal E}$, we have $|g_1|\leq b+\eta|g_1|+\eta|g_2|$, so $|g_1|\leq\frac{4}{3}(b+\eta|g_2|)\leq\frac{4}{3}b+\frac{1}{3}|g_2|$. Therefore, on ${\cal E}$,
\begin{align*}
    |g_1-b|\leq\eta(|g_1|+|g_2|)\leq 2\eta(b+|g_2|)=:w,\qquad \|g\|_2^2\leq\frac{32}{9}b^2+\frac{11}{9}g_2^2\leq 4(b+|g_2|)^2.
\end{align*}
Condition on $g_2$, which fixes $w$. For $|x-b|\leq w$, we have $x^2\geq\max\{b-w,0\}^2\geq b^2-2bw$, so $p(x)\leq p(b)\,e^{bw}$, and $bw=2\eta b^2+2\eta b|g_2|\leq\frac{1}{2}+\frac{1}{2}|g_2|$. Hence
\begin{align*}
    \Pr[{\cal E}\mid g_2]\leq\Pr\big[|g_1-b|\leq w\mid g_2\big]\leq 2w\,p(b)\,e^{bw}\leq 4e^{1/2}\,\eta\,(b+|g_2|)\,e^{|g_2|/2}\,p(b),
\end{align*}
and $\E[\|g\|_2^2{\bf 1}_{\cal E}\mid g_2]\leq 4(b+|g_2|)^2\Pr[{\cal E}\mid g_2]$. Since $b+|g_2|\leq(1+b)(1+|g_2|)$ and $\E[(1+|g_2|)^3e^{|g_2|/2}]=O(1)$, taking the expectation over $g_2$ gives $\Pr[{\cal E}]\leq O(\eta(1+b)p(b))$ and $\E[\|g\|_2^2{\bf 1}_{\cal E}]\leq O(\eta(1+b)^3p(b))$. The bound $(1+b)p(b)\leq 4G(b)$ completes the proof.
\end{proof}

The third tool is the continuity of the maps $\Psi$ and $\Pi$ at a covariance matrix with unit diagonal. Only the bound on $\Pi$ is used below, in the proof of Lemma~\ref{lem:init_min_eigen_L}.

\begin{lemma}[Continuity of the shifted kernel]\label{lem:kernel_cont}
Let $\Sigma,\wh\Sigma$ be $2\times 2$ PSD matrices with $\Sigma_{1,1}=\Sigma_{2,2}=1$ and $\|\wh\Sigma\|\leq 9/4$. Let $\eta:=\|\wh\Sigma^{1/2}-\Sigma^{1/2}\|$ and suppose $\eta(1+b^2)\leq 1/4$. Then
\begin{align*}
    |\Pi(\wh\Sigma)-\Pi(\Sigma)|\leq O(\eta)\cdot G(b),\qquad |\Psi(\wh\Sigma)-\Psi(\Sigma)|\leq O\big(\eta(1+b)^2\big).
\end{align*}
\end{lemma}
\begin{proof}
Let $g\sim{\cal N}(0,I_2)$, let $u,v$ be the rows of $\Sigma^{1/2}$, and let $\wh u,\wh v$ be the rows of $\wh\Sigma^{1/2}$. Then $(U,V):=(\langle u,g\rangle,\langle v,g\rangle)\sim{\cal N}(0,\Sigma)$ and $(\wh U,\wh V):=(\langle \wh u,g\rangle,\langle \wh v,g\rangle)\sim{\cal N}(0,\wh\Sigma)$, so we may compute $\Pi$ and $\Psi$ with this coupling. We have $\|u\|_2^2=\Sigma_{1,1}=1$, $\|v\|_2=1$, $\|\wh u\|_2,\|\wh v\|_2\leq\|\wh\Sigma^{1/2}\|\leq 3/2$, and $\|\wh u-u\|_2,\|\wh v-v\|_2\leq\eta$. Hence $|U|\leq\|g\|_2$, $|\wh U|\leq\frac{3}{2}\|g\|_2$ and $|\wh U-U|\leq\eta\|g\|_2$, and the same holds for $V$. Let ${\cal E}_U$ be the event ${\bf 1}[\wh U>b]\neq{\bf 1}[U>b]$. On ${\cal E}_U$, the threshold $b$ lies between $U$ and $\wh U$, so $|U-b|\leq|\wh U-U|\leq\eta\|g\|_2$. Thus ${\cal E}_U$ is contained in the event of Lemma~\ref{lem:crossing} for the unit vector $u$, and likewise ${\cal E}_V$ for $v$.

For $\Pi$, we have $|{\bf 1}[\wh U>b]{\bf 1}[\wh V>b]-{\bf 1}[U>b]{\bf 1}[V>b]|\leq{\bf 1}_{{\cal E}_U}+{\bf 1}_{{\cal E}_V}$, so $|\Pi(\wh\Sigma)-\Pi(\Sigma)|\leq\Pr[{\cal E}_U]+\Pr[{\cal E}_V]\leq O(\eta)\,G(b)$ by Lemma~\ref{lem:crossing}.

For $\Psi$, note that for real $x,\wh x$, if ${\bf 1}[\wh x>b]={\bf 1}[x>b]$ then $|\varphi_b(\wh x)-\varphi_b(x)|\leq|\wh x-x|\,{\bf 1}[x>b]$, and otherwise $|\varphi_b(\wh x)-\varphi_b(x)|\leq\max\{|x|,|\wh x|\}$. Hence
\begin{align*}
    |\varphi_b(\wh U)-\varphi_b(U)|\leq\eta\|g\|_2{\bf 1}[U>b]+\tfrac{3}{2}\|g\|_2{\bf 1}_{{\cal E}_U},
\end{align*}
the same bound holds for $V$, $|\varphi_b(U)|\leq\|g\|_2{\bf 1}[U>b]$, and, since $\eta\leq 1/4$, $|\varphi_b(\wh V)|\leq\frac{5}{4}\|g\|_2{\bf 1}[V>b]+\frac{3}{2}\|g\|_2{\bf 1}_{{\cal E}_V}$. We write $\varphi_b(\wh U)\varphi_b(\wh V)-\varphi_b(U)\varphi_b(V)=(\varphi_b(\wh U)-\varphi_b(U))\varphi_b(\wh V)+\varphi_b(U)(\varphi_b(\wh V)-\varphi_b(V))$, multiply out, and bound every indicator except one by $1$ in each product. With $\eta\leq 1/4$, this gives
\begin{align*}
    |\varphi_b(\wh U)\varphi_b(\wh V)-\varphi_b(U)\varphi_b(V)|\leq\|g\|_2^2\Big(\tfrac{9}{4}\,\eta\,{\bf 1}[U>b]+5\cdot{\bf 1}_{{\cal E}_U}+2\cdot{\bf 1}_{{\cal E}_V}\Big).
\end{align*}
Let $g_1:=U$ and let $g_2$ be the coordinate of $g$ orthogonal to $u$. Then $\E[\|g\|_2^2{\bf 1}[U>b]]=\E[(g_1^2+g_2^2){\bf 1}[g_1>b]]=G(b)+P(b)\leq 2G(b)$. By Lemma~\ref{lem:crossing}, $\E[\|g\|_2^2{\bf 1}_{{\cal E}_U}]$ and $\E[\|g\|_2^2{\bf 1}_{{\cal E}_V}]$ are at most $O(\eta(1+b)^2)\,G(b)$. Taking expectations and dividing by $G(b)$ gives the bound on $\Psi$.
\end{proof}

The fourth tool is a structural property of the kernel map $\psi$ of a hidden layer. Its power-series form is the dual-activation expansion of Daniely, Frostig and Singer~\cite{dfs16}. The proof uses the standard fact that the normalized Hermite polynomials form an orthonormal basis of $L^2({\cal N}(0,1))$ (see, for example, Chapter~11 of~\cite{o14}). The property takes the place of a Lipschitz bound for $\psi$. No such bound holds uniformly, because the jump of $\varphi_b$ at $b$ makes the slope of $\psi$ unbounded near the correlation $1$.

\begin{lemma}[The kernel map of a hidden layer]\label{lem:dual}
Let $b>0$. There are coefficients $c_0,c_1,\dots\geq 0$ with $\sum_{q\geq 0}c_q=1$ such that $\psi(\rho)=\sum_{q\geq 0}c_q\rho^q$ for all $\rho\in[-1,1]$. Consequently:
\begin{itemize}
\item[(a)] $|\psi(\rho)|\leq\psi(|\rho|)\leq 1$ for $\rho\in[-1,1]$, and $\psi$ is nondecreasing on $[0,1]$.
\item[(b)] $\psi$ is differentiable on $(-1,1)$, $|\psi'(\rho)|\leq\psi'(|\rho|)$ for $\rho\in(-1,1)$, and $\psi'$ is nondecreasing on $[0,1)$.
\item[(c)] $\psi'(s)\leq\frac{1-\psi(s)}{1-s}$ for every $s\in[0,1)$.
\end{itemize}
\end{lemma}
\begin{proof}
Let $\mathrm{He}_q$ be the probabilists' Hermite polynomials, defined by the generating function
\begin{align*}
    e^{tx-t^2/2}=\sum_{q\geq 0}\mathrm{He}_q(x)\,\frac{t^q}{q!}.
\end{align*}
The functions $\mathrm{He}_q/\sqrt{q!}$ form an orthonormal basis of $L^2({\cal N}(0,1))$. For $(U,V)\sim{\cal N}(0,\Sigma_\rho)$ and $s,t\in\R$, we have $\E[e^{sU-s^2/2}e^{tV-t^2/2}]=e^{\rho st}$, since $sU+tV\sim{\cal N}(0,s^2+t^2+2\rho st)$. We differentiate both sides $q$ times in $s$ and $q'$ times in $t$ at $s=t=0$. On the left we may differentiate under the expectation, since $U$ and $V$ have exponential moments. This gives $\E[\mathrm{He}_q(U)\mathrm{He}_{q'}(V)]=q!\,\rho^q$ if $q=q'$ and $0$ otherwise.

The function $\varphi_b$ lies in $L^2({\cal N}(0,1))$, with $\E[\varphi_b(U)^2]=G(b)$. Write $\varphi_b=\sum_q a_q\mathrm{He}_q/\sqrt{q!}$ in this basis, so that $\sum_qa_q^2=G(b)$, and let $f_N:=\sum_{q\leq N}a_q\mathrm{He}_q/\sqrt{q!}$. By the previous paragraph, $\E[f_N(U)f_N(V)]=\sum_{q\leq N}a_q^2\rho^q$. Since $U$ and $V$ are both standard Gaussians, the Cauchy--Schwarz inequality gives
\begin{align*}
    \big|\E[\varphi_b(U)\varphi_b(V)]-\E[f_N(U)f_N(V)]\big|\leq\|\varphi_b-f_N\|\,\|\varphi_b\|+\|f_N\|\,\|\varphi_b-f_N\|,
\end{align*}
where $\|\cdot\|$ is the norm of $L^2({\cal N}(0,1))$. The right side tends to $0$ as $N\to\infty$. Hence $\psi(\rho)=\sum_qc_q\rho^q$ with $c_q:=a_q^2/G(b)\geq 0$, and $\sum_qc_q=\psi(1)=1$.

Parts (a) and (b) follow from the nonnegativity of the coefficients. The series converges absolutely on $[-1,1]$, so $\psi$ is continuous there and $|\psi(\rho)|\leq\sum_qc_q|\rho|^q=\psi(|\rho|)$. It can be differentiated term by term on $(-1,1)$, and $|\sum_qqc_q\rho^{q-1}|\leq\sum_qqc_q|\rho|^{q-1}$. Every term of $\psi$ and of $\psi'$ is nondecreasing on $[0,1)$. For (c), let $s\in[0,1)$. Since $\psi'$ is nondecreasing on $[s,1)$ and $\psi$ is continuous at $1$,
\begin{align*}
    1-\psi(s)=\psi(1)-\psi(s)=\lim_{u\to 1^-}\int_s^u\psi'(x)\,\d x\geq(1-s)\,\psi'(s).
\end{align*}
\end{proof}

In the following lemma, we generalize Lemma C.3 in \cite{bpsw21} (also Lemma 3 in \cite{cgh+19}) into multiple layer neural networks. For $\ell\in\{0,\dots,L-1\}$, let $H_\ell\in\R^{n\times n}$ be given by $(H_\ell)_{i,j}:=h_{i,\ell}^\top h_{j,\ell}$, where $h_{i,0}:=x_i$, so that $H_0=X^\top X=\mathbf{K}_0$. When $h_{i,\ell}\neq 0$, let $\wh h_{i,\ell}:=h_{i,\ell}/\|h_{i,\ell}\|_2$, and let $\wh\rho_\ell(i,j):=\langle\wh h_{i,\ell},\wh h_{j,\ell}\rangle$ be the correlation of the hidden vectors of $x_i$ and $x_j$.

\begin{lemma}\label{lem:init_min_eigen_H}
Let $\delta\in(0,1)$, suppose $\lambda_*>0$ (Definition~\ref{def:gram_multilayer}), and let
\begin{align*}
    \epsilon_H:=C_H\sqrt{\frac{1+b}{k}}\,\log\frac{mnL}{\delta},
\end{align*}
where $k=m\exp(-b^2/2)$ is the sparsity parameter of Remark~\ref{remark:k} and $C_H$ is a large enough absolute constant. Suppose that
\begin{align*}
    12L^2\epsilon_H\leq\lambda_*^2.
\end{align*}
Then, with probability at least $1-\delta/2$ over $W_1,\dots,W_{L-1}$, we have for every $\ell\in\{0,1,\dots,L-1\}$:
\begin{itemize}
\item[(i)] $\|h_{i,\ell}\|_2^2\in[1-\ell\epsilon_H,1+2\ell\epsilon_H]\subseteq[0.9,1.1]$ for all $i\in[n]$,
\item[(ii)] $|\wh\rho_\ell(i,j)-(\mathbf{K}_\ell)_{i,j}|\leq 6\ell\epsilon_H/\lambda_*$ for all $i\neq j$,
\item[(iii)] $\max_{(i,j)\in[n]\times[n]}\big|(H_\ell)_{i,j}-(\mathbf{K}_\ell)_{i,j}\big|\leq e_\ell:=8\ell\epsilon_H/\lambda_*$.
\end{itemize}
\end{lemma}

\begin{proof}
We follow the induction on $\ell$ of \cite{dllwz19}, with two changes. First, the concentration at each layer has to use the sparsity of the activations. Second, we propagate the error from the previous layer with Lemma~\ref{lem:dual} instead of a Lipschitz bound for the kernel map. Since the layers are positively homogeneous, we can treat the norms of the hidden vectors and their correlations $\wh\rho_\ell(i,j)$ separately. The hypothesis gives $\epsilon_H\leq 1/(12L^2)$, so $\ell\epsilon_H\leq 1/(12L)\leq 1/24$ for $\ell\leq L$, as $L\geq 2$.

\textbf{Step 1: concentration at one layer.} Fix $\ell\in[L-1]$ and condition on $W_1,\dots,W_{\ell-1}$ such that $h_{i,\ell-1}\neq 0$ for all $i\in[n]$. For $i,j\in[n]$, including $i=j$, let
\begin{align*}
    S_\ell(i,j):=\frac{(H_\ell)_{i,j}}{\|h_{i,\ell-1}\|_2\,\|h_{j,\ell-1}\|_2}.
\end{align*}
We show that, with probability at least $1-\delta/(2L)$ over $W_\ell$,
\begin{align}\label{eq:H_one_layer}
    \max_{(i,j)\in[n]\times[n]}\big|S_\ell(i,j)-\psi(\wh\rho_{\ell-1}(i,j))\big|\leq\epsilon_H,
\end{align}
where $\wh\rho_{\ell-1}(i,i)=1$. Fix $i,j$. By Fact~\ref{fac:shifted_kernel} with $W=W_\ell$, we have $S_\ell(i,j)=\sum_{r\in[m]}Z_r$ with
\begin{align*}
    Z_r:=\frac{\varphi_b(U_r)\varphi_b(V_r)}{mG(b)}\geq 0,
\end{align*}
where the pairs $(U_r,V_r)$ are i.i.d.\ ${\cal N}(0,\Sigma_\rho)$ with $\rho:=\wh\rho_{\ell-1}(i,j)$, and $\E[\sum_rZ_r]=\psi(\rho)\leq 1$ by Lemma~\ref{lem:dual}(a). A summand is nonzero only if neuron $r$ fires on both inputs. A nonzero summand can be as large as order $\log(m)/(mG(b))$, which is about $\sqrt{\log m}/k$ for $b=\sqrt{2\gamma\log m}$. A Hoeffding bound with this range over all $m$ summands would give a deviation of order $\sqrt m/k$ up to logarithmic factors. We therefore use Bernstein's inequality, whose variance term accounts for the sparsity. Let
\begin{align*}
    \tau:=2\sqrt{\log(24m^2nL/\delta)},\qquad \wt Z_r:=Z_r\,{\bf 1}\big[|U_r|\leq\tau,~|V_r|\leq\tau\big].
\end{align*}
Then $0\leq\wt Z_r\leq B:=\tau^2/(mG(b))\leq 3\tau^2(1+b)/k$ by Fact~\ref{fact:gauss_quant}, and
\begin{align*}
    \sum_{r\in[m]}\E[\wt Z_r^2]\leq B\sum_{r\in[m]}\E[\wt Z_r]\leq B\sum_{r\in[m]}\E[Z_r]=B\,\psi(\rho)\leq B.
\end{align*}
Applying Lemma~\ref{lem:bernstein} to the variables $\wt Z_r-\E[\wt Z_r]$ and to their negatives, we get for $t\in(0,1]$
\begin{align*}
    \Pr\Big[\Big|\sum_{r\in[m]}(\wt Z_r-\E[\wt Z_r])\Big|>t\Big]\leq 2\exp\Big(-\frac{t^2/2}{B+Bt/3}\Big)\leq 2\exp\Big(-\frac{t^2}{3B}\Big).
\end{align*}
We take $t:=(3B\log(8n^2L/\delta))^{1/2}$, so that this probability is at most $\delta/(4n^2L)$.

Next, we bound the effect of the truncation. For every $i'\in[n]$ and $r\in[m]$, the variable $\sqrt{m/2}\,(W_\ell\wh h_{i',\ell-1})_r$ is a standard Gaussian, so by Fact~\ref{fact:gaussian_ulb}, the event that one of these $mn$ variables exceeds $\tau$ in absolute value has probability at most $2mn\,e^{-\tau^2/2}\leq\delta/(4L)$. Outside this event, $Z_r=\wt Z_r$ for all $r$ and all pairs $i,j$. The truncation also changes the mean only slightly. Since $\E[Z_r^2]\leq\E[U_r^2V_r^2]/(mG(b))^2\leq 3/(mG(b))^2$ and the truncation is active with probability at most $4e^{-\tau^2/2}$, the Cauchy--Schwarz inequality and Fact~\ref{fact:gauss_quant} give
\begin{align*}
    0\leq\sum_{r\in[m]}\big(\E[Z_r]-\E[\wt Z_r]\big)\leq\frac{4e^{-\tau^2/4}}{G(b)}\leq\frac{12(1+b)m}{k}\cdot\frac{\delta}{24m^2nL}\leq\frac{1+b}{m}.
\end{align*}
Hence, outside the two exceptional events, $|S_\ell(i,j)-\psi(\rho)|\leq t+(1+b)/m$. Since $\tau^2$ and $\log(8n^2L/\delta)$ are $O(\log(mnL/\delta))$, we have $t\leq O\big(\sqrt{(1+b)/k}\,\log(mnL/\delta)\big)$. The hypothesis gives $\epsilon_H\leq 1$, so $k\geq C_H^2(1+b)\log^2(mnL/\delta)\geq 1+b$, and with $k\leq m$ we get $(1+b)/m\leq\sqrt{(1+b)/k}$. So $t+(1+b)/m\leq\epsilon_H$ once $C_H$ is a large enough absolute constant, which also justifies $t\leq 1$. A union bound over the $n^2$ pairs and the truncation event proves~\eqref{eq:H_one_layer} with probability at least $1-\delta/(2L)$.

\textbf{Step 2: norms and correlations at one layer.} Suppose~\eqref{eq:H_one_layer} holds. Since $\psi(1)=1$, we have $S_\ell(i,i)\in[1-\epsilon_H,1+\epsilon_H]$, and $\|h_{i,\ell}\|_2^2=\|h_{i,\ell-1}\|_2^2\,S_\ell(i,i)$. In particular $h_{i,\ell}\neq 0$. Moreover $\wh\rho_\ell(i,j)=S_\ell(i,j)/(S_\ell(i,i)S_\ell(j,j))^{1/2}$, and $|S_\ell(i,j)|\leq(S_\ell(i,i)S_\ell(j,j))^{1/2}$ by the Cauchy--Schwarz inequality. Since $(S_\ell(i,i)S_\ell(j,j))^{1/2}\in[1-\epsilon_H,1+\epsilon_H]$,
\begin{align}\label{eq:corr_one_layer}
    \big|\wh\rho_\ell(i,j)-\psi(\wh\rho_{\ell-1}(i,j))\big|\leq\big|1-(S_\ell(i,i)S_\ell(j,j))^{1/2}\big|+\epsilon_H\leq 2\epsilon_H.
\end{align}

\textbf{Step 3: induction over the layers.} For $\ell\in[L-1]$, let ${\cal B}_\ell$ be the event that $h_{i,\ell-1}\neq 0$ for all $i$ and~\eqref{eq:H_one_layer} fails. By Step 1, $\Pr[{\cal B}_\ell]\leq\delta/(2L)$, so none of ${\cal B}_1,\dots,{\cal B}_{L-1}$ occurs with probability at least $1-\delta/2$. Suppose that none occurs. Since $h_{i,0}=x_i\neq 0$, Step 2 and induction on $\ell$ show that $h_{i,\ell}\neq 0$ and that~\eqref{eq:H_one_layer} and~\eqref{eq:corr_one_layer} hold for every $\ell\in[L-1]$.

\emph{Norms.} By Step 2, $\|h_{i,\ell}\|_2^2\in[(1-\epsilon_H)^\ell,(1+\epsilon_H)^\ell]\subseteq[1-\ell\epsilon_H,1+2\ell\epsilon_H]$, since $(1+\epsilon_H)^\ell\leq e^{\ell\epsilon_H}\leq 1+2\ell\epsilon_H$ for $\ell\epsilon_H\leq 1$. With $\ell\epsilon_H\leq 1/24$ this proves (i).

\emph{Correlations.} Fix $i\neq j$, and write $\rho_\ell:=(\mathbf{K}_\ell)_{i,j}$, $\wh\rho_\ell:=\wh\rho_\ell(i,j)$, $d_\ell:=1-|\rho_\ell|$ and $e'_\ell:=|\wh\rho_\ell-\rho_\ell|$. Then $e'_0=0$ since $\wh\rho_0=x_i^\top x_j=\rho_0$, and for $\ell\in[L-1]$, $\rho_\ell=\psi(\rho_{\ell-1})$, so by Lemma~\ref{lem:dual}(a),
\begin{align}\label{eq:gap_step}
    d_\ell=1-|\psi(\rho_{\ell-1})|\geq 1-\psi(|\rho_{\ell-1}|).
\end{align}
By~\eqref{eq:corr_gap}, $d_\ell\geq\lambda_*$ for $\ell\in[L-2]$, and $d_\ell\leq 1$ always. We claim that for $\ell\in[L-1]$,
\begin{align}\label{eq:corr_claim}
    e'_\ell\leq 2\epsilon_H\Big(1+\sum_{q=1}^{\ell-1}\Big(1+\frac{1}{L}\Big)^{\ell-q}\frac{d_\ell}{d_q}\Big)\leq\frac{6\ell\epsilon_H}{\lambda_*}.
\end{align}
The second inequality holds since for $q\leq\ell-1\leq L-2$ we have $d_q\geq\lambda_*$, $d_\ell\leq 1$ and $(1+1/L)^{L}\leq e<3$, and since $1\leq 3/\lambda_*$. For $\ell=1$, \eqref{eq:corr_one_layer} gives $e'_1=|\wh\rho_1-\psi(\rho_0)|\leq 2\epsilon_H$, which is~\eqref{eq:corr_claim}. Let $2\leq\ell\leq L-1$ and suppose~\eqref{eq:corr_claim} holds for $\ell-1$. Since $1\leq\ell-1\leq L-2$, we have $d_{\ell-1}\geq\lambda_*$, and the hypothesis $12L^2\epsilon_H\leq\lambda_*^2$ gives
\begin{align*}
    e'_{\ell-1}\leq\frac{6L\epsilon_H}{\lambda_*}\leq\frac{\lambda_*}{2L}\leq\frac{d_{\ell-1}}{2L}.
\end{align*}
Let $s^+:=|\rho_{\ell-1}|+e'_{\ell-1}$, so that $1-s^+\geq d_{\ell-1}(1-\frac{1}{2L})>0$. Every $x$ between $\rho_{\ell-1}$ and $\wh\rho_{\ell-1}$ satisfies $|x|\leq s^+<1$. So by parts (b), (c) and (a) of Lemma~\ref{lem:dual}, in this order, and by~\eqref{eq:gap_step},
\begin{align*}
    |\psi'(x)|\leq\psi'(s^+)\leq\frac{1-\psi(s^+)}{1-s^+}\leq\frac{1-\psi(|\rho_{\ell-1}|)}{d_{\ell-1}(1-\frac{1}{2L})}\leq\Big(1+\frac{1}{L}\Big)\frac{d_\ell}{d_{\ell-1}},
\end{align*}
where we used $(1-\frac{1}{2L})^{-1}\leq 1+\frac{1}{L}$. By the mean value theorem and~\eqref{eq:corr_one_layer},
\begin{align*}
    e'_\ell\leq|\wh\rho_\ell-\psi(\wh\rho_{\ell-1})|+|\psi(\wh\rho_{\ell-1})-\psi(\rho_{\ell-1})|\leq 2\epsilon_H+\Big(1+\frac{1}{L}\Big)\frac{d_\ell}{d_{\ell-1}}\,e'_{\ell-1}.
\end{align*}
Inserting~\eqref{eq:corr_claim} for $\ell-1$ gives~\eqref{eq:corr_claim} for $\ell$. This proves (ii).

\emph{Entries of $H_\ell$.} We have $(H_\ell)_{i,j}=\|h_{i,\ell}\|_2\|h_{j,\ell}\|_2\,\wh\rho_\ell(i,j)$, and by the bound on the norms, $\|h_{i,\ell}\|_2\|h_{j,\ell}\|_2\in[(1-\epsilon_H)^\ell,(1+\epsilon_H)^\ell]\subseteq[1-\ell\epsilon_H,1+2\ell\epsilon_H]$. For $i\neq j$, using $|\wh\rho_\ell(i,j)|\leq 1$ and (ii), $|(H_\ell)_{i,j}-(\mathbf{K}_\ell)_{i,j}|\leq 2\ell\epsilon_H+6\ell\epsilon_H/\lambda_*\leq e_\ell$. For $i=j$, $|(H_\ell)_{i,i}-1|\leq 2\ell\epsilon_H\leq e_\ell$. This proves (iii).
\end{proof}

\begin{remark}\label{remark:rate_H}
For $b=\sqrt{2\gamma\log m}$ with a constant $\gamma\in(0,1)$, so that $k=m^{1-\gamma}$, the rate $\epsilon_H=\wt O(k^{-1/2})$ of Lemma~\ref{lem:init_min_eigen_H} cannot be replaced by $\wt O(m^{-1/2})$, because only about $mP_b$ of the $m$ neurons of a layer fire. For instance, at layer $1$ and $i=j$, Fact~\ref{fac:shifted_kernel} gives $(H_1)_{i,i}=\sum_{r\in[m]}\varphi_b(U_r)^2/(mG(b))$ with i.i.d.\ $U_r\sim{\cal N}(0,1)$. Since $\E[\varphi_b(U)^4]=b^3p(b)+3G(b)\geq 3G(b)$ and $G(b)\leq 1$, its variance is
\begin{align*}
    \frac{\E[\varphi_b(U)^4]}{mG(b)^2}-\frac{1}{m}\geq\frac{2}{mG(b)}\geq\Omega\Big(\frac{1}{(1+b)k}\Big),
\end{align*}
where the last step uses $mG(b)\leq O((1+b)k)$. This holds since $G(b)\leq p(b)(b+1/b)\leq 2b\,p(b)$ and $m\,p(b)=k/\sqrt{2\pi}$ when $b\geq 1$, and since $mG(b)\leq m\leq 2k$ when $b<1$. So the fluctuation is of order at least $((1+b)k)^{-1/2}$, while $\epsilon_H$ is of order $((1+b)/k)^{1/2}$ up to the logarithmic factor.

The propagation over the layers costs the factor $\ell/\lambda_*$ in parts (ii) and (iii). The factor $\ell$ reflects that the errors of the layers add up. The factor $1/\lambda_*$ comes from pairs of data points whose kernel correlation $\rho$ at some hidden layer is close to $1$. For $\ell\in[L-1]$, the entries of $\mathbf{K}_\ell$ lie in $[0,1]$, since $\varphi_b\geq 0$. The kernel map $\psi$ pushes the correlations of such pairs away from $1$ and is steep there, so it amplifies the errors of such pairs. The proof bounds this amplification by the factor $1/(1-|\rho|)$. We do not know whether this factor is needed. Part (c) of Lemma~\ref{lem:dual} keeps the dependence on the depth polynomial. A Lipschitz bound $\Gamma$ for $\psi$ on the correlations that occur would only give an amplification by $\Gamma$ per layer, which is exponential in the depth whenever $\Gamma>1$.
\end{remark}

\begin{lemma}[Bounds on the least eigenvalue at initialization for layer $L$]\label{lem:init_min_eigen_L}
Let $\delta\in(0,1)$, suppose $\lambda_L>0$ and $\lambda_*>0$, and let $\epsilon_H$ be as in Lemma~\ref{lem:init_min_eigen_H}. As in Lemmas~\ref{lem:init_f_J} and~\ref{lem:perturb}, we write $W_L=\sqrt{2/m}\,\wt W_L$, where $\wt W_L$ has i.i.d.\ ${\cal N}(0,1)$ entries, and take the Jacobian $J_L$ of the last layer, and $G_L=J_LJ_L^\top$, with respect to $\wt W_L$ at initialization. There is an absolute constant $c_K>0$ such that, if
\begin{align}\label{eq:kernel_width_delta}
    \epsilon_H\,L^2n^2\lambda_L^{-2}\lambda_*^{-2}\leq c_K,\qquad\text{equivalently}\qquad k\geq C_K\,(1+b)\,L^4n^4\lambda_L^{-4}\lambda_*^{-4}\log^2\frac{mnL}{\delta}
\end{align}
with $C_K:=(C_H/c_K)^2$, then
\begin{align*}
    \Pr\Big[ \lambda_{\min}(G_L)\geq \frac{3}{4}\lambda_L \Big] \geq 1 - \delta.
\end{align*}
\end{lemma}

\begin{proof}
Let $C_\kappa$ be an absolute constant for which both bounds of Lemma~\ref{lem:kernel_cont} hold, let $c':=1/(64(1+C_\kappa)^2)$, and let $c_K:=c'/8$. Then $c_K\leq 1/12$, and since $n\geq 1$ and $\lambda_L\leq 1$, condition~\eqref{eq:kernel_width_delta} implies $12L^2\epsilon_H\leq\lambda_*^2$, so Lemma~\ref{lem:init_min_eigen_H} applies. Let ${\cal E}_H$ be its event, which has probability at least $1-\delta/2$ and depends only on $W_1,\dots,W_{L-1}$. Write $h_i:=h_{i,L-1}$, $\wh H:=H_{L-1}$, $\wh\rho_{i,j}:=\wh\rho_{L-1}(i,j)$ for $i\neq j$, $\wh\rho_{i,i}:=1$, and $e:=8L\epsilon_H/\lambda_*$. On ${\cal E}_H$, we have $\|h_i\|_2^2\in[0.9,1.1]$, $|\wh H_{i,j}|\leq 1.1$, and, by parts (ii) and (iii) of Lemma~\ref{lem:init_min_eigen_H}, both $|\wh H_{i,j}-(\mathbf{K}_{L-1})_{i,j}|$ and $|\wh\rho_{i,j}-(\mathbf{K}_{L-1})_{i,j}|$ are at most $e$ for all $i,j$. By~\eqref{eq:kernel_width_delta} and $\lambda_*\leq 1$,
\begin{align}\label{eq:L_err_small}
    \epsilon_H\leq e\leq\frac{8c_K\lambda_*\lambda_L^2}{Ln^2}\leq\frac{c'\lambda_L^2}{n^2}.
\end{align}

Row $i$ of $J_L$ is $\vect(\sqrt{2/m}\,D_{i,L}ah_i^\top)^\top$, and $a_r^2=1$, so
\begin{align*}
    (G_L)_{i,j}=\frac{2}{m}\,a^\top D_{i,L}D_{j,L}a\cdot h_i^\top h_j=\wh H_{i,j}\cdot\frac{2}{m}\,\tr[D_{i,L}D_{j,L}].
\end{align*}
Condition on $W_1,\dots,W_{L-1}$ in ${\cal E}_H$. By Fact~\ref{fac:shifted_kernel} with $W=W_L$, we have
\begin{align*}
    (G_L)_{i,j}=\wh H_{i,j}\sum_{r\in[m]}Y_r,\qquad Y_r:=\frac{{\bf 1}[U_r>b]\,{\bf 1}[V_r>b]}{mG(b)},
\end{align*}
where the pairs $(U_r,V_r)$ are i.i.d.\ ${\cal N}(0,\Sigma_{\wh\rho_{i,j}})$, and $\E[\sum_rY_r]=\Pi(\Sigma_{\wh\rho_{i,j}})/G(b)$. Let $(\wh{\mathbf{K}}_L)_{i,j}:=\wh H_{i,j}\,\Pi(\Sigma_{\wh\rho_{i,j}})/G(b)$. Then
\begin{align*}
    \lambda_{\min}(G_L)\geq\lambda_{\min}(\mathbf{K}_L)-\|G_L-\wh{\mathbf{K}}_L\|-\|\wh{\mathbf{K}}_L-\mathbf{K}_L\|,
\end{align*}
and we bound the last two terms by $\lambda_L/8$ each.

\textbf{Step 1: $\|G_L-\wh{\mathbf{K}}_L\|\leq\lambda_L/8$ with probability at least $1-\delta/2$ over $W_L$.} We have $0\leq Y_r\leq B_L:=1/(mG(b))\leq 3(1+b)/k$ by Fact~\ref{fact:gauss_quant}. Moreover, $\Pi(\Sigma_{\wh\rho_{i,j}})\leq P_b\leq G(b)$, so $\sum_r\E[Y_r^2]\leq B_L\sum_r\E[Y_r]\leq B_L$. Lemma~\ref{lem:bernstein}, applied to $Y_r-\E[Y_r]$ and to their negatives, gives for $t\in(0,1]$
\begin{align*}
    \Pr\Big[\Big|\sum_{r\in[m]}(Y_r-\E[Y_r])\Big|>t\Big]\leq 2\exp\Big(-\frac{t^2/2}{B_L+B_Lt/3}\Big)\leq 2\exp\Big(-\frac{t^2}{3B_L}\Big).
\end{align*}
We take $t_L:=(3B_L\log(4n^2/\delta))^{1/2}\leq O\big(\sqrt{(1+b)/k}\,\log(mnL/\delta)\big)\leq\epsilon_H\leq 1$, for $C_H$ large enough. Then each pair fails with probability at most $\delta/(2n^2)$, and by a union bound, with probability at least $1-\delta/2$, every entry satisfies $|(G_L)_{i,j}-(\wh{\mathbf{K}}_L)_{i,j}|\leq|\wh H_{i,j}|\,t_L\leq 1.1\,\epsilon_H$. By~\eqref{eq:L_err_small} and $c'\leq 1/64$,
\begin{align*}
    \|G_L-\wh{\mathbf{K}}_L\|\leq\|G_L-\wh{\mathbf{K}}_L\|_F\leq 1.1\,n\,\epsilon_H\leq 1.1\,c'\lambda_L^2/n\leq\lambda_L/8.
\end{align*}

\textbf{Step 2: $\|\wh{\mathbf{K}}_L-\mathbf{K}_L\|\leq\lambda_L/8$ on ${\cal E}_H$.} Fix $i,j$ and let $\rho:=(\mathbf{K}_{L-1})_{i,j}$, so that $|\rho|\leq 1$ and $(\mathbf{K}_L)_{i,j}=\rho\,\Pi(\Sigma_\rho)/G(b)$. We have $\|\Sigma_{\wh\rho_{i,j}}-\Sigma_\rho\|=|\wh\rho_{i,j}-\rho|\leq e$. The first bound of Lemma~\ref{lem:sqrt_perturb} needs no eigenvalue assumption, in particular not $\lambda_{L-1}>0$, and gives $\eta:=\|\Sigma_{\wh\rho_{i,j}}^{1/2}-\Sigma_\rho^{1/2}\|\leq e^{1/2}$. By~\eqref{eq:L_err_small} and $\lambda_L\leq 1/(1+b^2)$, we get $\eta(1+b^2)\leq\sqrt{c'}\,\lambda_L(1+b^2)/n\leq\sqrt{c'}\leq 1/4$. Also $\|\Sigma_{\wh\rho_{i,j}}\|\leq 2\leq 9/4$. Lemma~\ref{lem:kernel_cont} gives $|\Pi(\Sigma_{\wh\rho_{i,j}})-\Pi(\Sigma_\rho)|\leq C_\kappa\,\eta\,G(b)$. Using $\Pi(\Sigma_{\wh\rho_{i,j}})\leq G(b)$, we get
\begin{align*}
    |(\wh{\mathbf{K}}_L)_{i,j}-(\mathbf{K}_L)_{i,j}|\leq&~ \big|\wh H_{i,j}-\rho\big|\cdot\frac{\Pi(\Sigma_{\wh\rho_{i,j}})}{G(b)}+|\rho|\cdot\frac{|\Pi(\Sigma_{\wh\rho_{i,j}})-\Pi(\Sigma_\rho)|}{G(b)}\\
    \leq&~ e+C_\kappa e^{1/2}\leq (1+C_\kappa)\,e^{1/2},
\end{align*}
since $e\leq 1$. Hence, by~\eqref{eq:L_err_small}, $\|\wh{\mathbf{K}}_L-\mathbf{K}_L\|\leq n(1+C_\kappa)e^{1/2}\leq(1+C_\kappa)\sqrt{c'}\,\lambda_L=\lambda_L/8$.

On ${\cal E}_H$ and the event of Step 1, which together have probability at least $1-\delta$, we get $\lambda_{\min}(G_L)\geq\lambda_L-\lambda_L/8-\lambda_L/8=\frac{3}{4}\lambda_L$. The equivalence of the two forms of~\eqref{eq:kernel_width_delta} follows from the definition of $\epsilon_H$.
\end{proof}
Only Step 2 of this proof uses the square-root bound $\eta\leq e^{1/2}$. This bound costs the factor $n^4\lambda_L^{-4}$ in~\eqref{eq:kernel_width_delta}, but it needs no lower bound on $\lambda_{L-1}$. The factors $(1+b)/k$, $L^4$ and $\lambda_*^{-4}$ come from Lemma~\ref{lem:init_min_eigen_H}.

\subsection{Bounds on the Least Eigenvalue during Optimization}

In this section, we adapt Lemma C.5 in \cite{bpsw21} to the last layer of a multi-layer neural network, following the argument of Lemma C.2 in~\cite{syz21} for two-layer networks with a shifted ReLU. We treat the last layer as a two-layer network whose inputs are the unit vectors $\wh h_i:=h_{i,L-1}/\|h_{i,L-1}\|_2$. Throughout, we write $W_L=\sqrt{2/m}\,\wt W_L$, where $\wt W_L(0)$ has i.i.d.\ ${\cal N}(0,1)$ entries. We measure the movement of the last layer by the rows $\wt W_{L,r}$ of $\wt W_L$. The threshold of the last layer on input $x_i$ is $\sqrt{2/m}\,b\,\|h_{i,L-1}\|_2$, so neuron $r$ fires on $x_i$ if and only if $\langle\wt W_{L,r},\wh h_i\rangle>b$. For a matrix $\wt W_L$, let
\begin{align*}
    S_i(\wt W_L):=\{r\in[m]:\langle\wt W_{L,r},\wh h_i\rangle>b\}
\end{align*}
be the set of neurons of the last layer that fire on $x_i$, so that $D_{i,L}=\sqrt{c_b}\,\diag({\bf 1}_{S_i(\wt W_L)})$. The hidden vectors $h_{i,L-1}$ do not change during training, and neither do the $\wh h_i$.

\begin{lemma}[Changes of the activation pattern of the last layer]\label{lem:flip}
Fix $W_1,\dots,W_{L-1}$ such that $h_{i,L-1}\neq 0$ for all $i\in[n]$. Let $\wt R>0$ with $b\wt R\leq 1$, let $\delta\in(0,1)$, and let
\begin{align*}
    \zeta:=\wt R+\frac{(1+b)\log(n/\delta)}{k}.
\end{align*}
Then, with probability at least $1-\delta$ over $\wt W_L(0)$, for every $i\in[n]$ and every $\wt W_L$ with $\|\wt W_{L,r}-\wt W_{L,r}(0)\|_2\leq\wt R$ for all $r\in[m]$,
\begin{align*}
    \big|S_i(\wt W_L)\,\triangle\,S_i(\wt W_L(0))\big|\leq N:=4e\,m\wt R\,p(b)+4\log(n/\delta),
\end{align*}
and $N/(mG(b))\leq 44\,\zeta$.
\end{lemma}
\begin{proof}
Fix $i$. If neuron $r$ is in the symmetric difference, then the threshold $b$ lies between $\langle\wt W_{L,r}(0),\wh h_i\rangle$ and $\langle\wt W_{L,r},\wh h_i\rangle$, whose difference is at most $\|\wt W_{L,r}-\wt W_{L,r}(0)\|_2\leq\wt R$, since $\|\wh h_i\|_2=1$. So $|\langle\wt W_{L,r}(0),\wh h_i\rangle-b|\leq\wt R$. Let $N_i$ be the number of $r\in[m]$ with this property. It depends only on $\wt W_L(0)$, so it bounds the symmetric difference for all $\wt W_L$ at once. The variables $g_r:=\langle\wt W_{L,r}(0),\wh h_i\rangle$, $r\in[m]$, are i.i.d.\ ${\cal N}(0,1)$. Every $x$ with $|x-b|\leq\wt R$ satisfies $x^2\geq\max\{b-\wt R,0\}^2\geq b^2-2b\wt R$, so $p(x)\leq p(b)\,e^{b\wt R}\leq e\,p(b)$. Hence $q:=\Pr[|g_r-b|\leq\wt R]\leq 2e\,\wt R\,p(b)$, and $N_i$ has the binomial distribution ${\cal B}(m,q)$. Let $\ell_0:=\log(n/\delta)$ and $t:=mq+4\ell_0$. Lemma~\ref{lem:bernstein}, applied to the variables ${\bf 1}[|g_r-b|\leq\wt R]-q$, whose variances sum to at most $mq\leq t$, gives
\begin{align*}
    \Pr[N_i-mq>t]\leq\exp\Big(-\frac{t^2/2}{mq+t/3}\Big)\leq\exp\Big(-\frac{3t}{8}\Big)\leq e^{-\ell_0}=\frac{\delta}{n}.
\end{align*}
So $N_i\leq 2mq+4\ell_0\leq N$ with probability at least $1-\delta/n$, and a union bound over $i\in[n]$ proves the first claim. For the second, Fact~\ref{fact:gauss_quant} gives $p(b)/G(b)\leq 4/(1+b)$ and $1/(mG(b))\leq 3(1+b)/k$, so
\begin{align*}
    \frac{N}{mG(b)}\leq\frac{16e\,\wt R}{1+b}+\frac{12(1+b)\log(n/\delta)}{k}\leq 44\,\zeta.
\end{align*}
\end{proof}

\begin{lemma}[The Gram matrix after a small change of the last layer]
\label{lem:eigen_perturb}
Let $\wt R$, $\delta$ and $\zeta$ be as in Lemma~\ref{lem:flip}, and suppose $\|h_{i,L-1}\|_2\in[0.9,1.1]$ for all $i\in[n]$. Take the Gram matrix $G_L(W)=J_LJ_L^\top$ of the last layer with respect to $\wt W_L$. On the event of Lemma~\ref{lem:flip}, every $\wt W_L$ with $\|\wt W_{L,r}-\wt W_{L,r}(0)\|_2\leq\wt R$ for all $r\in[m]$ satisfies
\begin{align*}
    \|G_L(W)-G_L(W(0))\|_F\leq 107\,n\zeta,
\end{align*}
and hence $\lambda_{\min}(G_L(W))\geq\lambda_{\min}(G_L(W(0)))-107\,n\zeta$.
\end{lemma}
\begin{proof}
As in the proof of Lemma~\ref{lem:init_min_eigen_L}, since $2c_b/m=1/(mG(b))$,
\begin{align*}
    (G_L(W))_{i,j}=h_{i,L-1}^\top h_{j,L-1}\cdot\frac{2}{m}\tr[D_{i,L}D_{j,L}]=h_{i,L-1}^\top h_{j,L-1}\cdot\frac{|S_i(\wt W_L)\cap S_j(\wt W_L)|}{mG(b)}.
\end{align*} The symmetric difference of $S_i(\wt W_L)\cap S_j(\wt W_L)$ and $S_i(\wt W_L(0))\cap S_j(\wt W_L(0))$ is contained in the union of the symmetric differences for $i$ and for $j$, so by Lemma~\ref{lem:flip}, the two intersections differ in size by at most $2N$. With $|h_{i,L-1}^\top h_{j,L-1}|\leq 1.21$, every entry of $G_L(W)-G_L(W(0))$ is at most $1.21\cdot 2N/(mG(b))\leq 1.21\cdot 88\,\zeta\leq 107\,\zeta$ in absolute value. So $\|G_L(W)-G_L(W(0))\|_F\leq 107\,n\zeta$, and Weyl's inequality gives the second claim.
\end{proof}

\begin{corollary}
\label{cor:eigen_perturb}
Suppose the hypotheses of Lemma~\ref{lem:eigen_perturb} hold, $\lambda_{\min}(G_L(W(0)))\geq\frac{3}{4}\lambda_L$, and $428\,n\zeta\leq\lambda_L$. Then, on the event of Lemma~\ref{lem:flip}, $\lambda_{\min}(G_L(W))\geq\frac{\lambda_L}{2}$ for every $\wt W_L$ with $\|\wt W_{L,r}-\wt W_{L,r}(0)\|_2\leq\wt R$ for all $r\in[m]$.
\end{corollary}

\begin{proof}
By Lemma~\ref{lem:eigen_perturb}, $\lambda_{\min}(G_L(W))\geq\frac{3}{4}\lambda_L-107\,n\zeta\geq\frac{3}{4}\lambda_L-\frac{1}{4}\lambda_L=\frac{\lambda_L}{2}$.
\end{proof}

\begin{remark}
The factor $1/(mG(b))$ in the Gram matrix is of order $1/((1+b)k)$, since $mG(b)=\Theta((1+b)k)$ by Fact~\ref{fact:gauss_quant}(ii) and Remark~\ref{remark:rate_H}. This is much larger than $1/m$. Hoeffding's inequality bounds the number of changes of the activation pattern with an additive deviation of order $\sqrt m$, up to logarithmic factors. Because of this large factor, such a bound would need a much larger width when $\gamma<1/2$, and it would fail for every width polynomial in $n$ when $\gamma\geq 1/2$. Lemma~\ref{lem:flip} instead combines Bernstein's inequality with the fact that the pre-activations have density about $p(b)$ near the threshold, so that only about $m\wt R\,p(b)$ neurons can change. Since $p(b)/G(b)\leq 4/(1+b)$, the resulting change of the Gram matrix is $O(n\wt R)$ up to the additive term $(1+b)\log(n/\delta)/k$. The normalized inputs $\wh h_i$ have norm exactly $1$ because the threshold of the last layer is proportional to $\|h_{i,L-1}\|_2$.
\end{remark}

\section{Convergence Analysis of Our Algorithm}
\label{sec:analysis}
In this section, we present a convergence analysis of Algorithm~\ref{alg:complete}. We show that as long as the neural network width is large enough, the convergence of Algorithm~\ref{alg:complete} is linear, and the weight matrix does not change too much.
\subsection{Preliminary}
We recall the initialization of our neural network.

\begin{definition}[Initialization]\label{def:init_weights}
Let $m = m_{\ell}$ for all $\ell \in [L]$. Let $m_0 = d$. 
We assume weights are initialized as 
\begin{itemize}
    \item Each entry of weight vector $a \in \R^{m}$ is i.i.d. sampled from $\{ -1 , + 1 \}$ uniformly at random.
    \item Each entry of the weight matrices $W_1\in\R^{m\times d}$ and $W_\ell\in\R^{m\times m}$ for $\ell\in\{2,\dots,L\}$ is sampled from ${\cal N}(0,2/m)$.
\end{itemize}
\end{definition}

\begin{remark}
Later, we will also interpret $W_L$ as sampled from ${\cal N}(0,1)$ and then being scaled by $\sqrt{\frac{2}{m}}$.
\end{remark}

We also restate the architecture of our neural network here.

\begin{definition}[Architecture]\label{def:architecture}
Our neural network is a standard $L$-layer feed-forward neural network whose activations are scaled versions of the shifted ReLU, with a threshold proportional to the norm of the input of the layer. Let $c_b:=(2(1-\Phi(b) + b\,p(b)))^{-1}$, where $p$ is the standard Gaussian density and $b>0$ is a threshold parameter that we pick later. For a threshold $\tau\geq 0$, let
\begin{align*}
    \phi_\tau(x):=\sqrt{c_b}\,{\bf 1}[x>\tau]\,x,\qquad \phi_{L,\tau}(x):=\sqrt{c_b}\max\{x-\tau,0\}.
\end{align*}
For $i\in[n]$, let $h_{i,0}:=x_i$, and for $\ell\in[L]$, let $\tau_{i,\ell}:=\sqrt{2/m}\,b\,\|h_{i,\ell-1}(0)\|_2$ be the threshold of layer $\ell$ on input $x_i$, computed with the initial weights, and
\begin{align*}
    h_{i,\ell}:=\phi_{\tau_{i,\ell}}(W_\ell h_{i,\ell-1})~~~\text{for}~\ell\in[L-1],\qquad h_{i,L}:=\phi_{L,\tau_{i,L}}(W_Lh_{i,L-1}),
\end{align*}
with the activations applied coordinate-wise. At the last layer, we use a vector $a$ whose entries are Rademacher random variables. We define the neural network function on the data points as
\begin{align*}
    f(W,x_i) = & ~ a^\top h_{i,L}.
\end{align*}
We write $\phi$ for $\phi_{\tau_{i,\ell}}$ and $\phi_L$ for $\phi_{L,\tau_{i,L}}$, and leave the dependence of the threshold on $i$ and $\ell$ implicit, as in $h_{i,\ell}=\phi(W_\ell h_{i,\ell-1})$ and $D_{i,\ell}=\diag(\phi'(W_\ell h_{i,\ell-1}))=\sqrt{c_b}\,\diag({\bf 1}[W_\ell h_{i,\ell-1}>\tau_{i,\ell}])$. The hidden layers use the thresholded activation $\phi$, and the trained last layer uses the continuous activation $\phi_L$. Since $\phi_L'=\phi'$ almost everywhere and $|\phi_L|\leq|\phi|$, the matrices $D_{i,L}=\diag(\phi'(W_Lh_{i,L-1}))$ are unchanged, and $f(\cdot,x_i)$ is continuous and piecewise linear in $W_L$. In this section, $h_{i,L}:=\phi_L(W_Lh_{i,L-1})$. In contrast, Lemma~\ref{lem:als_71} below concerns the network that uses $\phi$ in every layer.
We measure the loss of the neural network via squared-loss function:
\begin{align*}
    {\cal L}(W) = & ~ \frac{1}{2}\sum_{i=1}^n (f(x_i)-y_i)^2.
\end{align*}
We use $f_t:\R^{d\times n}\rightarrow \R^n$ denote the prediction of our network:
\begin{align*}
    f_t(X) = & ~ [f(W(t),x_1),\ldots, f(W(t),x_n)]^\top.
\end{align*}
\end{definition}

\begin{remark}[Fixed thresholds and the change from the published version]\label{remark:threshold}
Algorithm~\ref{alg:complete} trains only $W_L$, so $h_{i,\ell}(t)=h_{i,\ell}(0)$ for $\ell\leq L-1$, and $\tau_{i,\ell}=\sqrt{2/m}\,b\,\|h_{i,\ell-1}(t)\|_2$ at every step $t$. Writing $\wh h:=h/\|h\|_2$ for $h\neq 0$, we have $h_{i,\ell}=\|h_{i,\ell-1}\|_2\,\phi_{\sqrt{2/m}\,b}(W_\ell\wh h_{i,\ell-1})$ at initialization. So the network first normalizes the input of a layer, then applies the activation with the fixed threshold $\sqrt{2/m}\,b$, and then multiplies the output by the norm of the input. In particular, the map $h\mapsto\|h\|_2\,\phi_{\sqrt{2/m}\,b}(W_\ell\wh h)$ that each hidden layer applies is positively homogeneous, that is, multiplying its input by a scalar $c>0$ multiplies its output by $c$. This map agrees with the network on the hidden inputs $h_{i,\ell-1}$, which never change.

The published version of this paper (ITCS 2024) uses the fixed threshold $\tau_{i,\ell}=\sqrt{2/m}\,b$ in every layer, and the thresholded activation $\phi$ also in the last layer. The two networks agree on the first layer, since $\|x_i\|_2=1$. They differ on the later layers, where $\|h_{i,\ell-1}\|_2$ is only close to $1$. We use the norm-scaled threshold because it makes each layer positively homogeneous. With the fixed threshold, the mean of $\|h_{i,\ell}\|_2^2$ given $\|h_{i,\ell-1}\|_2^2=s$ is $s\,G(b/\sqrt s)/G(b)$, with $G$ as in Section~\ref{sec:spectral_app}. This map fixes $s=1$. Since $G'(x)=-x^2p(x)$, its slope there is $1+b^3p(b)/(2G(b))\geq 1+b^4/(2+2b^2)$, where we used $G(b)\leq p(b)(1+b^2)/b$, a consequence of~\eqref{eq:mills}. So deviations of the norms from $1$ can grow geometrically with the depth. An analysis of the fixed-threshold network that keeps them small layer by layer therefore needs a factor $(C\log m)^{2L+O(1)}$ in the width, for an absolute constant $C$. With the norm-scaled threshold, the norm deviations add up over the layers instead (Lemma~\ref{lem:als_71}). The comparison with the kernels is then polynomial in the depth (Lemma~\ref{lem:init_min_eigen_H}). Each neuron of each layer still fires with probability exactly $P_b=1-\Phi(b)$ at initialization (Lemma~\ref{lem:init_sparse}), so the sparsity and the cost per iteration are as before.
\end{remark}

\subsection{Technical Lemmas}
We first show that during initialization, by using our shifted ReLU activation, the vector $h_{i,\ell}$ is sparse. Hence, the diagonal matrix $D_{i,\ell}$ is sparse as well.

\begin{lemma}[Sparse initialization]
\label{lem:init_sparse}
Let $P_b:=1-\Phi(b)$ and $k:=m\exp(-b^2/2)$. With probability at least
\begin{align*}
1-nL\cdot e^{-mP_b/3}\geq 1-nL\cdot e^{-k/(9(1+b))}
\end{align*}
over the initialization, for all $i\in [n]$ and all $\ell\in [L]$,
\begin{align*}
\|h_{i,\ell}(0)\|_0\leq\|D_{i,\ell}(0)\|_0\leq 2mP_b\leq 2k.
\end{align*}
\end{lemma}

\begin{proof}
We fix $i\in [n]$ and $\ell\in [L]$, since we will union bound over all $i$ and $\ell$ at last. Condition on $W_1,\dots,W_{\ell-1}$, so that $u:=h_{i,\ell-1}$ is a fixed vector. If $u=0$, then $W_\ell u=0$ and $\tau_{i,\ell}=0$, so $D_{i,\ell}=0$. Otherwise, let $W_{\ell,r}$ denote the $r$-th row of $W_\ell$. The variables $g_r:=\sqrt{m/2}\,\langle W_{\ell,r},u/\|u\|_2\rangle$, $r\in[m]$, are i.i.d.\ ${\cal N}(0,1)$, and the $r$-th diagonal entry of $D_{i,\ell}$ is nonzero if and only if $\langle W_{\ell,r},u\rangle>\tau_{i,\ell}=\sqrt{2/m}\,b\,\|u\|_2$, that is, if and only if $g_r>b$. So $\|D_{i,\ell}\|_0$ has the binomial distribution ${\cal B}(m,P_b)$, with mean $mP_b$. This step uses the norm-scaled threshold, which makes each neuron fire with probability exactly $P_b$, whatever the norm of the input. By the Chernoff bound (Lemma~\ref{lem:chernoff}) with $\epsilon=1$,
\begin{align*}
    \Pr\big[\|D_{i,\ell}\|_0\geq 2mP_b\big]\leq\exp(-mP_b/3).
\end{align*}
A union bound over $i$ and $\ell$ gives the probability bound. The inequalities $mP_b\geq k/(3(1+b))$ and $mP_b\leq k$ hold by Fact~\ref{fact:gauss_quant}. Finally, for $\ell\leq L-1$ we have $h_{i,\ell}=D_{i,\ell}W_\ell h_{i,\ell-1}$, since $\phi(x)=\phi'(x)x$, and $h_{i,L}=\phi_L(W_Lh_{i,L-1})$ vanishes outside the support of $D_{i,L}$. So $\|h_{i,\ell}\|_0\leq\|D_{i,\ell}\|_0$.
\end{proof}

\begin{remark}\label{remark:k}
The above lemma shows that by using the shifted ReLU activation, we make sure that all $h_{i,\ell}$ are sparse after initialization. We use $k:=m\exp(-b^2/2)$ as the sparsity parameter. If we choose $b=\sqrt{2\gamma\log m}$ for a constant $\gamma\in(0,1)$, then $k=m^{1-\gamma}$, and hence with high probability, $\|h_{i,\ell}(0)\|_0\leq\|D_{i,\ell}(0)\|_0\leq 2m^{1-\gamma}$ for all $i\in[n]$ and $\ell\in[L]$. Since the threshold of each layer is proportional to the norm of its input, this needs no bound on the norms of the hidden vectors. The failure probability $nL\,e^{-mP_b/3}$ is at most $\delta$ once $k\geq 9(1+b)\log(nL/\delta)$, by Fact~\ref{fact:gauss_quant}.

We write $\bar k:=k/((1+b)L^2)$. By Lemma~\ref{lem:als_71} below with $\epsilon=1/10$, we have $\|h_{i,\ell}\|_2\in[0.9,1.1]$ for all $i\in[n]$ and $\ell\in[L]$ with probability at least $1-O(nL)\cdot e^{-\Omega(\bar k)}$. Since $k/(9(1+b))\geq\bar k/9$, the failure probability of Lemma~\ref{lem:init_sparse} is also at most $nL\cdot e^{-\Omega(\bar k)}$. For $b=\sqrt{2\gamma\log m}$ and $m\geq 3$, we have $1+b\leq 3\sqrt{\log m}$, so $\bar k\geq m^{1-\gamma}/(3L^2\sqrt{\log m})$.

As a direct consequence, we note that all initial $D_{i,\ell}$ are $O(k)$-sparse as well.
\end{remark}

We state a lemma that handles the $\ell_2$ norm of $h_{i,\ell}$ when one uses \emph{truncated Gaussian distribution} instead. We defer its proof to Section~\ref{sec:modified_als}.
\begin{lemma}[Restatement of Lemma~\ref{lem:als_71_shift}]
\label{lem:als_71}
Let $b>0$, suppose $\|x_i\|_2=1$ for all $i\in[n]$, and consider the network of Definition~\ref{def:architecture} at initialization, with the thresholded activation $\phi$ in every layer, including layer $L$. Let $P_b:=1-\Phi(b)$, $k:=m\exp(-b^2/2)$ and $\bar k:=k/((1+b)L^2)$. There is an absolute constant $c_1>0$ such that, for every $\epsilon\in(0,1]$, with probability at least
\begin{align*}
    1-6nL\cdot \exp\Big(-\frac{c_1\epsilon^2P_b\,m}{4L^2}\Big)\geq 1-6nL\cdot\exp\big(-\Omega(\epsilon^2\bar k)\big)
\end{align*}
over $W(0)$, we have $\|h_{i,\ell}\|_2^2\in[1-\epsilon,1+\epsilon]$, and hence $\|h_{i,\ell}\|_2\in[1-\eps,1+\eps]$, for all $i\in [n]$ and $\ell\in [L]$. In particular, with probability at least $1-O(nL)\cdot e^{-\Omega(\bar k)}$, we have $\|h_{i,\ell}\|_2\in[0.9,1.1]$ for all $i\in[n]$ and $\ell\in[L]$.
\end{lemma}

\subsection{Bounds on Initialization}
In the following lemma, we generalize the Lemma C.2 in \cite{bpsw21} into multiple layer neural networks.
\begin{lemma}[Bounds on initialization, multiple layer version of Lemma C.2 in \cite{bpsw21}]
\label{lem:init_f_J}
Suppose $m = \Omega(nL\log(nL))$, then we have the following
\begin{itemize}
    \item $\Pr[ f(W,x_i) = \wt O(1), ~ \forall i \in [n] ] \geq 1- e^{-\Omega(\log^2 n)}-O(nL)\cdot e^{-\Omega(\bar k)}$.
    \item  $\Pr[ \| J_{L,0,i} \| \leq 1.6, ~ \forall i \in [n] ] \geq 1-O(nL)\cdot e^{-\Omega(\bar k)}$, for the Jacobian with respect to $\wt W_L:=\sqrt{m/2}\,W_L$.
\end{itemize}
Here $\bar k$ is as in Lemma~\ref{lem:als_71}.
\end{lemma}

\begin{proof}
We will prove the two parts of the statement separately.
\paragraph{Part 1:}
By definition, for any $i\in [n]$, we have
\begin{align*}
        f(W,x_i) = a^\top \phi_L( W_L ( \phi ( \cdots \phi(W_1 x_i ) ) ) ).
\end{align*}

We shall make use of Lemma~\ref{lem:als_71} here, with $\epsilon=1/10$. It gives $\|\phi(W_Lh_{i,L-1})\|_2\leq 1.1$ for all $i\in[n]$, and since $|\phi_L|\leq|\phi|$,
\begin{align*}
\Pr \big[ \|h_{i,L}\|_2\leq 1.1, \forall i \in [n] \big] \geq 1-O(nL)\cdot \exp(-\Omega(\bar k)) .
\end{align*}

Recall that $a\in \R^m$ has each of its entry being a Rademacher random variable, hence it's 1-subgaussian. Use the concentration of subgaussian~(Lemma~\ref{lem:subgaussian}), we know that
\begin{align*}
    \Pr[|a^\top h_{i,L}|\geq 1.1t] \leq & ~ 2\exp(-\frac{t^2}{2}),
\end{align*}
setting $t=2\log(2n)$, and union bound over all $i\in [n]$, we conclude our desired result, with $|f(W,x_i)|\leq 2.2\log(2n)$ for all $i$.

\paragraph{Part 2:}

For the last layer, we write $W_L=\sqrt{2/m}\,\wt W_L$, where each entry of $\wt W_L$ is sampled from ${\cal N}(0,1)$, and we take the Jacobian with respect to $\wt W_L$. Its $i$-th row, viewed as an $m\times m$ matrix, is $\sqrt{2/m}\,D_{i,L}a h_{i,L-1}^\top$, which has rank one. The nonzero diagonal entries of $D_{i,L}$ are equal to $\sqrt{c_b}$, and $a_r^2=1$, so $\|D_{i,L}a\|_2^2=c_b\|D_{i,L}\|_0$. Hence, using $2c_b/m=1/(mG(b))$,
\begin{align*}
    \|J_{L,0,i}\|= & ~ \sqrt{2/m}\,\|D_{i,L}a\|_2\,\|h_{i,L-1}\|_2 \\
    = & ~ \Big(\frac{\|D_{i,L}\|_0}{mG(b)}\Big)^{1/2}\|h_{i,L-1}\|_2.
\end{align*}
By Lemma~\ref{lem:init_sparse}, $\|D_{i,L}(0)\|_0\leq 2mP_b$ for all $i$ with probability at least $1-nL\cdot e^{-\Omega(\bar k)}$, and by Lemma~\ref{lem:als_71}, $\|h_{i,L-1}\|_2\leq 1.1$ for all $i$ with probability at least $1-O(nL)\cdot \exp(-\Omega(\bar k))$. On both events, since $P_b\leq G(b)$ (Fact~\ref{fact:gauss_quant}), $\|J_{L,0,i}\|^2\leq 2.42\,P_b/G(b)\leq 2.42$, so $\|J_{L,0,i}\|\leq 1.6$.
\end{proof}

\subsection{Bounds on Small Perturbation}
In the following, we generalize the Lemma C.4 in \cite{bpsw21} into multiple layer neural network. As in Section~\ref{sec:spectral_app}, we write $W_L=\sqrt{2/m}\,\wt W_L$, where $\wt W_L(0)$ has i.i.d.\ ${\cal N}(0,1)$ entries, and we take the Jacobian $J_{W_L}$ of the last layer with respect to $\wt W_L$. We measure the movement of the last layer by the rows $\wt W_{L,r}$ of $\wt W_L$. We write $J_{W_L,x_i}$ for the $i$-th row of $J_{W_L}$, viewed as an $m\times m$ matrix.

\begin{lemma}[multiple layer version of Lemma C.4 in \cite{bpsw21}]
\label{lem:perturb}
Let $\wt R>0$ with $b\wt R\leq 1$, let $\delta\in(0,1)$, and suppose that $\zeta:=\wt R+(1+b)\log(n/\delta)/k\leq 1/300$. Then, over the random initialization of
\begin{align*}
W(0) = \{ W_1(0), W_2(0), \cdots W_L(0) \},
\end{align*}
the following holds with probability at least $1-\delta-O(nL)\cdot e^{-\Omega(\bar k)}$ ($\bar k$ as in Lemma~\ref{lem:als_71}), for every $\wt W_L$ satisfying $\| \wt W_{L,r} - \wt W_{L,r}(0) \|_2 \leq \wt R$ for each $r\in [m]$:
\begin{itemize}
    \item  $\|\wt W_L-\wt W_L(0)\|_F \leq R:=\sqrt m\,\wt R$.
    \item $\| J_{W_L,x_i} - J_{W_L(0), x_i} \| \leq\sqrt{54\zeta}$ for all $i\in[n]$.
    \item $\| J_{W_{L}} - J_{ W_{L}(0) }\|_F \leq\sqrt{54n\zeta}$.
    \item $\| J_{W_{L}(0)} \|_F \leq 1.6\sqrt n$ and $\| J_{W_{L}} \|_F \leq 2\sqrt n$.
\end{itemize}
\end{lemma}
\begin{proof}
We use three events. By Lemma~\ref{lem:als_71} with $\epsilon=1/10$, $\|h_{i,L-1}\|_2\in[0.9,1.1]$ for all $i$ with probability at least $1-O(nL)\cdot e^{-\Omega(\bar k)}$, and this event depends only on $W_1,\dots,W_{L-1}$. By Lemma~\ref{lem:init_sparse}, $|S_i(\wt W_L(0))|=\|D_{i,L}(0)\|_0\leq 2mP_b$ for all $i$ with probability at least $1-nL\cdot e^{-\Omega(\bar k)}$, where $S_i$ is the set of neurons of the last layer that fire on $x_i$, as in Lemma~\ref{lem:flip}. On the first event, Lemma~\ref{lem:flip} applies conditionally on $W_1,\dots,W_{L-1}$, and its event has probability at least $1-\delta$. We work on the intersection of the three events.

{\bf Part 1.} Note that
\begin{align*}
   \|\wt W_L-\wt W_L(0)\|_F^2 = & ~ \sum_{r=1}^m \|\wt W_{L,r}-\wt W_{L,r}(0)\|_2^2 \leq m\wt R^2 = R^2.
\end{align*}

{\bf Part 2.} To simplify the notation, we fix $i$ and write $h:=h_{i,L-1}$, $S:=S_i(\wt W_L)$ and $S_0:=S_i(\wt W_L(0))$. The $i$-th row of the Jacobian is $\sqrt{2/m}\,D_{i,L}ah^\top$ with $D_{i,L}=\sqrt{c_b}\,\diag({\bf 1}_S)$. So $J_{W_L,x_i}-J_{W_L(0),x_i}=\sqrt{2c_b/m}\,\diag({\bf 1}_S-{\bf 1}_{S_0})\,a\,h^\top$, a matrix of rank one. Since $a_r^2=1$ and $2c_b/m=1/(mG(b))$, Lemma~\ref{lem:flip} gives
\begin{align*}
    \|J_{W_L,x_i}-J_{W_L(0),x_i}\|^2 = & ~ \frac{\|h\|_2^2\,|S\triangle S_0|}{mG(b)} \leq 1.21\cdot 44\,\zeta\leq 54\,\zeta.
\end{align*}

{\bf Part 3.} The squared Frobenius norm is the sum of the squared norms of the rows, hence
\begin{align*}
    \|J_{W_L}-J_{W_{L}(0)}\|_F^2 \leq 54\,n\zeta.
\end{align*}

{\bf Part 4.} As in the proof of Lemma~\ref{lem:init_f_J}, $\|J_{W_L(0),x_i}\|^2=\|h_{i,L-1}\|_2^2\,|S_i(\wt W_L(0))|/(mG(b))\leq 2.42\,P_b/G(b)\leq 2.42$, so $\|J_{W_L(0)}\|_F\leq\sqrt{2.42\,n}\leq 1.6\sqrt n$. By the triangle inequality and Part 3,
\begin{align*}
    \|J_{W_L}\|_F \leq & ~ \|J_{W_{L}(0)}\|_F+\|J_{W_L}-J_{W_L(0)}\|_F \leq \big(\sqrt{2.42}+\sqrt{54/300}\big)\sqrt n\leq 2\sqrt n.
\end{align*}
\end{proof}

\subsection{Putting It All Together}
In this section, we will prove the following core theorem that analyzes the convergence behavior of Algorithm~\ref{alg:complete}. We first state the width that we need.

\begin{definition}[Width condition]
\label{def:m}
We say that the width $m$ satisfies the width condition if $\lambda_L>0$ and $\lambda_*>0$ (Definition~\ref{def:gram_multilayer}), and the sparsity parameter $k=m\,e^{-b^2/2}$ satisfies
\begin{align}\label{eq:kernel_width}
    k\geq C_W\,(1+b)\,n^4L^4\lambda_L^{-4}\lambda_*^{-4}\log^{c_W}(mnL)
\end{align}
for large enough absolute constants $C_W$ and $c_W$, where $c_W$ depends on the constant $c$ of Definition~\ref{def:R}.
\end{definition}

For $b=\sqrt{2\gamma\log m}$ with $\gamma\in(0,1)$ and $m\geq 3$, we have $k=m^{1-\gamma}$ and $1+b\leq 3\sqrt{\log m}\leq 3\log(mnL)$, so the width condition holds whenever $m^{1-\gamma}\geq 3C_W\,n^4L^4\lambda_L^{-4}\lambda_*^{-4}\log^{c_W+1}(mnL)$. Since $m\geq k$, the width condition implies $m\geq n^4$.

\begin{theorem}[Formal version of Theorem~\ref{thm:conv_main}]
\label{thm:conv_app}
Suppose the width $m$ satisfies the width condition of Definition~\ref{def:m}, and run Algorithm~\ref{alg:complete} with the parameters of Definition~\ref{def:R}. Then, for every $T\geq 1$, with probability at least $1-(T+1)\exp(-\Omega(\log^2 n))$ over the randomness of the initialization of the neural network and the randomness of the algorithm, we have
\begin{align*}
   \|f_{t+1}-y\|_2 \leq \frac{1}{3}\|f_t-y\|_2~~~\text{for all}~t\in\{0,1,\dots,T-1\}.
\end{align*}
\end{theorem}

The proof uses the following events, where $\delta_0:=e^{-\log^2 n}$. We assume $n\geq 2$ here and below, so that $\delta_0\in(0,1)$, as Lemma~\ref{lem:flip} requires.
\begin{itemize}
    \item[(E1)] The norm bound $\|h_{i,\ell}\|_2\in[0.9,1.1]$ for all $i$ and $\ell$ (Lemma~\ref{lem:als_71} with $\epsilon=1/10$), the sparsity bound of Lemma~\ref{lem:init_sparse}, and the two bounds of Lemma~\ref{lem:init_f_J}.
    \item[(E2)] The bound $\lambda_{\min}(G_L(W(0)))\geq\frac{3}{4}\lambda_L$ of Lemma~\ref{lem:init_min_eigen_L} with $\delta=\delta_0$.
    \item[(E3)] The event of Lemma~\ref{lem:flip} with the radius $\wt R$ of Definition~\ref{def:R} and $\delta=\delta_0$.
    \item[(E4)] For each $t\in\{0,\dots,T-1\}$, the solver \textsc{FastTensorRegression} at step $t$, which is run with failure probability $\delta_0$, returns a vector with the guarantees of Theorem~\ref{thm:fast_kronecker_regression}.
\end{itemize}
The events (E1) to (E3) concern only the initialization $W(0)$ and $a$. Under the width condition, each of them fails with probability at most $e^{-\Omega(\log^2 n)}$. The event (E4) fails with probability at most $T\delta_0$. For (E1), the failure probabilities are $O(nL)\cdot e^{-\Omega(\bar k)}+e^{-\Omega(\log^2 n)}$. Since $\bar k=k/((1+b)L^2)\geq C_W\log^{c_W}(mnL)$, we have $O(nL)\cdot e^{-\Omega(\bar k)}\leq e^{-\Omega(\log^2 n)}$ once $c_W\geq 2$ and $C_W$ is large enough. For (E2), we have $\log(mnL/\delta_0)=\log(mnL)+\log^2 n\leq 2\log^2(mnL)$, so condition~\eqref{eq:kernel_width_delta} of Lemma~\ref{lem:init_min_eigen_L} holds once $C_W\geq 4C_K$ and $c_W\geq 4$. For (E3), Definition~\ref{def:R} and Fact~\ref{fact:gauss_quant} give $\wt R\leq 5\sqrt 3\,n\sqrt{1+b}\,\log^{c}(nm)/(\lambda_L\sqrt k)$, and $\log(n/\delta_0)\leq 2\log^2(mnL)$, so~\eqref{eq:kernel_width} with $c_W\geq 2c$ and $C_W$ large enough gives
\begin{align}\label{eq:zeta_small}
    \zeta=\wt R+\frac{(1+b)\log(n/\delta_0)}{k}\leq\frac{\lambda_L}{40000\,n}.
\end{align}
In particular $\wt R\leq\lambda_L\leq 1/(1+b^2)\leq 1/b$, so $b\wt R\leq 1$. Hence Lemmas~\ref{lem:flip}, \ref{lem:eigen_perturb} and~\ref{lem:perturb} and Corollary~\ref{cor:eigen_perturb} apply with this $\wt R$ and $\delta=\delta_0$. By~\eqref{eq:zeta_small}, $\zeta\leq 1/300$ and $428\,n\zeta\leq\lambda_L$, so on (E1) to (E3), the conclusions of Lemma~\ref{lem:perturb} and Corollary~\ref{cor:eigen_perturb} hold for every $\wt W_L$ whose rows are within $\wt R$ of the initial ones. From now on we work on the intersection of (E1) to (E4), which has probability at least $1-(T+1)e^{-\Omega(\log^2 n)}$. Since only about $k$ neurons of a layer fire, the width condition is on $k$ rather than on $m$ (Remark~\ref{remark:rate_H}).

Before moving on, we introduce several definitions and prove some useful facts related to them.
\begin{definition}[function $\sf J$]\label{def:sf_J}
For matrices $Z=(Z_1,\dots,Z_L)$ in place of the weight matrices and for $i\in[n]$, let
\begin{align*}
    h_i(Z_1,\dots,Z_{L-1}):=&~ \phi(Z_{L-1}(\phi(Z_{L-2}\cdots (\phi(Z_1 x_i))))) && \in \R^{m}, \\
    D_{i,L}(Z_L) := &~ \diag(\phi'(Z_L h_i(Z_1,\dots,Z_{L-1}))) && \in \R^{m \times m},
\end{align*}
with the thresholds of Definition~\ref{def:architecture}, and let
\begin{align*}
    {\sf J}_L(Z)_i := \sqrt{2/m}\,D_{i,L}(Z_L)\, a\, (h_i(Z_1,\dots,Z_{L-1}))^\top ~~~ \in \R^{m\times m}.
\end{align*}
We define ${\sf J}_L(Z)\in\R^{n\times m^2}$ as the matrix whose $i$-th row is $\vect({\sf J}_L(Z)_i)^\top$. So ${\sf J}_L(Z)$ is the Jacobian of the outputs of the network with respect to $\wt W_L=\sqrt{m/2}\,W_L$ at the weights $Z$, in the scaling that we use for $J_{L,t}$ below.
\end{definition}

\begin{fact}\label{fact:f_diff}
Let ${\sf J}_{L}$ be the function of Definition~\ref{def:sf_J}. For any $t\in \{0,\dots, T-1\}$, we have
\begin{align*}
    f_{t+1}-f_t = \left(\int_0^1 {\sf J}_{L}((1-s)W(t)+sW(t+1))  \d s\right) \cdot  \vect(\wt W_L (t+1)-\wt W_L (t)).
\end{align*}
\end{fact}
\begin{proof}
For $i\in [n]$, consider the $i$-th coordinate. Only $W_L$ changes, and $f(\cdot,x_i)$ is continuous and piecewise linear in $W_L$ (Definition~\ref{def:architecture}). So, along the segment from $W(t)$ to $W(t+1)$, it is differentiable except at finitely many points, and elsewhere its gradient with respect to $\wt W_L$ is ${\sf J}_L(\cdot)_i$. Hence
\begin{align*}
    (f_{t+1}-f_t)_i = &~ \int_0^1 \frac{\d}{\d s} f((1-s)W(t) + sW(t+1), x_i)\,\d s\\
    = &~ \int_0^1 \vect\big({\sf J}_{L}((1-s)W(t)+sW(t+1))_i\big)^\top \vect(\wt W_L (t+1)-\wt W_L (t))\,\d s,
\end{align*}
which is the $i$-th coordinate of the right-hand side.
\end{proof}

\begin{fact}\label{fact:J_equiv}
For any $t\in \{0,\ldots,T\}$, we have ${\sf J}_L(W_1(t),\ldots,W_L(t))=J_{L,t}$, where $J_{L,t}$ is the Jacobian with respect to $\wt W_L$.
\end{fact}

\begin{proof}
We drop $t$ from the notation. The $i$-th row of $J_{L,t}$, viewed as an $m\times m$ matrix, is
\begin{align*}
    \frac{\partial f(W,x_i)}{\partial \wt W_L}=\sqrt{2/m}\,D_{i,L}\,a\,h_{i,L-1}^\top,
\end{align*}
where $D_{i,L}=\diag(\phi'(W_Lh_{i,L-1}))$ and $h_{i,L-1}=\phi(W_{L-1}(\phi(W_{L-2}\cdots (\phi(W_1x_i)))))$. These are $D_{i,L}(W_L)$ and $h_i(W_1,\ldots,W_{L-1})$ of Definition~\ref{def:sf_J}. This completes the proof.
\end{proof}
We state the range we require for parameter $\wt R$ and $R$:
	\begin{definition}\label{def:R}
	We choose $\wt R$ as
	\begin{align*}
	   \wt R := \frac{5\, n\log^{c}(nm)}{\lambda_L\sqrt{mG(b)}},
	\end{align*}
	where $c$ is an absolute constant large enough that $\|f_0-y\|_2\leq\sqrt n\log^{c}(nm)$ on the event of Lemma~\ref{lem:init_f_J}. This holds once $nm$ is larger than an absolute constant, since $|f_0(x_i)|\leq 2.2\log(2n)$ there and $|y_i|=O(1)$. Recall that $R$ is the scale-up version of $\wt R$, hence
	\begin{align*}
	   R := \sqrt{m}\,\wt R = \frac{5\, n\log^{c}(nm)}{\lambda_L\sqrt{G(b)}}.
	\end{align*}
	We also run \textsc{FastTensorRegression} in Algorithm~\ref{alg:complete} with precision $\epsilon_0:=\sqrt{\lambda_L/n}/30$ and failure probability $e^{-\log^2 n}$.
	\end{definition}
	
The factor $1/\sqrt{G(b)}$ in $\wt R$ comes from the entries $\sqrt{c_b}=1/\sqrt{2G(b)}$ of $D_{i,L}$, which scale the gradient (Section~\ref{sec:movement}). Since $mG(b)\geq k/(3(1+b))$ (Fact~\ref{fact:gauss_quant}), $\wt R$ is at most $\wt O(n/(\lambda_L\sqrt k))$. Hence~\eqref{eq:zeta_small} holds under the width condition.

\begin{remark}
For the last layer, Lemma~\ref{lem:flip} bounds the number of changes of the activation pattern on each data point by $O(m\wt R\,p(b)+\log(n/\delta))$. The running time of Algorithm~\ref{alg:complete} (Theorem~\ref{thm:alg_app}) does not depend on this number. The cost bound of the incremental implementation uses it (Theorem~\ref{thm:plain_incremental}(ii)).
\end{remark}

We use induction to prove the following two claims recursively. 
\begin{definition}[Induction hypothesis 1]
    Let $t \in [T]$ be a fixed integer. We have
    \begin{align*}
    \| \wt W_{L,r}(t) - \wt W_{L,r}(0) \|_2 \leq \wt R
    \end{align*}
    holds for any $r \in [m]$, where $\wt W_L=\sqrt{m/2}\,W_L$ and $\wt R$ is as in Definition~\ref{def:R}.
\end{definition}

\begin{definition}[Induction Hypothesis 2]
Let $t \in [T]$ be a fixed integer. We have
\begin{align*}
    \|f_{t} - y\|_2 \leq \frac{1}{3} \|f_{ t-1} - y\|_2.
\end{align*}
\end{definition}

	Throughout, the Jacobians $J_{L,t}$ and the weights are taken with respect to $\wt W_L$. We work on the intersection of the events (E1) to (E4) above. The Gauss--Newton step does not depend on the scaling of the parameters. Let $J'_{L,t}=\sqrt{m/2}\,J_{L,t}$ denote the Jacobian with respect to $W_L$, whose rows $u_{i,L}\otimes v_{i,L}$ Algorithm~\ref{alg:complete} passes to the solver, and let $g'_{L,t}$ be the solver output. Then $g_{L,t}:=\frac{m}{2}g'_{L,t}$ satisfies the guarantees of Theorem~\ref{thm:fast_kronecker_regression} for $J_{L,t}J_{L,t}^\top=\frac{2}{m}J'_{L,t}J'^\top_{L,t}$. The update of the algorithm is $\vect(\wt W_L(t+1)-\wt W_L(t))=-J_{L,t}^\top g_{L,t}$. The function ${\sf J}_L$ of Definition~\ref{def:sf_J} is in the same scaling, so ${\sf J}_L(W(t))=J_{L,t}$ by Fact~\ref{fact:J_equiv}.

	Suppose the above two claims hold up to $t$, we prove they continue to hold for time $t + 1$. In Section~\ref{sec:movement}, we first prove the first claim at time $t+1$, using the two claims only up to time $t$. The second claim is more delicate and also uses the first claim at time $t+1$. For it, we define
	\begin{align*}
	J_{L, t, t+1} := \int_{0}^{1} {\sf J}_{L} \Big( (1 - s) W_{t} + s W_{t+1} \Big) \d s,
	\end{align*}
	where ${\sf J}_{L} $ is defined in Definition~\ref{def:sf_J}.

	\begin{lemma}
	Let $g_{L}^{\star} := (J_{L,t} J_{L,t}^{\top})^{-1} (f_t - y)$. 
	We have
	\begin{align}\label{eq:second-order3}
	    \| f_{t+1} - y \|_2 
	    \leq & ~ \|f_{t}  - y - J_{L,t}J_{L,t}^\top g_{L,t} \|_2 \notag\\
	    + & ~  \|(J_{L,t}-J_{L, t, t+1})J_{L, t}^{\top} g^{\star}_L\|_2 \notag \\
	    + & ~ \|(J_{L,t}-J_{L, t, t+1})J_{L,t}^{\top} (g_{L,t} - g^{\star}_L)\|_2. 
	\end{align}
	
	\end{lemma}
	\begin{proof}
	Consider the following computation:
	\begin{align*}
	&~\|f_{t+1} - y\|_2 \notag\\
	= &~ \|f_{t}  - y + (f_{t + 1}  - f_{t})\|_2\notag\\
	= &~ \|f_{t}  - y + J_{L,t, t+1} \cdot \vect(\wt W_L(t+1) - \wt W_L(t))\|_2\notag\\
	= &~ \|f_{t}  - y - J_{L,t, t+1} \cdot J_{L,t}^{\top} g_{L,t}\|_2\notag\\
	= &~ \|f_{t}  - y -  J_{L, t}J_{L,t}^{\top} g_{L,t} +  J_{L, t}J_{L,t}^{\top} g_{L,t} -  J_{L,t, t+1}J_{L,t}^{\top} g_{L,t}\|_2 \notag\\
	\leq &~ \|f_{t}  - y -  J_{L,t} J_{L,t}^{\top} g_{L,t}\|_2 +  \|(J_{L,t}-J_{L, t, t+1})J_{L,t}^{\top} g_{L,t} \|_2\notag\\
	\leq &~ \|f_{t}  - y - J_{L,t} J_{L,t}^{\top} g_{L,t} \|_2 + \|(J_{L,t}-J_{L, t, t+1})J_{L, t}^{\top} g^{\star}_L\|_2 +   \|(J_{L,t}-J_{L, t, t+1})J_{L,t}^{\top} (g_{L,t} - g^{\star}_L)\|_2 ,
	\end{align*}
	The second step follows from Fact~\ref{fact:f_diff} and the definition of $J_{L,t, t+1}$, and the third step follows from the update rule. 
	\end{proof}
	
\begin{claim}[1st term in Eq.~\eqref{eq:second-order3}]
	We have 
	\begin{align*}
	    \|f_{t}  - y - J_{L,t} J_{L,t}^{\top} g_{L,t} \|_2 \leq \frac{1}{9}\|f_t - y\|_2.
	\end{align*}
\end{claim}
	
\begin{proof}
We have
	\begin{align}
	\label{eq:second-order5}
	\|f_{t}  - y - J_{L,t} J_{L,t}^{\top} g_{L,t} \|_2 \leq & ~ \epsilon_0\sqrt{\kappa_t} \|f_t - y\|_2 \notag \\
	\leq & ~ \frac{1}{9}\|f_t - y\|_2,
	\end{align}
	since $g_{L,t}$ is the output of Theorem~\ref{thm:fast_kronecker_regression} with precision $\epsilon_0$ (Definition~\ref{def:R}) for the regression problem
\begin{align*}
	\min_{g} \| J_{L,t} J_{L,t}^\top g - (f_t - y) \|_2,
\end{align*}
where $\kappa_t:=\lambda_{\max}(J_{L,t}J_{L,t}^\top)/\lambda_{\min}(J_{L,t}J_{L,t}^\top)\leq \|J_{L,t}\|_F^2/(\lambda_L/2)\leq 8n/\lambda_L$ by Lemma~\ref{lem:perturb} and Corollary~\ref{cor:eigen_perturb}, which apply to $\wt W_L(t)$ by Induction hypothesis 1. Hence $\epsilon_0\sqrt{\kappa_t}\leq\sqrt 8/30\leq 1/9$.
\end{proof}

\begin{claim}[2nd term in Eq.~\eqref{eq:second-order3}]
    We have
    \begin{align*}
       \|(J_{L,t}-J_{L,t, t+1})J_{L,t}^{\top} g^{\star}_L\|_2  \leq \frac{1}{9} \| f_t - y \|_2.
    \end{align*}
\end{claim}
\begin{proof}
	We bound the second term in Eq.~\eqref{eq:second-order3} as follows:
	\begin{align}
	\|(J_{L,t}-J_{L,t, t+1})J_{L,t}^{\top} g^{\star}_L\|_2
	\leq & ~ \|J_{L,t}-J_{L,t, t+1}\| \cdot \|J_{L,t}^{\top} g^{\star}_L\|_2\notag\\
	 = & ~ \|J_{L,t}-J_{L,t, t+1}\| \cdot \|J_{L,t}^{\top} (J_{L,t} J_{L,t}^{\top})^{-1} \cdot (f_t - y) \|_2\notag\\
	 \leq & ~\|J_{L,t}-J_{L,t, t+1}\| \cdot \|J_{L,t}^{\top} (J_{L,t} J_{L,t}^{\top})^{-1} \| \cdot \| f_t - y\|_2 \label{eq:second-order9}.
	\end{align}
	We bound these term separately.

	For the first term in Eq.~\eqref{eq:second-order9},
    \begin{align}\label{eq:second-order6}
        \|J_{L, t} - J_{L, t, t+1}\| = &~  \left\| \J_{L}(W_t) - \int_0^1 \J_L((1-s)W_t + sW_{t+1})\d s \right\|\notag \\
        \leq &~ \int_0^1 \left\| \J_{L}(W_t) - \J_L((1-s)W_t + sW_{t+1}) \right\|\d s \notag \\
        \leq & ~ \int_0^1 \| \J_{L}(W_t) - \J_{L} (W_0) \| + \| \J_{L} (W_0) - \J_L((1-s)W_t + sW_{t+1}) \|~\d s \notag \\
        \leq & ~ \| \J_{L}(W_t) - \J_{L} (W_0) \|+\int_0^1 \| \J_{L} (W_0) - \J_L((1-s)W_t + sW_{t+1}) \|~\d s \notag \\
        \leq & ~ 2\sqrt{54\,n\zeta}.
    \end{align}
    Here, by Fact~\ref{fact:J_equiv}, $\|\J_{L}(W_t) - \J_{L} (W_0)\| = \|J_{W_L(t)}-J_{W_L(0)}\|\leq\|J_{W_L(t)}-J_{W_L(0)}\|_F\leq\sqrt{54\,n\zeta}$ by Lemma~\ref{lem:perturb}, which applies to $\wt W_L(t)$ by Induction hypothesis 1 at time $t$. For the term $\int_0^1 \| \J_{L} (W_0) - \J_L((1-s)W_t + sW_{t+1}) \|~\d s$, note that by convexity of the norm and Induction hypothesis 1 at times $t$ and $t+1$, for every $r\in[m]$ and $s\in[0,1]$,
		\begin{align*}
		\|(1-s)\wt W_{L,r}(t)+s\wt W_{L,r}(t+1)-\wt W_{L,r}(0)\|_2 \leq \wt R.
		\end{align*}
		So $(1-s)\wt W_L(t)+s\wt W_{L}(t+1)$ satisfies the row bound of Lemma~\ref{lem:perturb}, hence $\| \J_{L} (W_0) - \J_L((1-s)W_t + sW_{t+1}) \|\leq\sqrt{54\,n\zeta}$.

	Furthermore, we have
	\begin{align}
		\|J_{L,t}^{\top} (J_{L,t} J_{L,t}^{\top})^{-1}\| = \frac{1}{\sigma_{\min}(J_{L,t}^{\top})} \leq \sqrt{2/\lambda_L}, \label{eq:second-order10}
	\end{align}
	where the second inequality follows from $\sigma_{\min}(J_{L,t}) = \sqrt{\lambda_{\min}(J_{L,t}J_{L,t}^\top)} \geq \sqrt{\lambda_L/2}$ (see Corollary~\ref{cor:eigen_perturb}).

	Combining Eq.~\eqref{eq:second-order9}, \eqref{eq:second-order6} and \eqref{eq:second-order10}, we have
	\begin{align}
	\label{eq:second-order11}
	\|(J_{L,t} - J_{L,t, t+1})J_{L,t}^{\top} g^{\star}_L\|_2
	\leq & ~ \Big(\frac{432\,n\zeta}{\lambda_L}\Big)^{1/2}\cdot \|f_t - y\|_2 \notag \\
	\leq & ~ \frac{1}{9}\|f_t - y\|_2,
	\end{align}
	where the last step follows from~\eqref{eq:zeta_small}, since $432/40000\leq 1/81$.
\end{proof}

\begin{claim}[3rd term in Eq.~\eqref{eq:second-order3}]
    We have
    \begin{align*}
      \|(J_{L,t}-J_{L,t, t+1})J_{L,t}^{\top} (g_{L,t} - g_{L}^{\star})\|_2  \leq \frac{1}{9} \| f_t - y \|_2
    \end{align*}
\end{claim}
\begin{proof}
We can show
	\begin{align}
	\label{eq:second-order4}
	\|(J_{L,t}-J_{L,t, t+1})J_{L,t}^{\top} (g_{L,t} - g_{L}^{\star})\|_2  \leq\|J_{L,t} - J_{L,t, t+1}\| \cdot \|J_{L,t}^{\top}\| \cdot \|g_{L,t} - g_{L}^{\star}\|_2.
	\end{align}

	Moreover, one has
	\begin{align}\label{eq:second-order2}
	\|g_{L,t} - g_{L}^{\star}\|_2
	\leq & ~ \epsilon_0\, \|(J_{L,t} J_{L,t}^{\top})^{-1}\| \cdot \|f_t - y\|_2 \notag \\
	\leq & ~ \frac{2\epsilon_0}{\lambda_L} \cdot \|f_t - y\|_2 \notag \\
	= & ~ \frac{1}{15\sqrt{n\lambda_L}} \cdot \|f_t - y\|_2 .
	\end{align}
	The first step is the second guarantee of Theorem~\ref{thm:fast_kronecker_regression} (with target vector $f_t-y$ and precision $\epsilon_0$). The second step comes from $\lambda_{\min}(J_{L,t} J_{L,t}^{\top}) = \lambda_{\min}(G_L(W(t))) \geq \lambda_L / 2$ (see Corollary~\ref{cor:eigen_perturb}).
 The last step follows from $\epsilon_0 = \sqrt{\lambda_L/n}/30$ (Definition~\ref{def:R}).

	 Consequently, using Eq.~\eqref{eq:second-order6} and~\eqref{eq:second-order2} and $\|J_{L,t}\|\leq\|J_{L,t}\|_F\leq 2\sqrt n$ (Lemma~\ref{lem:perturb}), we have
	\begin{align}\label{eq:second-order8}
	\|(J_{L,t}-J_{L,t, t+1})J_{L,t}^{\top} (g_{L,t} - g_{L}^{\star})\|_2
	\leq & ~ \|J_{L,t} - J_{L,t, t+1}\| \cdot \|J_{L,t}^{\top}\| \cdot \| g_{L,t} - g_{L}^{\star} \|_2 \notag \\
	\leq & ~2\sqrt{54\,n\zeta} \cdot 2\sqrt n \cdot \frac{1}{15\sqrt{n\lambda_L}} \cdot \| f_t - y \|_2 \notag \\
	= & ~ \frac{4}{15}\Big(\frac{54\,n\zeta}{\lambda_L}\Big)^{1/2}\cdot \|f_t-y\|_2 \notag\\
	\leq & ~ \frac{1}{9}\|f_t-y\|_2.
	\end{align}
	The last step follows from~\eqref{eq:zeta_small}.
\end{proof}

\begin{lemma}[Putting it all together]
We have
\begin{align}
	\label{eq:second-order12}
	\|f_{t + 1} - y\|_2 \leq \frac{1}{3}\|f_t - y\|_2.
	\end{align}
\end{lemma}
\begin{proof}
Combining Eq.~\eqref{eq:second-order3}, \eqref{eq:second-order5}, \eqref{eq:second-order11}, and \eqref{eq:second-order8}, we have proved the second claim, i.e.,
	\begin{align*}
	\|f_{t + 1} - y\|_2 \leq \frac{1}{3}\|f_t - y\|_2.
	\end{align*}
\end{proof}

By induction on $t$, the two claims hold for all $t\leq T$ on the intersection of the events (E1) to (E4), which proves Theorem~\ref{thm:conv_app}.

\subsection{Bounds on the Movement of Weights}\label{sec:movement}

\begin{lemma}
Let $\wt R$ be chosen as in Definition~\ref{def:R}, and suppose that the two induction hypotheses hold up to time $t$. Then the following holds for every $r\in[m]$:
\begin{align*}
    \|\wt W_{L,r}(t+1)-\wt W_{L,r}(0)\|_2 \leq & ~ \wt R.
\end{align*}
\end{lemma}
\begin{proof}
 First, we have
	 	 \begin{align}\label{eq:second-order7}
	 	 \|g_{L,t}\|_2
	 	 \leq & ~ \|g^{\star}_L\|_2 + \|g_{L,t} - g^{\star}_L\|_2 \notag \\
	 	 = & ~ \|(J_{L,t}J_{L,t}^{\top})^{-1}(f_t - y)\|_2 + \|g_{L,t} - g_L^{\star}\|_2 \notag \\
	 	 \leq & ~ \|(J_{L,t}J_{L,t}^{\top})^{-1}\| \cdot \|(f_t - y)\|_2 + \|g_{L,t} - g_L^{\star}\|_2 \notag \\
	 	 \leq & ~ \frac{2}{\lambda_L}\cdot \|f_t - y\|_2 + \frac{1}{15\sqrt{n\lambda_L}} \cdot \|f_t - y\|_2 \notag \\
	 	 \leq & ~ \frac{3}{\lambda_L} \cdot \|f_t - y\|_2
	 	 \end{align}
	 where the fourth step is owing to Corollary~\ref{cor:eigen_perturb} and Eq.~\eqref{eq:second-order2}, and the final step is due to the  fact that $1/\sqrt{n\lambda_L} \leq 1/\lambda_L$. The same bound holds at every time $t'\leq t$.

    The $i$-th row of $J_{L,t'}$, viewed as an $m\times m$ matrix, is $\sqrt{2/m}\,D_{i,L}(t')ah_{i,L-1}^\top$, and its $r$-th row is $\frac{1}{\sqrt{mG(b)}}\,a_r{\bf 1}[r\in S_i(\wt W_L(t'))]\,h_{i,L-1}^\top$, since $\sqrt{2c_b/m}=1/\sqrt{mG(b)}$. Here $S_i$ is the set of neurons of the last layer that fire on $x_i$ (Lemma~\ref{lem:flip}). This large factor $1/\sqrt{G(b)}$ is the source of the factor $1/\sqrt{G(b)}$ in $\wt R$. Hence, for $t'\leq t$,
    \begin{align*}
        \| \wt W_{L,r}(t'+1)-\wt W_{L,r}(t')\|_2 = & ~ \Big\| \sum_{i=1}^n \frac{1}{\sqrt{mG(b)}}\, a_r {\bf 1}[r\in S_i(\wt W_L(t'))]\, g_{L,t',i}\, h_{i,L-1} \Big\|_2 \\
        \leq & ~ \frac{1.1}{\sqrt{mG(b)}} \sum_{i=1}^n |g_{L,t',i}| \\
        \leq & ~ \frac{1.1\sqrt n}{\sqrt{mG(b)}} \cdot  \|g_{L,t'}\|_2 \\
        \leq & ~ \frac{3.3\sqrt n}{\lambda_L\sqrt{mG(b)}} \cdot \|f_{t'}-y\|_2 \\
        \leq & ~  \frac{3.3\sqrt n}{3^{t'}\lambda_L\sqrt{mG(b)}}\cdot \|f_0-y\|_2.
    \end{align*}
    The first step follows from the update rule $\vect(\wt W_L(t'+1)-\wt W_L(t'))=-J_{L,t'}^\top g_{L,t'}$, the second step is by the triangle inequality and $\|h_{i,L-1}\|_2\leq 1.1$, the third step uses the fact the $\ell_1$ norm of a vector is upper bounded by $\sqrt n$ times the $\ell_2$ norm, the fourth step is by Eq.~\eqref{eq:second-order7}, and the last step is by Induction Hypothesis 2 up to time $t'\leq t$.

    Consequently, using $\|f_0-y\|_2\leq\sqrt n\log^c(nm)$ (Definition~\ref{def:R}), we have
    \begin{align*}
        \|\wt W_{L,r}(t+1)-\wt W_{L,r}(0)\|_2
        \leq & ~ \sum_{t'=0}^t \|\wt W_{L,r}(t'+1)-\wt W_{L,r}(t')\|_2 \\
        \leq & ~ \sum_{t'=0}^t \frac{3.3\, n\log^c(nm)}{3^{t'}\lambda_L\sqrt{mG(b)}} \\
        \leq & ~ \frac{5\, n\log^c(nm)}{\lambda_L\sqrt{mG(b)}}=\wt R.
    \end{align*}
    This concludes our proof.
    \end{proof}

\subsection{The Exact Gram Solve}\label{sec:exact_solve}

The algorithms of Theorems~\ref{thm:plain_gram} and~\ref{thm:plain_incremental} solve the Gram regression exactly. We show that the proof of Theorem~\ref{thm:conv_app} also applies to them. For $t\geq 0$, let $J'_{L,t}\in\R^{n\times m^2}$ be the Jacobian of the last layer with respect to $W_L$ at $W_L(t)$, whose $i$-th row is $\vect(u_{i,L}(t)h_{i,L-1}^\top)^\top$ with $u_{i,L}(t)=D_{i,L}(t)a$. If $J'_{L,t}J'^\top_{L,t}$ is invertible, then the Gauss-Newton step with the exact Gram solve is
\begin{align}\label{eq:exact_step}
    g'(t):=(J'_{L,t}J'^\top_{L,t})^{-1}(f_t-y),\qquad W_L(t+1):=W_L(t)-\sum_{i=1}^n g'_i(t)\,u_{i,L}(t)h_{i,L-1}^\top.
\end{align}
This is the update of Algorithm~\ref{alg:complete} when its solver returns the exact solution.

\begin{lemma}[Convergence with the exact Gram solve]\label{lem:exact_solve}
Suppose that the width $m$ satisfies the width condition of Definition~\ref{def:m}, and that the events (E1), (E2) and (E3) hold. Then, for every $t\geq 0$, the iterate $W_L(t)$ of~\eqref{eq:exact_step} is defined and satisfies Induction hypothesis~1, the matrix $J'_{L,t}J'^\top_{L,t}$ is invertible, and $\|f_{t+1}-y\|_2\leq\frac{1}{3}\|f_t-y\|_2$.
\end{lemma}

\begin{proof}
We follow the proof of Theorem~\ref{thm:conv_app}. As there, let $J_{L,t}=\sqrt{2/m}\,J'_{L,t}$ be the Jacobian with respect to $\wt W_L$, and let $g_{L,t}:=\frac{m}{2}g'(t)$. Then
\begin{align*}
    g_{L,t}=\frac{m}{2}(J'_{L,t}J'^\top_{L,t})^{-1}(f_t-y)=(J_{L,t}J_{L,t}^\top)^{-1}(f_t-y)=g^\star_L,
\end{align*}
and~\eqref{eq:exact_step} is the update rule $\vect(\wt W_L(t+1)-\wt W_L(t))=-J_{L,t}^\top g_{L,t}$ of that proof. That proof uses the output of the solver only through $g_{L,t}$, and only in three places. The first guarantee of Theorem~\ref{thm:fast_kronecker_regression} with precision $\epsilon_0$ is used in the claim on the first term of~\eqref{eq:second-order3}. The second guarantee is used in~\eqref{eq:second-order2}, and through it in the claim on the third term and in~\eqref{eq:second-order7}. The third place is the update rule. Since $g_{L,t}=g^\star_L$, both guarantees hold with left-hand side zero, and the update rule holds. The event (E4) is used only for these guarantees, so the induction goes through on the events (E1) to (E3), as long as $J'_{L,t}J'^\top_{L,t}$ is invertible at every step. In the induction, Induction hypothesis~1 at time $t$ is established before step $t$ is analyzed. Then Corollary~\ref{cor:eigen_perturb} gives $\lambda_{\min}(J_{L,t}J_{L,t}^\top)\geq\lambda_L/2>0$, so $J'_{L,t}J'^\top_{L,t}=\frac{m}{2}J_{L,t}J_{L,t}^\top$ is invertible and step $t$ is defined. The induction never uses $T$, so it gives the claims for every $t\geq 0$.
\end{proof}

\section{An Incremental Implementation of Last-Layer Training}\label{sec:plain_incremental}

In this section, we show that the Gauss-Newton method for the last layer can be implemented with a cost per step that is sublinear in the width $m$ for every constant $\gamma>1/3$, which our width condition allows only at widths $m>n^6$. The implementations of Theorems~\ref{thm:plain_dense} and~\ref{thm:plain_gram} still touch all $m$ neurons of the last layer at every step. For instance, the method that maintains the matrix $W_LH$, whose columns are the pre-activations $W_Lh_{i,L-1}$ of the last layer, updates all $m$ rows of this matrix. We observe that a step changes only the rows of the neurons that fire on at least one data point, and that a neuron that fires on no data point never fires again. On the events (E1) to (E3) of Section~\ref{sec:analysis}, each data point makes at most $O(k)$ neurons fire at every step. So it suffices to keep track of $O(nk)$ rows, and this takes $O(n^2k)$ time per step.

\paragraph*{Notation.} Throughout this section, we write $h_i:=h_{i,L-1}$ and $\tau_i:=\tau_{i,L}$ for $i\in[n]$, $H:=[h_1,\dots,h_n]\in\R^{m\times n}$ and $K:=H^\top H\in\R^{n\times n}$. These do not change during training (Remark~\ref{remark:threshold}). For a step $t$, let $Z(t):=W_L(t)H\in\R^{m\times n}$, whose $i$-th column is the pre-activation $W_L(t)h_i$ of the last layer on $x_i$. Let
\begin{align*}
    A_i(t):=\{r\in[m]:Z_{r,i}(t)>\tau_i\}~~\text{for}~i\in[n],\qquad {\cal R}_t:=\bigcup_{i=1}^nA_i(t),\\
    B_r(t):=\{i\in[n]:r\in A_i(t)\}~~\text{for}~r\in {\cal R}_t.
\end{align*}
So $A_i(t)$ is the set of neurons of the last layer that fire on $x_i$, ${\cal R}_t$ is the set of neurons that fire on at least one data point, and $B_r(t)$ is the set of data points on which neuron $r$ fires. By Definition~\ref{def:architecture}, $D_{i,L}(t)=\sqrt{c_b}\,\diag({\bf 1}_{A_i(t)})$, so the vector $u_i(t):=D_{i,L}(t)a$ has the entries $(u_i(t))_r=\sqrt{c_b}\,a_r{\bf 1}[r\in A_i(t)]$. Writing $W_L=\sqrt{2/m}\,\wt W_L$ as in Section~\ref{sec:spectral_app}, for $h_i\neq 0$ we have $Z_{r,i}(t)>\tau_i=\sqrt{2/m}\,b\,\|h_i\|_2$ if and only if $\langle\wt W_{L,r}(t),\wh h_i\rangle>b$, where $\wh h_i=h_i/\|h_i\|_2$. Hence, if $h_i\neq 0$, which holds on the event (E1) of Section~\ref{sec:analysis}, $A_i(t)=S_i(\wt W_L(t))$ in the notation of Lemma~\ref{lem:flip}.

Let $J'_t\in\R^{n\times m^2}$ be the Jacobian of the last layer with respect to $W_L$ at step $t$. Its $i$-th row is $\vect(u_i(t)h_i^\top)^\top$. The Gauss-Newton step with the exact Gram solve is
\begin{align*}
    g'(t):=(J'_tJ'^\top_t)^{-1}(f_t-y),\qquad W_L(t+1):=W_L(t)-\sum_{i=1}^n g'_i(t)\,u_i(t)h_i^\top.
\end{align*}
This is the update of Algorithm~\ref{alg:complete} when the solver returns the exact solution. Let $\wt U_t\in\R^{m\times n}$ be the matrix whose $i$-th column is $-g'_i(t)\,u_i(t)$, so that $W_L(t+1)=W_L(t)+\wt U_tH^\top$, and let $\Gamma_t:=\sum_{s=0}^{t-1}\wt U_s$, with $\Gamma_0=0$.

\begin{algorithm}[H]\caption{Incremental training of the last layer.}
\label{alg:plain_incremental}
\begin{algorithmic}[1]
\Procedure{IncrementalTraining}{$X\in\R^{d\times n},y\in\R^n,T$} \Comment{Theorem~\ref{thm:plain_incremental}}
    \State Initialize $W_1(0),\dots,W_L(0)$ and $a$, and compute $h_i,\tau_i$ for $i\in[n]$ \Comment{As in Algorithm~\ref{alg:complete}}
    \State $K\gets H^\top H$, $Z\gets W_L(0)H$ and $\Gamma\gets 0\in\R^{m\times n}$
    \State $A_i\gets\{r\in[m]:Z_{r,i}>\tau_i\}$ for $i\in[n]$
    \For{$t=0\to T-1$}
        \State ${\cal R}\gets\bigcup_{i=1}^nA_i$, and $B_r\gets\{i\in[n]:r\in A_i\}$ for $r\in {\cal R}$
        \State $f_i\gets\sqrt{c_b}\sum_{r\in A_i}a_r(Z_{r,i}-\tau_i)$ for $i\in[n]$
        \State $Q_{ij}\gets c_b\,K_{ij}\,|\{r\in {\cal R}:i\in B_r\text{ and }j\in B_r\}|$ for $i,j\in[n]$ \Comment{The count is $|A_i\cap A_j|$}
        \State $g'\gets Q^{-1}(f-y)$, and stop if $Q$ is singular
        \For{$r\in {\cal R}$}
            \State $\wt U_{r,i}\gets-\sqrt{c_b}\,a_r\,g'_i$ for $i\in B_r$
            \State $Z_{r,j}\gets Z_{r,j}+\sum_{i\in B_r}\wt U_{r,i}K_{i,j}$ for $j\in[n]$, and $\Gamma_{r,i}\gets\Gamma_{r,i}+\wt U_{r,i}$ for $i\in B_r$
        \EndFor
        \State $A_i\gets\{r\in {\cal R}:Z_{r,i}>\tau_i\}$ for $i\in[n]$
    \EndFor
    \State \Return $W_L(0)+\Gamma H^\top$
\EndProcedure
\end{algorithmic}
\end{algorithm}

Algorithm~\ref{alg:plain_incremental} uses no randomness besides the initialization. It never forms $W_L(t)$ during training, and it uses neither the low-rank data structure nor the sketch. Our main result in this section is the following.

\begin{theorem}[Incremental last-layer training]\label{thm:plain_incremental}
Let $b=\sqrt{2\gamma\log m}$ for a constant $\gamma\in(0,1)$, so that $k=m^{1-\gamma}$, and suppose that the width $m$ satisfies the condition of Theorem~\ref{thm:conv_main} for this $\gamma$. Assume $n\geq 2$, as in Section~\ref{sec:analysis}. Let $\delta_0:=e^{-\log^2 n}$, and let $N$ be the bound of Lemma~\ref{lem:flip} for the radius $\wt R$ of Definition~\ref{def:R} and $\delta=\delta_0$, that is, $N=4e\,m\wt R\,p(b)+4\log(n/\delta_0)$. Then $N\leq k/n$. Moreover, with probability at least $1-e^{-\Omega(\log^2 n)}$ over the initialization, Algorithm~\ref{alg:plain_incremental} has the following properties for every number of steps $T\geq 1$.
\begin{itemize}
    \item[(i)] It computes the Gauss-Newton iterates with the exact Gram solve, in the sense that $Z=W_L(t)H$ at the start of iteration $t$, and it returns $W_L(T)$. Algorithm~\ref{alg:plain_gram} computes the same iterates, in the sense that its matrix $Z$ also equals $W_L(t)H$ at the start of step $t$, and it also returns $W_L(T)$. Moreover, $\|f_{t+1}-y\|_2\leq\frac{1}{3}\|f_t-y\|_2$ for all $t\in\{0,\dots,T-1\}$.
    \item[(ii)] For all $t\in\{0,\dots,T\}$ and $i\in[n]$, we have $|A_i(t)\triangle A_i(0)|\leq N$ and $|A_i(t)|\leq 2k+N$.
    \item[(iii)] For all $t\in\{0,\dots,T-1\}$, step $t$ changes $W_L$ and $Z=W_LH$ only in the rows in ${\cal R}_t$, and ${\cal R}_{t+1}\subseteq {\cal R}_t$.
    \item[(iv)] The initialization takes $O({\cal T}_{\mathrm{init}})=O(m^2nL)$ time, every step takes $O(n^2k)$ time, with no amortization, and forming the output $W_L(T)$ takes $O(nmk)$ time.
\end{itemize}
In particular, the cost per step is $O(n^2k)$, and $n^2k\leq m^{3(1-\gamma)/2}$. This is $o(m)$ for every constant $\gamma>1/3$, and it is $O(m^{3/4})$ for $\gamma=1/2$.
\end{theorem}

We first collect the identities behind the algorithm. They hold for every choice of the vectors $g'(t)$, not only for the Gauss-Newton step.

\begin{lemma}\label{lem:plain_incremental_rows}
Let $W_L(0)\in\R^{m\times m}$ and let $g'(0),g'(1),\ldots\in\R^n$ be arbitrary vectors. Define $W_L(t+1):=W_L(t)-\sum_{i=1}^ng'_i(t)\,u_i(t)h_i^\top$ for $t\geq 0$, and define $Z(t)$, $A_i(t)$, ${\cal R}_t$, $B_r(t)$, $\wt U_t$ and $\Gamma_t$ as above. Then the following hold for every $t\geq 0$.
\begin{itemize}
    \item[(a)] $Z(t+1)=Z(t)+\wt U_tK$ and $W_L(t)=W_L(0)+\Gamma_tH^\top$.
    \item[(b)] For $r\in {\cal R}_t$ and $i\in[n]$, we have $(\wt U_t)_{r,i}=-\sqrt{c_b}\,a_r\,g'_i(t)$ if $i\in B_r(t)$ and $(\wt U_t)_{r,i}=0$ otherwise. Every row of $\wt U_t$ outside ${\cal R}_t$ is zero. Hence $W_L(t+1)$ and $Z(t+1)$ agree with $W_L(t)$ and $Z(t)$ in every row outside ${\cal R}_t$.
    \item[(c)] For every $i\in[n]$, we have $A_i(t+1)=\{r\in {\cal R}_t:Z_{r,i}(t+1)>\tau_i\}$. In particular, ${\cal R}_{t+1}\subseteq {\cal R}_t\subseteq {\cal R}_0$, and every row of $\Gamma_t$ outside ${\cal R}_0$ is zero.
    \item[(d)] For every $i\in[n]$, we have $f(W(t),x_i)=\sqrt{c_b}\sum_{r\in A_i(t)}a_r\,(Z_{r,i}(t)-\tau_i)$.
    \item[(e)] For all $i,j\in[n]$, we have $(J'_tJ'^\top_t)_{i,j}=c_b\,|A_i(t)\cap A_j(t)|\,K_{i,j}$.
\end{itemize}
\end{lemma}

\begin{proof}
(a) Multiplying $W_L(t+1)=W_L(t)+\wt U_tH^\top$ on the right by $H$ gives $Z(t+1)=Z(t)+\wt U_tK$. The second identity follows by induction on $t$, since $\Gamma_0=0$ and $\Gamma_{t+1}=\Gamma_t+\wt U_t$.

(b) The $i$-th column of $\wt U_t$ is $-g'_i(t)\,u_i(t)$, whose $r$-th entry is $-\sqrt{c_b}\,a_r\,g'_i(t)\,{\bf 1}[r\in A_i(t)]$. For $r\in {\cal R}_t$, we have $r\in A_i(t)$ if and only if $i\in B_r(t)$. For $r\notin {\cal R}_t$, we have $r\notin A_i(t)$ for every $i$, so the $r$-th row of $\wt U_t$ is zero. Then the $r$-th rows of $\wt U_tH^\top$ and of $\wt U_tK$ are zero, and the last claim follows from $W_L(t+1)=W_L(t)+\wt U_tH^\top$ and from (a).

(c) Let $r\in[m]\setminus {\cal R}_t$ and $i\in[n]$. Since $r\notin A_i(t)$, we have $Z_{r,i}(t)\leq\tau_i$. By (b), $Z_{r,i}(t+1)=Z_{r,i}(t)\leq\tau_i$, so $r\notin A_i(t+1)$. Hence $A_i(t+1)\subseteq {\cal R}_t$, which gives the formula for $A_i(t+1)$. Taking the union over $i$ gives ${\cal R}_{t+1}\subseteq {\cal R}_t$, and induction on $t$ gives ${\cal R}_t\subseteq {\cal R}_0$. By (b), every row of $\wt U_s$ outside ${\cal R}_s\subseteq {\cal R}_0$ is zero, so the same holds for $\Gamma_t=\sum_{s<t}\wt U_s$.

(d) By Definition~\ref{def:architecture}, $f(W(t),x_i)=a^\top\phi_L(W_L(t)h_i)=\sqrt{c_b}\sum_{r=1}^ma_r\max\{Z_{r,i}(t)-\tau_i,0\}$, and the $r$-th term is zero unless $r\in A_i(t)$.

(e) The $i$-th row of $J'_t$ is $\vect(u_ih_i^\top)^\top$, and $\langle u_ih_i^\top,u_jh_j^\top\rangle=\tr[h_iu_i^\top u_jh_j^\top]=(u_i^\top u_j)(h_i^\top h_j)$. Since $a_r^2=1$, we have $u_i(t)^\top u_j(t)=c_b\sum_{r\in A_i(t)\cap A_j(t)}a_r^2=c_b\,|A_i(t)\cap A_j(t)|$, and $h_i^\top h_j=K_{i,j}$.
\end{proof}

\begin{lemma}\label{lem:plain_incremental_exact}
Let $T\geq 1$, let $W_L(t)$ be the Gauss-Newton iterates with the exact Gram solve, and suppose that $J'_tJ'^\top_t$ is invertible for every $t\in\{0,\dots,T-1\}$. Then, at the start of iteration $t\in\{0,\dots,T-1\}$ of Algorithm~\ref{alg:plain_incremental}, we have $Z=Z(t)$, $A_i=A_i(t)$ for all $i\in[n]$ and $\Gamma=\Gamma_t$. In iteration $t$, we have ${\cal R}={\cal R}_t$, $B_r=B_r(t)$, $f=f_t$, $Q=J'_tJ'^\top_t$ and $g'=g'(t)$. The algorithm returns $W_L(T)$.
\end{lemma}

\begin{proof}
We use induction on $t$. After the initialization, $Z=W_L(0)H=Z(0)$, $A_i=A_i(0)$ by definition and $\Gamma=0=\Gamma_0$. Suppose that $Z=Z(t)$, $A_i=A_i(t)$ and $\Gamma=\Gamma_t$ at the start of iteration $t$. Then ${\cal R}={\cal R}_t$ and $B_r=B_r(t)$ by definition, and $f=f_t$ by Lemma~\ref{lem:plain_incremental_rows}(d). Since $A_i(t)\cap A_j(t)\subseteq {\cal R}_t$, the count in the definition of $Q_{ij}$ is $|A_i(t)\cap A_j(t)|$, so $Q=J'_tJ'^\top_t$ by Lemma~\ref{lem:plain_incremental_rows}(e). Since $Q$ is invertible, the algorithm does not stop, and $g'=g'(t)$. By Lemma~\ref{lem:plain_incremental_rows}(b), the entries $\wt U_{r,i}$ set in the inner loop are the entries $(\wt U_t)_{r,i}$ with $r\in {\cal R}_t$ and $i\in B_r(t)$, and all other entries of $\wt U_t$ are zero. So the inner loop adds the $r$-th row of $\wt U_tK$ to the $r$-th row of $Z$ and the $r$-th row of $\wt U_t$ to the $r$-th row of $\Gamma$ for every $r\in {\cal R}_t$, and it leaves the other rows unchanged. By Lemma~\ref{lem:plain_incremental_rows}(a) and~(b), this gives $Z=Z(t+1)$ and $\Gamma=\Gamma_{t+1}$. By Lemma~\ref{lem:plain_incremental_rows}(c), the last line of the iteration gives $A_i=A_i(t+1)$. Finally, the algorithm returns $W_L(0)+\Gamma_TH^\top=W_L(T)$ by Lemma~\ref{lem:plain_incremental_rows}(a).
\end{proof}

\begin{proof}[Proof of Theorem~\ref{thm:plain_incremental}]
The condition of Theorem~\ref{thm:conv_main} implies the width condition of Definition~\ref{def:m} (see the paragraph after that definition), so the discussion after the events (E1) to (E4) in Section~\ref{sec:analysis} applies.

\textbf{The bound on $N$.} Since $m\,p(b)=m\,e^{-b^2/2}/\sqrt{2\pi}=k/\sqrt{2\pi}$ and $4e/\sqrt{2\pi}\leq 5$, the first term of $N$ is at most $5k\wt R$. Both terms of $\zeta=\wt R+(1+b)\log(n/\delta_0)/k$ are nonnegative, so by~\eqref{eq:zeta_small} each of them is at most $\lambda_L/(40000\,n)$, and $\lambda_L\leq 1/(1+b^2)\leq 1$ by Theorem~\ref{thm:conv_main}. Hence $\wt R\leq 1/(40000\,n)$ and $\log(n/\delta_0)\leq k/(40000\,n)$, and
\begin{align*}
    N\leq 5k\wt R+4\log(n/\delta_0)\leq\frac{5k}{40000\,n}+\frac{4k}{40000\,n}\leq\frac{k}{n}.
\end{align*}
This bound does not depend on the initialization.

Let ${\cal E}$ be the intersection of the events (E1), (E2) and (E3) of Section~\ref{sec:analysis}. These events concern only the initialization, and by the discussion after their definition, ${\cal E}$ fails with probability at most $e^{-\Omega(\log^2 n)}$. We work on ${\cal E}$ from now on. None of the arguments below depends on $T$.

\textbf{Convergence.} The Gauss-Newton iterates with the exact Gram solve defined above are the iterates~\eqref{eq:exact_step} of Section~\ref{sec:exact_solve}, with $J'_t=J'_{L,t}$. On ${\cal E}$, Lemma~\ref{lem:exact_solve} shows that, for every $t\geq 0$, Induction hypothesis~1 holds at time $t$, the matrix $J'_tJ'^\top_t$ is invertible, and $\|f_{t+1}-y\|_2\leq\frac{1}{3}\|f_t-y\|_2$. By Lemma~\ref{lem:plain_incremental_exact}, Algorithm~\ref{alg:plain_incremental} computes these iterates and returns $W_L(T)$. Since the Gram matrix is invertible at every step, Algorithm~\ref{alg:plain_gram} does not stop either, and the claim in the proof of Theorem~\ref{thm:plain_gram} shows that its matrix $Z$ equals $W_L(t)H$ at the start of step $t$ and that it returns $W_L(T)$. This proves (i).

\textbf{Active sets.} Fix $t\geq 0$. By Induction hypothesis~1 at time $t$, every row of $\wt W_L(t)$ is within distance $\wt R$ of the corresponding row of $\wt W_L(0)$. The event (E3) is the event of Lemma~\ref{lem:flip}, which holds for all such matrices at once, so it applies at every step without a union bound over $t$. Its hypotheses hold, since $h_i\neq 0$ by $\|h_i\|_2\geq 0.9$ in (E1), and $b\wt R\leq 1$ by the discussion after~\eqref{eq:zeta_small}. Since $A_i(s)=S_i(\wt W_L(s))$ for every $s$, the event (E3) gives $|A_i(t)\triangle A_i(0)|\leq N$ for all $i\in[n]$. By (E1) and Lemma~\ref{lem:init_sparse}, $|A_i(0)|=\|D_{i,L}(0)\|_0\leq 2mP_b\leq 2k$. Hence
\begin{align*}
    |A_i(t)|\leq|A_i(0)|+|A_i(t)\setminus A_i(0)|\leq 2k+N.
\end{align*}
This proves (ii).

\textbf{Rows.} Claim (iii) is Lemma~\ref{lem:plain_incremental_rows}(b) and~(c). It holds for every initialization.

\textbf{Cost.} We store $W_L(0)$, $Z$ and $\Gamma$ as dense arrays, each $A_i$ as a list, and the lists $B_r$ in an array indexed by $r\in[m]$. We also keep an array of $m$ flags that marks the members of ${\cal R}$. These two arrays are allocated once during the initialization, and at the end of each iteration we reset only the entries for $r\in {\cal R}$.

The initialization first does what Algorithm~\ref{alg:complete} does before its data structure is created, in $O({\cal T}_{\mathrm{init}})$ time by the proof of Theorem~\ref{thm:alg_app}. Computing $K=H^\top H$ takes $O(n^2m)$ time. Computing $Z=W_L(0)H$ takes $O(nmk)$ time, since each $h_i$ has at most $2k$ nonzero entries by (E1) and Lemma~\ref{lem:init_sparse}. The sets $A_i(0)$, the matrix $\Gamma=0$ and the two arrays take $O(nm)$ time. All of this is $O({\cal T}_{\mathrm{init}})$.

Consider iteration $t$ and let $q_t:=\sum_{i=1}^n|A_i(t)|$. By (ii) and $N\leq k/n$, we have $q_t\leq n(2k+N)\leq 2nk+k\leq 3nk$. We also have $|{\cal R}_t|\leq q_t$, $|B_r(t)|\leq n$ and $\sum_{r\in {\cal R}_t}|B_r(t)|=q_t$.
\begin{itemize}
    \item One pass over the lists $A_i$ builds ${\cal R}$ and the lists $B_r$, in $O(q_t)$ time. Computing $f$ also takes $O(q_t)$ time.
    \item We compute the counts $|\{r\in {\cal R}_t:i\in B_r(t)\text{ and }j\in B_r(t)\}|$ by adding $1$ to the count of the pair $(i,j)$ for every $r\in {\cal R}_t$ and every pair $(i,j)\in B_r(t)\times B_r(t)$. This takes $O(n^2+\sum_{r\in {\cal R}_t}|B_r(t)|^2)\leq O(n^2+n\,q_t)$ time. Forming $Q$ from the counts and $K$ takes $O(n^2)$ time.
    \item Gaussian elimination decides whether $Q$ is singular and, if it is not, gives $g'$ in $O(n^3)$ time. By the convergence part and Lemma~\ref{lem:plain_incremental_exact}, $Q=J'_tJ'^\top_t$ is positive definite, so the algorithm does not stop.
    \item In the loop over $r\in {\cal R}_t$, row $r$ takes $O(n\,|B_r(t)|)$ time, so the loop takes $O(n\,q_t)$ time.
    \item The new sets $A_i$ are found by scanning ${\cal R}_t$ once for each $i$, in $O(n\,|{\cal R}_t|)\leq O(n\,q_t)$ time. Resetting the flags and the lists takes $O(|{\cal R}_t|)$ time.
\end{itemize}
So iteration $t$ takes $O(n\,q_t+n^3)=O(n^2k+n^3)$ time. As in the proof of Corollary~\ref{cor:alg_cost}, the width condition gives $k=m^{1-\gamma}\geq n^4$, so $n^3\leq n^2k$, and each step takes $O(n^2k)$ time. Moreover, $n\leq k^{1/4}$ gives $n^2k\leq k^{3/2}=m^{3(1-\gamma)/2}$, and $3(1-\gamma)/2<1$ for $\gamma>1/3$.

Finally, the output is $W_L(0)+\Gamma_TH^\top=W_L(0)+\sum_{i=1}^n(\Gamma_T)_{:,i}h_i^\top$. The $i$-th term has nonzero entries only in the at most $2k$ columns in the support of $h_i$, so adding it to the stored $W_L(0)$ takes $O(mk)$ time. The total is $O(nmk)$. This proves (iv) and the theorem.
\end{proof}

\begin{remark}
We collect some consequences of Theorem~\ref{thm:plain_incremental}.
\begin{itemize}
    \item The bound $O(n^2k)$ uses that the update $\wt U_t$ has at most $q_t=O(nk)$ nonzero entries, so that the product $\wt U_tK$ takes $O(n\,q_t)$ time. Fast matrix multiplication is not needed. Multiplying the rows ${\cal R}_t$ of $\wt U_t$ by $K$ as dense matrices costs $\Tmat(|{\cal R}_t|,n,n)$, and with $|{\cal R}_t|\leq 3nk$ this gives only the weaker bound $n^{\omega+o(1)}k$.
    \item Given the invertibility of $Q$ from the convergence part, Lemma~\ref{lem:flip} enters the cost analysis only through the bound $|A_i(t)|\leq 2k+N$. Without it, Lemma~\ref{lem:plain_incremental_rows}(c) and Lemma~\ref{lem:init_sparse} still give $|A_i(t)|\leq|{\cal R}_t|\leq|{\cal R}_0|\leq 2nk$, so $q_t\leq 2n^2k$ and each step takes $O(n^3k)$ time.
    \item Theorem~\ref{thm:alg_main} allows every constant $\gamma\in[a,1)$, where $a=\min\{\alpha,1/2\}$. For $\gamma=1/2$, the cost per step here is $O(m^{3/4})$. For $\gamma=a$, it is at most $m^{3(1-a)/2}$, which is $O(m^{0.8025})$ by the bound $\alpha>0.465$ over every field of characteristic zero~\cite[Theorem~1.1 and Corollaries~14.1--14.2]{OpenAIMM26}. The bounds on the cost per step in Theorems~\ref{thm:alg_main}, \ref{thm:plain_dense} and~\ref{thm:plain_gram} are at least linear in $m$. For $\gamma>1/3$, the bound here is sublinear in $m$ because a step reads and writes only the rows of $Z$ of the $O(nk)$ neurons in ${\cal R}_t$.
    \item The initialization still takes $O({\cal T}_{\mathrm{init}})$ time. We do not improve it.
\end{itemize}
\end{remark}

\section{Bounds on the Intermediate Layer Output with Shifted ReLU}
\label{sec:modified_als}
In this section, we prove a technical lemma (Lemma~\ref{lem:als_71}) involving truncated Gaussian distribution, which correlates to the shifted ReLU activation we use, with the threshold of each layer proportional to the norm of the layer input.

\begin{definition}[Truncated Gaussian distribution]
Suppose $X\sim \mathcal{N}(0, \sigma^2)$. Let $b\in \R$. Then, we say a random variable $Y$ follows from a truncated Gaussian distribution $\mathcal{N}_{b}(0, \sigma^2)$ if $Y = X | X\geq b$. The probability density function for $\mathcal{N}_{b}(0, \sigma^2)$ is as follows:
\begin{align*}
    f(y) = \frac{1}{\sigma (1-\Phi(b/\sigma))}\cdot \frac{1}{\sqrt{2\pi}}e^{-y^2/(2\sigma^2)}~~~y\in [b, \infty),
\end{align*}
where $\Phi(\cdot)$ is the standard Gaussian distribution's CDF.  
\end{definition}

\begin{fact}[Properties of truncated Gaussian distribution]
For $b\in \R$, suppose $X\sim \mathcal{N}_b(0, \sigma^2)$. Let $\beta := b/\sigma$. Then, we have
\begin{itemize}
    \item $\E[X] = \frac{\sigma \phi(\beta)}{1-\Phi(\beta)}$, where $\phi(x):=\frac{1}{\sqrt{2\pi}}e^{-x^2/2}$.
    \item $\Var[X] = \sigma^2(1+\beta\phi(\beta)/(1-\Phi(\beta)) - (\phi(\beta)/(1-\Phi(\beta)))^2)$.
    \item $X/\sigma \sim \mathcal{N}_{b/\sigma}(0, 1)$.
    \item When $\sigma=1$, $X$ is $C(b+1)$-subgaussian, where $C>0$ is an absolute constant.
\end{itemize}
\end{fact}

\begin{lemma}[Concentration inequality for $b$-truncated chi-square distribution]\label{lem:truncated_chi_concentration}
For $b\in \R$, $n>0$, let $X\sim \chi_{b,n}^2$; that is, $X=\sum_{i=1}^n Y_i^2$ where $Y_1,\dots, Y_n\sim \mathcal{N}_b(0, 1)$ are independent $b$-truncated Gaussian random variables. Then, there exist two constants $C_1, C_2$ such that for any $t>0$,
\begin{align*}
    \Pr\left[\left|X - n(1+ \frac{b\phi(b)}{1-\Phi(b)})\right|\geq nt\right] \leq 2\exp\left(-C_1 nt^2 / (b+1)^4\right) +  2\exp\left(-C_2 nt/(b+1)^2\right).
\end{align*}
In particular, we have
\begin{align*}
        \Pr\left[\left|X - n\big(1+ \tfrac{b\phi(b)}{1-\Phi(b)}\big)\right| \geq t\right] \leq 2\exp\left(-C_1 t^2 / (n(b+1)^4)\right) +  2\exp\left(-C_2 t/(b+1)^2\right).
\end{align*}
\end{lemma}
\begin{proof}
Since we know that $Y_i\sim \mathcal{N}_b(0, 1)$ is $C(b+1)$-subgaussian, it implies that $Y_i^2$ is a sub-exponential random variable with parameters $(4\sqrt{2}C^2(b+1)^2, 4C^2(b+1)^2)$. Hence, by the standard concentration of sub-exponential random variables, we have
\begin{align*}
    \Pr\left[\left|\sum_{i=1}^n Y_i^2 - n\E[Y_i^2]
    \right|\geq nt\right] \leq &~ \begin{cases}
    2\exp\left(-\frac{nt^2}{2\cdot 32C^4 (b+1)^4}\right) & \text{if}~nt\leq 8C^2(\textbf{}b+1)^2\\
    2 \exp\left(-\frac{nt}{2\cdot 4C^2 (b+1)^2}\right) & \text{otherwise}
    \end{cases}\\
    \leq &~ 2\exp\left(-C_1 nt^2 / (b+1)^4\right) + 2 \exp\left(-C_2 nt/(b+1)^2\right).
\end{align*}
\end{proof}
\begin{fact}
\label{fact:biased_gaussian}
Let $h\in\R^{d'}$ be a fixed nonzero vector, let $W\in\R^{m\times d'}$ have i.i.d.\ entries $W_{r,j}\sim{\cal N}(0,\frac{2}{m})$, and let $v:=\phi(Wh)$ be the output of a layer of Definition~\ref{def:architecture} with the thresholded activation $\phi$ on the input $h$, with the threshold $\sqrt{2/m}\,b\,\|h\|_2$. Then $g:=\sqrt{m/2}\,Wh/\|h\|_2$ has distribution ${\cal N}(0,I_m)$, and
\begin{align*}
    \|v\|_2^2=\frac{\|h\|_2^2}{mG(b)}\sum_{r\in A}g_r^2,\qquad A:=\{r\in[m]:g_r>b\}.
\end{align*}
The size $N_A:=|A|$ has the binomial distribution ${\cal B}(m,1-\Phi(b))$, and conditioned on $A$, the variables $g_r$ with $r\in A$ are i.i.d.\ with distribution ${\cal N}_{b}(0,1)$. In particular, $\frac{mG(b)\|v\|_2^2}{\|h\|_2^2}$ has the $b$-truncated chi-square distribution $\chi^2_{b,N_A}$ of order $N_A$, and the distribution of $\|v\|_2^2/\|h\|_2^2$ does not depend on $h$.
\end{fact}

\begin{proof}
The rows $W_r$ are i.i.d.\ ${\cal N}(0,\frac{2}{m}I_{d'})$, so the coordinates $\langle W_r,h/\|h\|_2\rangle$ are i.i.d.\ ${\cal N}(0,\frac{2}{m})$, which gives $g$. Since $(Wh)_r=\sqrt{2/m}\,\|h\|_2\,g_r$, it exceeds the threshold $\sqrt{2/m}\,b\,\|h\|_2$ if and only if $g_r>b$. So coordinate $r$ of $v$ equals $\sqrt{c_b}\,{\bf 1}[g_r>b]\sqrt{2/m}\,\|h\|_2\,g_r$, and the formula for $\|v\|_2^2$ follows from $2c_b/m=1/(mG(b))$. The events $\{g_r>b\}$ are independent and each has probability $1-\Phi(b)$. Since the $g_r$ are independent, conditioning on $A$ leaves the $g_r$ with $r\in A$ independent, each distributed as $g_r$ conditioned on $g_r>b$, that is, as ${\cal N}_b(0,1)$.
\end{proof}

\begin{fact}[Gaussian tail bound]\label{fact:gaussian_ulb}
For any $b>0$, we have
\begin{align*}
    \frac{e^{-b^2/2}}{C(b+1)}\leq 1- \Phi(b) \leq e^{-b^2/2},
\end{align*}
where $C$ is an absolute constant.
\end{fact}

We prove a truncated Gaussian version of Lemma 7.1 of~\cite{als19_dnn}. For ReLU, each layer is positively homogeneous, so the ratio $\|h_\ell\|_2^2/\|h_{\ell-1}\|_2^2$ has a distribution that does not depend on the earlier layers. Hence the norm deviations of the layers add up. By Fact~\ref{fact:biased_gaussian}, the same holds for our activation, because the threshold of each layer is proportional to the norm of its input. This fails for the fixed threshold $\sqrt{2/m}\,b$ of the published version, with which the deviations of the squared norms from $1$ can grow geometrically with the depth (Remark~\ref{remark:threshold}).
\begin{lemma}
\label{lem:als_71_shift}
Let $b>0$, suppose $\|x_i\|_2=1$ for all $i\in[n]$, and consider the network of Definition~\ref{def:architecture} at initialization, with the thresholded activation $\phi$ in every layer, including layer $L$. Let $P_b:=1-\Phi(b)$, $k:=m\exp(-b^2/2)$ and $\bar k:=k/((1+b)L^2)$. There is an absolute constant $c_1>0$ such that, for every $\epsilon\in(0,1]$, with probability at least
\begin{align*}
    1-6nL\cdot \exp\Big(-\frac{c_1\epsilon^2P_b\,m}{4L^2}\Big)\geq 1-6nL\cdot\exp\big(-\Omega(\epsilon^2\bar k)\big)
\end{align*}
over $W(0)$, we have $\|h_{i,\ell}\|_2^2\in[1-\epsilon,1+\epsilon]$, and hence $\|h_{i,\ell}\|_2\in[1-\eps,1+\eps]$, for all $i\in [n]$ and $\ell\in [L]$. In particular, with probability at least $1-O(nL)\cdot e^{-\Omega(\bar k)}$, we have $\|h_{i,\ell}\|_2\in[0.9,1.1]$ for all $i\in[n]$ and $\ell\in[L]$.
\end{lemma}

\begin{proof}
Throughout the proof, $P(x):=1-\Phi(x)$, so that $P_b=P(b)$, and
\begin{align*}
    G(x)=\E_{g\sim{\cal N}(0,1)}\big[g^2\,{\bf 1}[g>x]\big]=P(x)+x\,p(x),
\end{align*}
where the second form follows from the integration by parts $\int_x^\infty t^2p(t)\,\d t=x\,p(x)+\int_x^\infty p(t)\,\d t$, which uses $p'(t)=-t\,p(t)$. In particular, $c_b=1/(2G(b))$, $G'(x)=-x^2p(x)$, and $G(x)\geq x\,p(x)$ for $x\geq 0$.

\textbf{Gaussian tail facts.} For $x\geq 0$,
\begin{align}\label{eq:mills}
    \frac{p(x)}{1+x}\leq P(x), \qquad\text{and}\qquad P(x)\leq \frac{p(x)}{x}~~\text{if}~x>0.
\end{align}
The upper bound holds since $P(x)=\int_x^\infty p(t)\,\d t\leq\int_x^\infty \frac{t}{x}\,p(t)\,\d t=\frac{p(x)}{x}$. For the lower bound, the function $u(x):=P(x)-p(x)/(1+x)$ satisfies $u'(x)=-x\,p(x)/(1+x)^2\leq 0$ and $u(x)\to 0$ as $x\to\infty$, so $u\geq 0$ on $[0,\infty)$. As a consequence, the mean of the square of a variable with distribution ${\cal N}_b(0,1)$ is
\begin{align*}
    \mu_b:=\frac{G(b)}{P(b)}=1+\frac{b\,p(b)}{P(b)}\geq 1+b^2\geq \frac{(1+b)^2}{2},
\end{align*}
by the upper bound in~\eqref{eq:mills}.

\textbf{Step 1: one layer.} We first show a concentration bound for the ratio of the squared norms at one layer.

\begin{claim}\label{clm:one_layer}
Let $g\sim{\cal N}(0,I_m)$ and $X:=\frac{1}{mG(b)}\sum_{r\in[m]}g_r^2\,{\bf 1}[g_r>b]$. There is an absolute constant $c_1>0$ such that for every $\delta\in(0,1]$,
\begin{align*}
    \Pr\big[\,|X-1|>\delta\,\big]\leq 6\exp(-c_1\delta^2 m P_b).
\end{align*}
\end{claim}
\begin{proof}[Proof of Claim~\ref{clm:one_layer}]
Let $A:=\{r\in[m]:g_r>b\}$ and $N_A :=|A|\sim{\cal B}(m,P_b)$, and let $X':=\sum_{r\in A}g_r^2=mG(b)\,X$. Conditioned on $A$, the $g_r$ with $r\in A$ are i.i.d.\ with distribution ${\cal N}_b(0,1)$. By the Chernoff bound (Lemma~\ref{lem:chernoff}) with parameter $\delta/3$, the event $|N_A-mP_b|\leq\frac{\delta}{3}mP_b$ fails with probability at most $2\exp(-\delta^2mP_b/27)$. Condition on a set $A$ with $N_A\geq(1-\frac{\delta}{3})mP_b\geq\frac{2}{3}mP_b$. Apply Lemma~\ref{lem:truncated_chi_concentration} to $X'$, with $N_A $ terms, truncation level $b$ and $t:=\delta\mu_b/3$. Since $\mu_b\geq(1+b)^2/2$ and $\delta^2\leq\delta$, we get
\begin{align*}
    \Pr\Big[|X'-N_A\mu_b|\geq\frac{\delta}{3}\,N_A\mu_b~\Big|~A\Big]\leq &~ 2\exp\Big(-\frac{C_1N_A\delta^2}{36}\Big)+2\exp\Big(-\frac{C_2N_A\delta}{6}\Big)\\
    \leq &~ 4\exp\Big(-\frac{2c'}{3}\,\delta^2mP_b\Big),
\end{align*}
where $c':=\min\{C_1/36,C_2/6\}$. Outside these two events,
\begin{align*}
    X'\in\Big[\big(1-\tfrac{\delta}{3}\big)^2,\big(1+\tfrac{\delta}{3}\big)^2\Big]\cdot mP_b\mu_b\subseteq[1-\delta,1+\delta]\cdot mG(b),
\end{align*}
since $P_b\mu_b=G(b)$. So $X\in[1-\delta,1+\delta]$. The claim follows with $c_1:=\min\{1/27,2c'/3\}$.
\end{proof}

\textbf{Step 2: induction over the layers.} Fix $i\in[n]$ and drop it from the notation, so that $h_\ell=h_{i,\ell}$ and $h_0=x_i$. For $\ell\in[L]$, condition on $W_1,\dots,W_{\ell-1}$ such that $h_{\ell-1}\neq 0$. By Fact~\ref{fact:biased_gaussian} with $W=W_\ell$ and $h=h_{\ell-1}$, we have $\|h_\ell\|_2^2=\|h_{\ell-1}\|_2^2\,X_\ell$, where $X_\ell:=\frac{1}{mG(b)}\sum_{r\in A}g_r^2$ has the distribution of $X$ in Claim~\ref{clm:one_layer}, whatever $W_1,\dots,W_{\ell-1}$ are. Let $\delta:=\epsilon/(2L)\leq 1/2$, and let $B_\ell$ be the event that $h_{\ell-1}\neq 0$ and $|X_\ell-1|>\delta$. By Claim~\ref{clm:one_layer},
\begin{align*}
    \Pr[B_\ell]\leq 6\exp(-c_1\delta^2mP_b)=6\exp\Big(-\frac{c_1\epsilon^2P_b\,m}{4L^2}\Big).
\end{align*}
Suppose that none of $B_1,\dots,B_L$ occurs. Since $\|h_0\|_2=\|x_i\|_2=1$, induction on $\ell$ shows that $h_\ell\neq 0$ and
\begin{align*}
    \|h_\ell\|_2^2=\prod_{\ell'=1}^{\ell}X_{\ell'}\in\big[(1-\delta)^\ell,(1+\delta)^\ell\big]
\end{align*}
for every $\ell\in[L]$. Since $(1-\delta)^\ell\geq 1-L\delta=1-\epsilon/2$ and $(1+\delta)^\ell\leq e^{L\delta}=e^{\epsilon/2}\leq 1+\epsilon$ for $\epsilon\leq 1$, we get $\|h_\ell\|_2^2\in[1-\epsilon/2,1+\epsilon]$, and hence $\|h_\ell\|_2\in[\sqrt{1-\epsilon/2},\sqrt{1+\epsilon}]\subseteq[1-\epsilon,1+\epsilon]$. So the deviations of the layers add up, and no factor depending exponentially on $L$ appears. A union bound over $\ell\in[L]$ and $i\in[n]$ shows that this fails with probability at most $6nL\exp(-c_1\epsilon^2P_bm/(4L^2))$.

Finally, $mP_b\geq k/(3(1+b))$ by Fact~\ref{fact:gauss_quant}, so $c_1\epsilon^2P_bm/(4L^2)\geq c_1\epsilon^2\bar k/12$, which gives the second form of the probability bound. The last statement is the case $\epsilon=1/10$.
\end{proof}

\fi
\end{document}